\documentclass[9pt,twocolumn,letterpaper]{article}
\usepackage{graphicx,epsfig,fullpage,svg}
\usepackage[hyphens]{url}
\usepackage{caption}
\usepackage{subfig}
\usepackage{float}
\usepackage{comment}
\usepackage{caption}
\usepackage{mathtools} 
\usepackage{adjustbox}
\usepackage{tabularx}
\usepackage{multirow}
\usepackage{booktabs} 
\usepackage{hyperref}
\usepackage{dirtytalk}

\title{An overview on the evaluated video retrieval tasks at\\ TRECVID 2022}
\author{
George Awad
  \{gawad@nist.gov\}
  Keith Curtis
  \{keith.curtis@nist.gov\}\\
  Information Access Division,
National Institute of Standards and Technology, USA\\\\
Asad A. Butt
  \{asad.butt@nist.gov\}\\
  Johns Hopkins University;\\ Information Access Division,
National Institute of Standards and Technology, USA\\\\
Jonathan Fiscus
  \{jfiscus@nist.gov\}
Afzal Godil
   \{godil@nist.gov\}\\
Yooyoung Lee
  \{yooyoung@nist.gov\}
Andrew Delgado
  \{andrew.delgado@nist.gov\}\\
Eliot Godard
  \{eliot.godard@nist.gov\}\\
Information Access Division,
National Institute of Standards and Technology, USA\\\\
Lukas Diduch
  \{lukas.diduch@nist.gov\}\\
  Dakota-consulting,  USA\\\\
Jeffrey Liu
  \{jeffrey.liu@ll.mit.edu\}\\
  MIT Lincoln Laboratory, USA\\\\
Yvette Graham
  \{graham.yvette@gmail.com\}\\
 ADAPT Centre, Trinity College Dublin, Ireland\\\\
Georges Qu\'{e}not
  \{Georges.Quenot@imag.fr\}\\
  Laboratoire d'Informatique de Grenoble, France\\\\
\\\\
}

\newlength{\spacelen}
\begin{document}
\graphicspath{{tv21.figures/}}
\maketitle

\section{Introduction}
The TREC Video Retrieval Evaluation (TRECVID) is a TREC-style
video analysis and retrieval evaluation with the goal of
promoting progress in research and development of content-based exploitation and retrieval of 
information from digital video via open, tasks-based evaluation supported by metrology. 

Over the last twenty-one years this effort has yielded a better understanding of how systems can effectively accomplish such processing and how one can reliably benchmark their performance. 
TRECVID has been funded by NIST (National Institute of Standards and Technology) and other US government agencies. In addition, many organizations and individuals worldwide contribute significant time and effort.

TRECVID 2022 planned for the following six tasks. From which, four tasks (AVS, VTT, DSDI, \& ActEV) continued from previous years,  while two pilot tasks (DVU and MSUM) were introduced. In total, 35 teams from various research organizations worldwide signed up to join the evaluation campaign this year, where 20 teams
(Table \ref{participants}) completed one or more of the following six tasks, and 15 teams registered but did not submit any runs.

\begin{enumerate} \itemsep0pt \parskip0pt
\item Ad-hoc Video Search (AVS)
\item Video to Text (VTT)
\item Deep Video Understanding (DVU)
\item Disaster Scene Description and Indexing (DSDI)
\item Activities in Extended Video (ActEV)
\item Movie Summarization (MSUM)
\end{enumerate}

This year TRECVID continued the usage of the Vimeo Creative Commons collection dataset (V3C1 and V3C2) \cite{rossetto2019v3c} of about 2,300 hours in total and segmented into 1.5 million short video shots to support the Ad-hoc video search task. The dataset is drawn from the Vimeo video sharing website under the Creative Commons licenses and reflects a wide variety of content, style, and source device determined only by the self-selected donors. The VTT task also adopted a subset of 2008 short videos from the Vimeo V3C1 dataset.

For the ActEV task, about 16 hours of the Multiview Extended Video with Activities (MEVA) dataset was used which was designed to be realistic, natural and challenging for video surveillance domains in terms of its resolution, background clutter, diversity in scenes, and human activity/event categories.

A new licensed movie dataset of about 15 hours was acquired this year from KinoLorberEdu\footnote{https://www.kinolorberedu.com/} and applied to both the MSUM task as well as the DVU tasks. In addition, a set of 14 Creative Common (CC) movies (total duration of 17.5 hr) previously utilized in 2020 and 2021 ACM Multimedia DVU Grand Challenges including their movie-level and scene-level annotations are being utilized as development dataset for the DVU task. The movies have been collected from public websites such as Vimeo and the Internet Archive. In total, the 14 movies consist of 621 scenes, 1572 entities, 650 relationships, and 2491 interactions.

The DSDI task continued to test systems on a new collected public natural disaster 6\thinspace h videos supplied by the Federal Emergency Management Agency (FEMA) from different events inside the USA.

The AVS, DVU, and MSUM results were judged by NIST human assessors, while the VTT task ground-truth was created by NIST human assessors and scored automatically later using Machine Translation (MT) metrics and Direct Assessment (DA) by Amazon Mechanical Turk workers on sampled runs. Full ground-truth was also built for the Disaster Scene Description and Indexing tasks and later on used to score teams' runs.

The systems submitted for the ActEV task evaluations were scored by NIST using reference annotations created by Kitware, Inc.

\begin{table*} 
\caption{Participants and tasks}
\label{participants}
  \vspace{1.0cm}
  \centering{
   \scriptsize{
    \begin{tabular}{|c|c|c|c|c|c|l|l|l|}
      \hline 
\multicolumn{6}{|c|}{Task}&Location&TeamID&Participants \\
      \hline 
$DS$ & $AV$ & $DV$ & $MS$ & $VT$& $AH$ &  &  &    \\
\hline
$--$ & $--$ & $**$ & $--$ & $--$ & $--$ & $Eur$ & $Adapt$&Adapt research centre Dublin City University\\
$--$ & $AV$ & $DV$ & $--$ & $--$ & $--$ & $Asia$ & $alisec\_video$&Alibaba group\\
$--$ & $---$ & $--$ & $--$ & $VT$ & $--$ & $NAm$ & $Arete$&Arete Associates Machine-Learning@arete.com\\
$--$ & $**$ & $--$ & $--$ & $--$ & $--$ & $Asia$ & $BUPT\_MCPRL$&Beijing University of Posts and Telecommunications\\
$--$ & $--$ & $--$ & $--$ & $**$ & $--$ & $NAm$ & $VIDION$&Carnegie Mellon University\\
$--$ & $AV$ & $DV$ & $MS$ & $VT$ & $AH$ & $NAm$ & $drylwlsn$&drylwlsn\_visual\_intelligence\\
$--$ & $--$ & $--$ & $--$ & $**$ & $--$ & $Afr$ & $ELT\_01$&Elyadata\\
$--$ & $--$ & $DV$ & $MS$ & $--$ & $--$ & $Eur$ & $EURECOM$&EURECOM\\
$--$ & $--$ & $DV$ & $--$ & $--$ & $--$ & $NAm$ & $spacetime\_memory$&Facebook AI Research\\
$--$ & $--$ & $--$ & $--$ & $--$ & $**$ & $Asia$ & $kindai\_ogu\_osaka$&Kindai University, Osaka Gakuin University,\\&&&&&&&& Osaka University\\
$--$ & $--$ & $**$ & $--$ & $VT$ & $--$ & $NAm$ & $columbia\_graphen$&Graphen,Inc Columbia University\\
$--$ & $**$ & $--$ & $--$ & $**$ & $--$ & $Asia$ & $MLVC\_HDU$&Hangzhou Dianzi University\\
$DS$ & $--$ & $--$ & $--$ & $--$ & $--$ & $Asia$ & $AIV4$&Hitachi, Ltd.\\
$--$ & $**$ & $--$ & $MS$ & $--$ & $**$ & $Eur$ & $ITI\_CERTH$&Information Technologies Institute, \\&&&&&&&& Centre for Research and Technology Hellas\\
$--$ & $AV$ & $DV$ & $--$ & $--$ & $--$ & $NAm$ & $INF$&Language Technologies Institute - \\&&&&&&&& Carnegie Mellon University\\
$DS$ & $AV$ & $--$ & $--$ & $--$ & $--$ & $Eur$ & $ActiVisionLinks$&LINKS Foundation\\
$--$ & $--$ & $--$ & $--$ & $--$ & $AH$ & $Eur$ & $LIG$&Multimedia Information Modeling and Retrieval group \\&&&&&&&& of LIG Explainable and Responsible Artificial Intelligence \\&&&&&&&& Chair of the MIAI Institute.\\
$--$ & $--$ & $--$ & $--$ & $**$ & $--$ & $Asia$ & $kslab$&Nagaoka University of Technology\\
$DS$ & $AV$ & $DV$ & $**$ & $VT$ & $AH$ & $Asia$ & $NII\_UIT$&National Institute of Informatics, Japan \\&&&&&&&& and University of Information Technology,\\&&&&&&&& VNU-HCMC, Vietnam\\
$--$ & $--$ & $**$ & $--$ & $--$ & $--$ & $Asia$ & $WHU\_NERCMS$& Wuhan University, Wuhan City, Hubei Province, China\\
$DS$ & $--$ & $DV$ & $MS$ & $VT$ & $AH$ & $Eur+Asia$ & $OzuCod$&Ozyegin University\\
$**$ & $AV$ & $--$ & $--$ & $--$ & $AH$ & $Asia$ & $PKU\_WICT$&Peking University\\
$--$ & $--$ & $--$ & $--$ & $**$ & $**$ & $Asia$ & $RUCAIM3-Tencent$&Renmin University of China\\
$--$ & $--$ & $DV$ & $MS$ & $VT$ & $**$ & $Asia$ & $RUCMM$&Renmin University of China\\
$--$ & $--$ & $--$ & $--$ & $--$ & $**$ & $Asia$ & $VIREO$&Singapore Management University \\&&&&&&&& City University of Hong Kong\\
$DS$ & $--$ & $--$ & $--$ & $--$ & $--$ & $Asia$ & $SEUGraphDy$&Southeast University\\
$--$ & $**$ & $--$ & $--$ & $--$ & $--$ & $Asia$ & $TokyoTech$&Tokyo Institute of Technology\\
$--$ & $--$ & $--$ & $--$ & $VT$ & $**$ & $SAm$ & $CamiloUchile$&Uchile\\
$--$ & $AV$ & $--$ & $--$ & $--$ & $--$ & $Asia$ & $MMGofLDS$&University of Science and Technology of China\\
$--$ & $--$ & $DV$ & $MS$ & $VT$ & $AH$ & $Eur$ & $MiguelUA$&University of Alicante, Alicante, SPAIN.\\
$**$ & $--$ & $--$ & $--$ & $--$ & $--$ & $NAm$ & $UMKC$&University of Missouri-Kansas City\\
$--$ & $--$ & $--$ & $--$ & $--$ & $--$ & $NAm$ & $USF\_EE$&University of South Florida\\
$--$ & $**$ & $--$ & $--$ & $**$ & $**$ & $Asia$ & $WasedaMeiseiSoftbank$&Waseda University, Meisei University,\\&&&&&&&& SoftBank Corporation\\
$DS$ & $AV$ & $DV$ & $MS$ & $VT$ & $AH$ & $NAm$ & $yorku22$&York University\\
$--$ & $**$ & $--$ & $--$ & $--$ & $--$ & $NAm$ & $UMD$&University of Maryland\\
      \hline 
   \end{tabular} \\
\vspace{.2cm}
Task legend. DV:Deep Video Understanding; VT:Video to Text; AV:Activities in Extended videos; AH:Ad-hoc search; DS: Disaster Scene Description and Indexing; MS: Movie Summarization; $--$:no run planned; $**$:submitted run(s)\\
   } 
  } 
\end{table*}

This paper is an introduction to the tasks, data, evaluation framework, and performance measures used in the 2022 evaluation campaign. For detailed information about the approaches and results, the reader should see the various site reports and the results pages available at the workshop proceeding online page \cite{tv22pubs}.
Finally, we would like to acknowledge that all work presented here has been cleared by RPO (Research Protection Office)\footnote{ under RPO number: \#ITL-17-0025}

\emph{Disclaimer: Certain commercial equipment, instruments, software, or materials, commercial or non-commercial, are identified in this paper in order to specify the experimental procedure adequately. Such identification does not imply recommendation or endorsement of any product or service by NIST, nor does it imply that the materials or equipment identified are necessarily the best available for the purpose. }

\section{Datasets}
Many datasets have been adopted and used across the years since TRECVID started in 2001 and all available resources and datasets from previous years can be accessed from our website\footnote{https://trecvid.nist.gov/past.data.table.html}. In the following sections we will give an overview of the main datasets used this year across the different tasks.

\subsection{DVU Movies Training Dataset}
A set of 14 Creative Common (CC) movies (total duration of 17.5 hr) collected from public websites such as Vimeo and the Internet Archive and previously utilized in 2020 and 2021 ACM Multimedia DVU Grand Challenges has been introduced this year to support the Deep Video Understanding task. The dataset includes movie-level and scene-level annotations for 621 scenes, 1572 entities, 650 relationships, and 2491 interactions. In addition, key characters and locations image snapshots were provided.

\subsection{Kinolorberedu Testing Dataset}
A set of 10 movies licensed from Kino Lorber Edu (https://www.kinolorberedu.com/) are made available to support the MSUM and DVU tasks. All movies are in English with duration between 1.5 - 2 hrs each. This year, 6 movies have been used as testing dataset for the DVU task, while for movie summarization, the dataset was split into 5 movies for training and 5 for testing.

\subsection{Vimeo Creative Commons Collection (V3C) Dataset}
Two sub-collections (V3C1 and V3C2) \cite{rossetto2019v3c} have been adopted to support the AVS task. 
Together, they are composed of about 17,000 Vimeo videos (2.9 TB, 2300 h) with Creative Commons licenses and mean duration of 8 min. All videos have some metadata available such as title, keywords, and description in json files. They have been segmented into 2\,508\,113 short video segments according to the provided master shot boundary files. In addition, keyframes and thumbnails per video segment have been extracted and made available. V3C2 was used for testing, while V3C1 was available for development along with the previous Internet Archive datasets (IACC.1-3) of about 1800 h. In addition to the above, a subset of short videos from V3C1 dataset was used to test the Video to Text systems.

\subsection{MEVA Dataset} 
The ActEV Sequestered Data Leaderboard (SDL) competition is based on the Multiview Extended Video with Activities (MEVA) dataset (\cite{MEVAdata} mevadata.org) which was collected and annotated specifically for the development and evaluation of public safety video activity detection capabilities at the Muscatatuck Urban Training Center by Kitware, Inc. for the IARPA DIVA (Deep Intermodal Video Analytics) program and the broader research community. This dataset contains time-synchronized multi-camera, continuous, long-duration video, often taken at significant stand-off ranges from the activities. Metadata and auxiliary data for the site were provided as is typical for public-safely scenarios where detailed knowledge of the site is available to systems. Provided data will include a map and 3D site model of the test area, approximate camera locations for the publicly released video data, and camera models for released sensor video. The dataset was collected with both EO (Electro-Optical) and IR (Infrared) sensors, with over 100 actors performing in various scripted and non-scripted activities in various scenarios. The activities included person and multi-person activities, person-object interaction activities, vehicle activities, and person-vehicle interaction activities.

The dataset was captured with off-the-shelf cameras with fields of view, which contains both overlapping and non-overlapping, and 25 of them are EO cameras and 4 are IR cameras. The IR cameras are paired with EO cameras with roughly the same location and orientation. The spatial resolution of the EO cameras is 1920x1080 or 1920x1072 and the IR cameras is 352x240. All the video cameras have a frame rate of 30 frames/second, have a fixed orientation except one, and all are synchronized with the GPS time signal. The number of indoor cameras is 11 and the number of outdoor cameras is 18, all the IR cameras which are paired with EO cameras having the same position and orientation are outdoor. 

\subsubsection{Test Data}
The TRECVID’22 ActEV Self-Reported Leaderboard (SRL) test dataset is a 16-hour collection of videos with 20 activities, which only consists of Electro-Optics (EO) camera modalities from public cameras. The TRECVID’22 ActEV SRL test dataset is the same as the one used for CVPR ActivityNet 2022 ActEV SRL and the WACV’22 ActEV SRL challenges.

\subsubsection{Training and Development Data}
In December 2019, the public MEVA dataset was released with 328 hours of ground-camera data and 4.2 hours of Unmanned Arial Vehicle video. 160 hours of the ground camera video have been annotated by the same team that has annotated the ActEV test set. Additional annotations have been performed by the public and are also available in the annotation repository.

\begin{figure}[hbt]
\begin{centering}
\includegraphics[width=3.16667in,height=1.86000in]{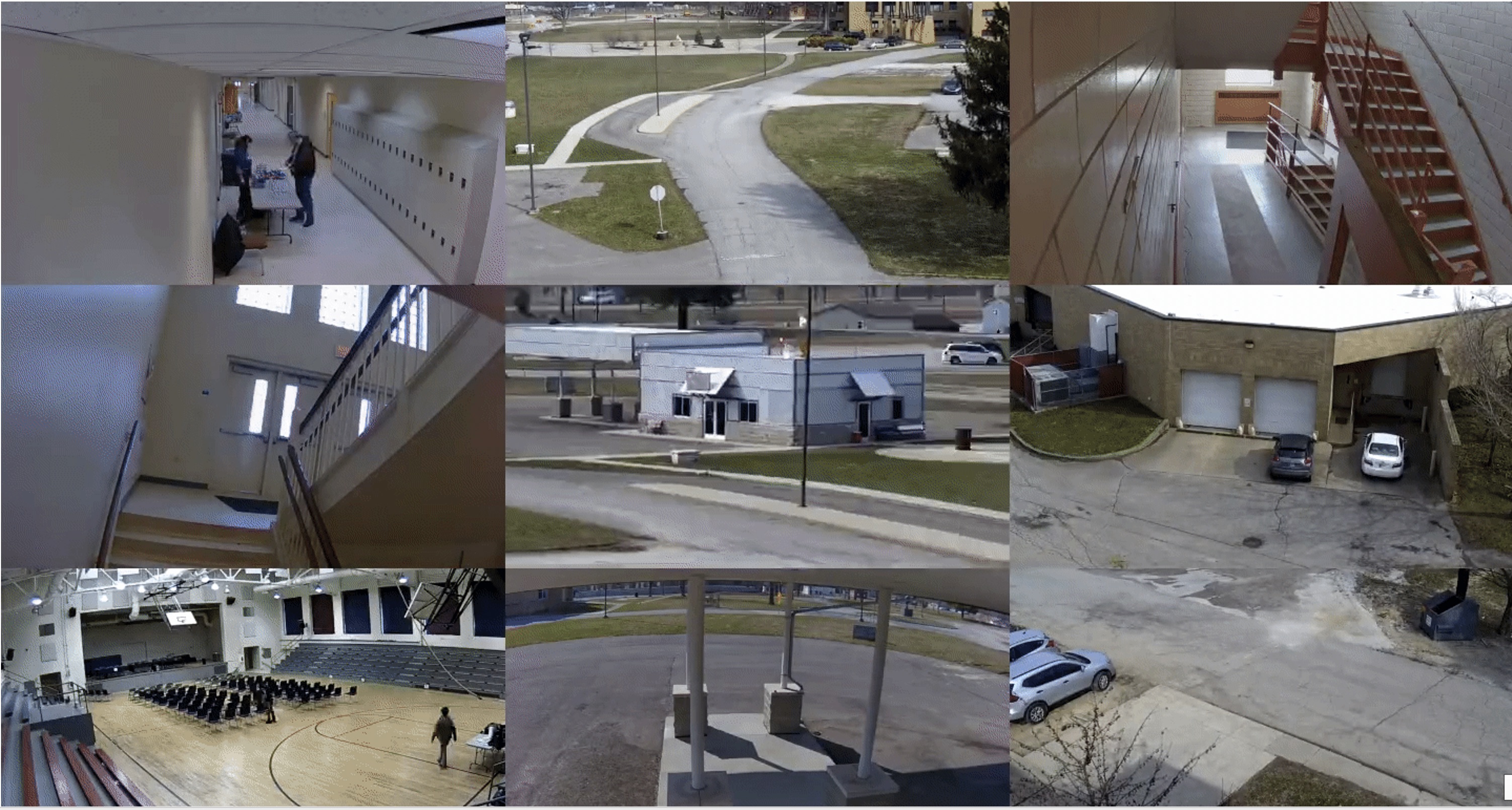}
\caption{Montage of randomly selected video clips}
\label{fig_actev_dataset}
\end{centering}
\end{figure}

\subsection{TRECVID-VTT}
This dataset contains short videos that are between 3 seconds and 10 seconds long. The video sources are from Twitter Vine, Flickr, and V3C2. The dataset is being updated annually and in total, there are 10,862 videos with captions. Each video has between 2 and 5 captions, which have been written by dedicated annotators. The collection includes 6475 URLs from Twitter Vine and 4,387 video files in webm format and Creative Commons License. Those 4,387 videos have been extracted from Flickr and the V3C2 dataset. This year a newly selected 2008 V3C1 videos were used as a testing set.

\subsection{Low Altitude Disaster Imagery (LADI)}
The LADI dataset consists of over 20\,000 annotated images, each at least 4 MB in size, and was available as a development dataset for the DSDI systems. The images were collected by the Civil Air Patrol from various natural disaster events. The raw images were previously released into the public domain. Key distinctions of the dataset are the low altitude (less than 304.8 m (1000 ft)), oblique perspective of the imagery, and disaster-related features, which are rarely featured in computer vision benchmarks and datasets. The dataset currently employs a hierarchical labeling scheme of five coarse categories and then more specific annotations for each category. The initial dataset focuses on the Atlantic Hurricane and spring flooding seasons since 2015.


\section{Evaluated Tasks}
\subsection{Ad-hoc Video Search}
The Ad-hoc Video Search (AVS) task finished a 3-year cycle, 2019-2021, running on the Vimeo dataset (V3C1) as the testing dataset. 
Teams and coordinators found it useful to continue the task again starting this year for a second 3-year cycle where the V3C2 dataset is adopted as the testing dataset. The task is aiming to model the end user video search use case, who is looking for segments of video containing people, objects, activities, locations, etc., and combinations of the former. Compared to previous years, a higher focus on more fine-grained descriptions was given to provided queries. The task was coordinated by NIST and by the Laboratoire d'Informatique de Grenoble.

The task for participants was defined as the following: given a standard set of master shot boundaries (about 1.4 million shots defined by starting time and ending time in the original whole videos) from the V3C2 test collection and a list of 30 ad-hoc textual queries (see Appendix \ref{appendixA} and \ref{appendixB}), participants were asked to return for each query, at most the top 1000 video clips from the master shot boundary reference set, ranked according to the highest probability of containing the target query. The presence of each query was assumed to be binary, i.e., it was either present or absent in the given standard video shot. 
Judges at NIST followed several rules in evaluating system output. For example, if the query was true for some frame (sequence) within the shot, then it was true for the shot. In addition, query definitions such as “contains x” or words to that effect are short for “contains x to a degree sufficient for x to be recognizable as x by a human”. This means among other things that unless explicitly stated, partial visibility or audibility may suffice. Lastly, the fact that a segment contains video of a physical object representing the query target, such as photos, paintings, models, or toy versions of the target (e.g picture of Barack Obama vs Barack Obama himself), was NOT grounds for judging the query to be true for the segment. Containing video of the target within video (such as a television showing the target query) may be grounds for doing so. Three main submission types were accepted:

\begin{itemize}
\item{Fully automatic runs (no human input in the loop): The system takes a query as input and produces results without any human intervention.}
\item{Manually-assisted runs: where a human can formulate the initial query based on topic and query interface, not on knowledge of collection or search results. The system takes the formulated query as input and produces results without further human intervention.}
\item{Relevance-Feedback: The system takes the official query as input and produces initial results, then a human judge can assess the top-30 results and input this information as feedback to the system to produce a final set of results. This feedback loop is strictly permitted for only up to 3 iterations.}
\end{itemize}

In general, runs submitted were allowed to choose any of the below four training types:

\begin{itemize}
\item{A - used only V3C1 training data}
\item{D - used any other training data (except the testing dataset V3C2)}
\item{E - used only training data collected automatically using only the official query textual description}
\item{F - used only training data collected automatically using a query built manually from the given official query textual description}
\end{itemize}

The training categories “E” and “F” are motivated by the idea of promoting the development of methods that permit the indexing of concepts in video clips using only data from the web or archives without the need of additional annotations. The training data could for instance consist of images or videos retrieved by a general-purpose search engine (e.g., Google) using only the query definition with only automatic processing of the returned images or videos. 

The progress subtask previously introduced in 2019-2021 with the objective of measuring system progress on a set of 20 fixed topics (Appendix \ref{appendixB}) was planned again starting this year. As a result, teams were allowed to submit results for 20 common topics (not evaluated this year) that will be fixed for three years (2022-2024). In general, the 20 fixed progress topics are divided equally into two sets of 10 topics. The first set will be evaluated in 2023 to measure system progress for two years (2022-2023), while the second set will be evaluated in 2024 to measure progress over three years.

A Novelty run type was also allowed to be submitted within the main task. The goal of this run type is to encourage systems to submit novel and unique relevant shots not easily discovered by other runs. In other words, to find rare true positive shots. Finally, teams were allowed to submit an optional explainability parameter with each shot. This was formulated as a keyframe and bounding box to localize the region that supports the query evidence.

\subsubsection{Dataset}
The V3C2 dataset (drawn from a larger V3C video dataset \cite{rossetto2019v3c}) was adopted as a testing dataset. It is composed of 9760 Vimeo videos (1.6 TB, 1300 h) with Creative Commons licenses and mean duration of 8 min. All videos have some metadata available e.g., title, keywords, and description in json files. The dataset has been segmented into 1\,425\,454 short video segments according to the provided master shot boundary files. In addition, keyframes and thumbnails per video segment have been extracted and made available.
For training and development, all previous V3C1 dataset (1000 h) and Internet Archive datasets (IACC.1-3) with about 1\,800 h were made available with their ground truth and xml meta-data files. Throughout this report we do not differentiate between a clip and a shot and thus they may be used interchangeably.

\subsubsection{Evaluation}
Each group was allowed to submit up to 4 prioritized runs per submission type and per task type (main or progress), and two additional if they were of training type “E” or “F” runs. In addition, one novelty run type was allowed to be submitted within the main task.

In fact, 7 groups submitted a total of 61 runs with 33 main runs and 28 progress runs. One team submitted a novelty run. The 33 main runs consisted of 28 fully automatic, and 5 manually-assisted runs, while the progress runs consisted of 23 fully automatic and 5 manually-assisted runs. 

To prepare the results from teams for human judgments, a workflow was adopted to pool results from runs submitted. For each query topic, a top pool was created using 100 \% of clips at ranks 1 to 300 across all submissions after removing duplicates. A second pool was created using a sampling rate at 25 \% of clips at ranks 301 to 1000, not already in the top pool, across all submissions and after removing duplicates. Using these two master pools, we divided the clips in them into small pool files with about 1000 clips in each file. Five human judges (assessors) were presented with the pools - one assessor per topic - and they judged each shot by watching the associated video and listening to the audio then voting if the clip contained the query topic or not. Once the assessor completed judging for a topic, a second round of confirmation judging was conducted to take into consideration close neighborhood shots with opposite judging decisions as well as clips submitted by at least 10 runs at ranks 1 to 200 that were voted as false positive by the assessor. This final step was done as a secondary check on the assessors' judging work and to give them an opportunity to fix any judgment mistakes. 

In all, 148\,234 clips were judged while 150\,332 clips fell into the unjudged part of the overall samples. Total hits across the 30 topics reached 20\,125 with 7762 hits at submission ranks from 1 to 100, 7745 hits at submission ranks 101 to 300, and 4618 hits at submission ranks between 301 to 1000. Table \ref{avsstats} presents information about the pooling and judging per topic.

\subsubsection{Measures}
Work at Northeastern University \cite{yilmaz06} has resulted in methods for estimating standard system performance measures using relatively small samples of the usual judgment sets so that larger numbers of features can be evaluated using the same amount of judging effort. Tests on past data showed the measure inferred average precision (infAP) to be a good estimator of average precision \cite{tv6overview}. This year mean extended inferred average precision (mean xinfAP) was used which permits sampling density to vary \cite{yilmaz08}. This allowed the evaluation to be more sensitive to clips returned below the lowest rank ($\approx$300) previously pooled and judged. It also allowed adjustment of the sampling density to be greater among the highest ranked items that contribute more average precision than those ranked lower.
The {\em sample\_eval} software \footnote{http://www-nlpir.nist.gov/projects/trecvid/\\trecvid.tools/sample\_eval/}, a tool implementing xinfAP, was used to calculate inferred recall, inferred precision, inferred average precision, etc., for each result, given the sampling plan and a submitted run. Since all runs provided results for all evaluated topics, runs can be compared in terms of the mean inferred average precision across all evaluated query topics. 

\begin{table*}[t]
\caption{Ad-hoc search pooling and judging statistics}
\label{avsstats} 
  \vspace{0.5cm}
  \centering{
   \small{
    \begin{tabular}{|c|c|c|c|c|c|c|c|}
      \hline 

\parbox{1.3cm}{Topic \\ number} & 
\parbox{1.5cm}{Total \\ submitted} & 
\parbox{1.5cm}{Unique \\ submitted} & 
\parbox{1.0cm}{total \\ that \\ were \\ unique\\ \%} & 
\parbox{1.2cm}{Number \\ judged} & 
\parbox{1.0cm}{unique \\ that \\ were \\ judged\\ \%} & 
\parbox{1.2cm}{Number \\ relevant} & 
\parbox{1.2cm}{judged \\ that \\ were \\ relevant\\ \%}

 \\ \hline
1701 & 33000 & 29188 & 88.45 & 4274 & 14.64 & 528 & 12.35\\ \hline 
1702 & 33000 & 29232 & 88.58 & 5323 & 18.21 & 430 & 8.08\\ \hline 
1703 & 33000 & 29196 & 88.47 & 4426 & 15.16 & 1159 & 26.19\\ \hline 
1704 & 33000 & 29190 & 88.45 & 4179 & 14.32 & 1148 & 27.47\\ \hline 
1705 & 33000 & 29175 & 88.41 & 4691 & 16.08 & 920 & 19.61\\ \hline 
1706 & 33000 & 29166 & 88.38 & 4794 & 16.44 & 1134 & 23.65\\ \hline 
1707 & 33000 & 29178 & 88.42 & 4271 & 14.64 & 205 & 4.80\\ \hline 
1708 & 33000 & 29241 & 88.61 & 5786 & 19.79 & 1151 & 19.89\\ \hline 
1709 & 33000 & 29213 & 88.52 & 5351 & 18.32 & 1554 & 29.04\\ \hline 
1710 & 33000 & 29230 & 88.58 & 6794 & 23.24 & 304 & 4.47\\ \hline 
1711 & 33000 & 29239 & 88.60 & 5397 & 18.46 & 213 & 3.95\\ \hline 
1712 & 33000 & 29220 & 88.55 & 5990 & 20.50 & 1018 & 16.99\\ \hline 
1713 & 33000 & 29268 & 88.69 & 6450 & 22.04 & 338 & 5.24\\ \hline 
1714 & 33000 & 29185 & 88.44 & 4653 & 15.94 & 192 & 4.13\\ \hline 
1715 & 33000 & 29212 & 88.52 & 4584 & 15.69 & 641 & 13.98\\ \hline 
1716 & 33000 & 29167 & 88.38 & 5327 & 18.26 & 109 & 2.05\\ \hline 
1717 & 33000 & 29178 & 88.42 & 3815 & 13.07 & 400 & 10.48\\ \hline 
1718 & 33000 & 29221 & 88.55 & 5168 & 17.69 & 1481 & 28.66\\ \hline 
1719 & 33000 & 29162 & 88.37 & 4389 & 15.05 & 378 & 8.61\\ \hline 
1720 & 33000 & 29099 & 88.18 & 4328 & 14.87 & 327 & 7.56\\ \hline 
1721 & 33000 & 29247 & 88.63 & 6085 & 20.81 & 1049 & 17.24\\ \hline 
1722 & 33000 & 29209 & 88.51 & 4928 & 16.87 & 129 & 2.62\\ \hline 
1723 & 33000 & 29219 & 88.54 & 4272 & 14.62 & 1865 & 43.66\\ \hline 
1724 & 33000 & 29222 & 88.55 & 4381 & 14.99 & 648 & 14.79\\ \hline 
1725 & 33000 & 29134 & 88.28 & 4407 & 15.13 & 232 & 5.26\\ \hline 
1726 & 33000 & 28913 & 87.62 & 5122 & 17.72 & 1398 & 27.29\\ \hline 
1727 & 33000 & 29041 & 88.00 & 4032 & 13.88 & 73 & 1.81\\ \hline 
1728 & 33000 & 29236 & 88.59 & 4354 & 14.89 & 295 & 6.78\\ \hline 
1729 & 33000 & 29179 & 88.42 & 4524 & 15.50 & 618 & 13.66\\ \hline 
1730 & 33000 & 29259 & 88.66 & 6139 & 20.98 & 188 & 3.06\\ \hline 

    \end{tabular}
   } 
  } 
\end{table*}

\subsubsection{Ad-hoc Results}

\begin{figure*}[hbtp]
\begin{center}
\includegraphics[height=3.5in,width=6.5in,angle=0]{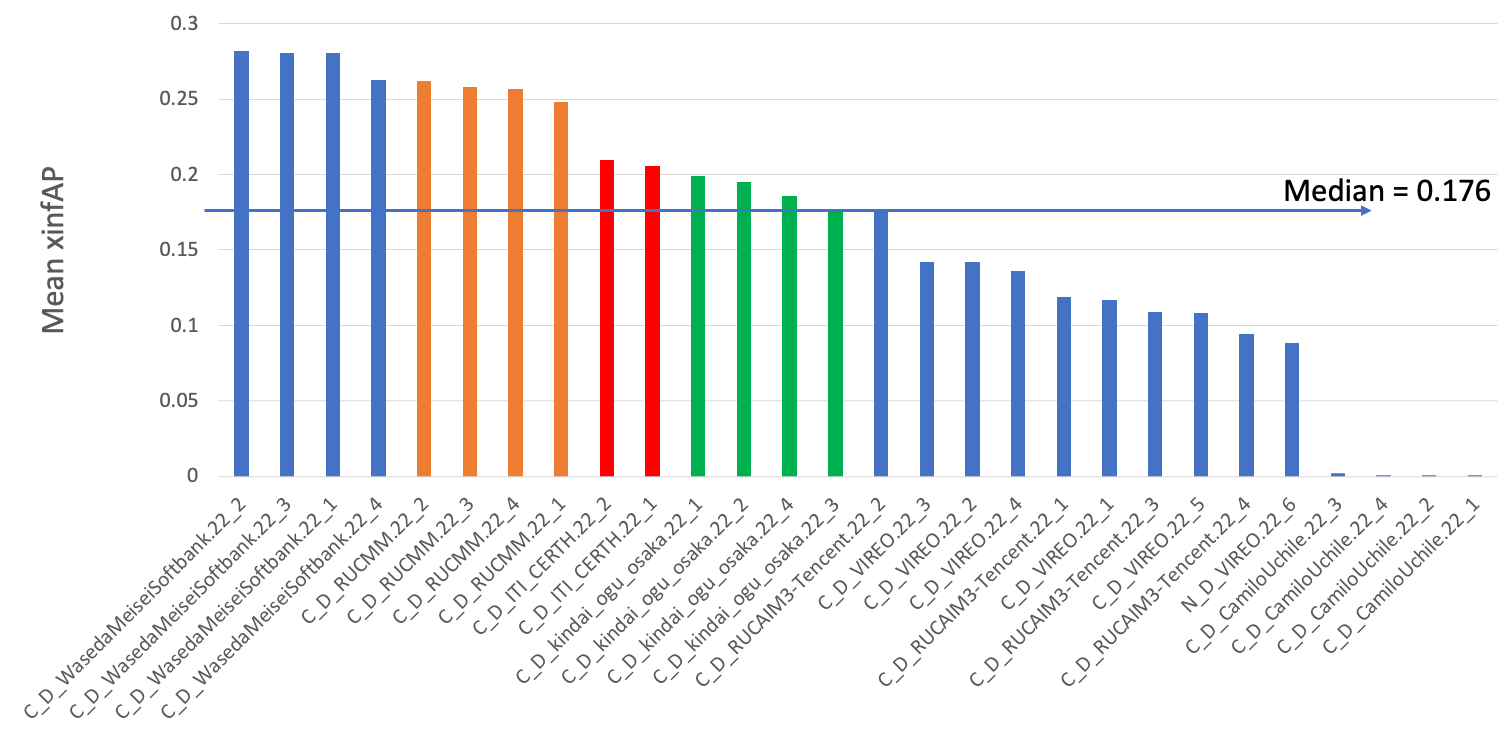}
\caption{AVS: 28 automatic runs across 30 main queries}
\label{avs.f.all}
\end{center}
\end{figure*}

\begin{figure}[hbtp]
\begin{center}
\includegraphics[height=1.5in,width=3.0in,angle=0]{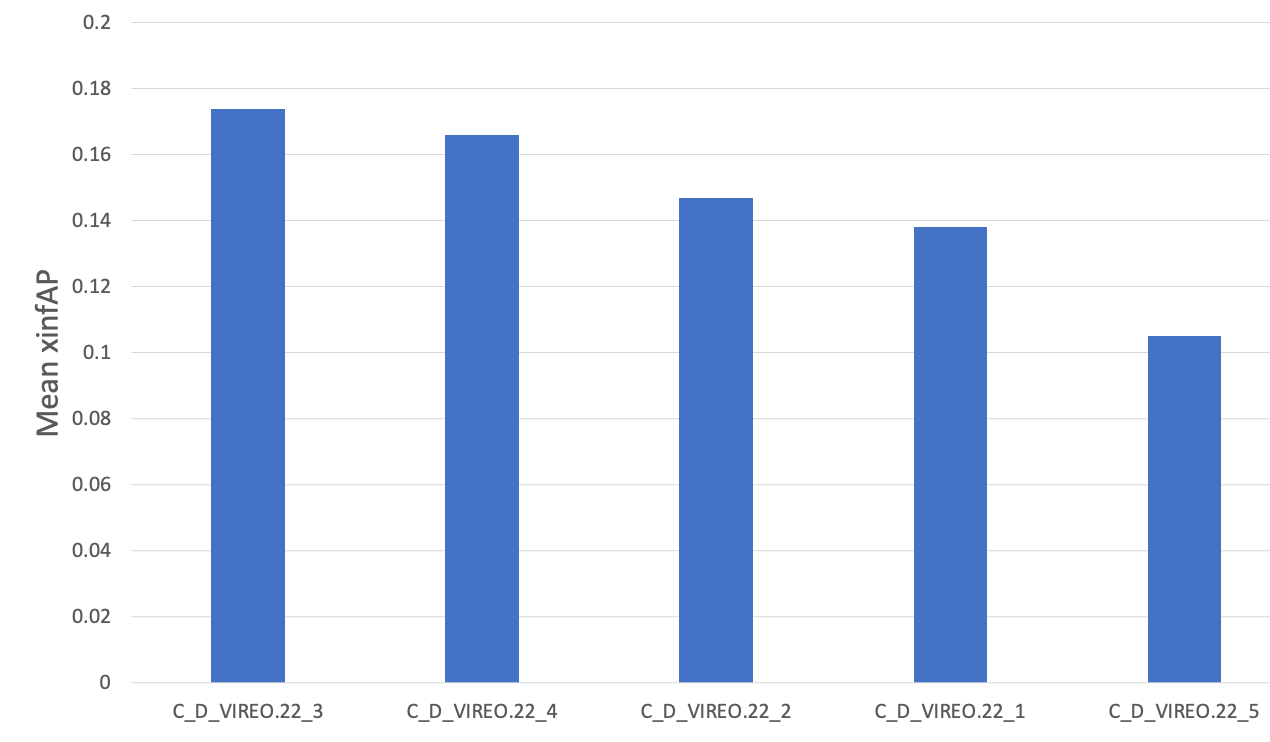}
\caption{AVS: 5 manually-assisted runs across 30 main queries}
\label{avs.m.all}
\end{center}
\end{figure}

Figures \ref{avs.f.all} and \ref{avs.m.all} show the results of all the 28 fully automatic runs and 5 manually-assisted submissions respectively. All manually-assisted runs came from only 1 team (VIREO) where its top run performed higher than its best automatic run. In general each team's runs are very close to each other's performance and we can see that the top 4 teams had their automatic runs clustered together. The top performance this year is less than the highest achieved in the past two years in this task. This may be due to the fact that queries focused on more fine-grained information needs. Overall, automatic runs scored higher than manually-assisted ones and had a median score of 0.176 compared to 0.147 for manually-assisted runs. We should also note here that all submissions were of type 'D', and no runs using category “E” or “F” were submitted. Also, while the evaluation supported relevance feedback run types, this year no submissions were received under this category.

To test if there were significant differences between the runs submitted, we applied a randomization test \cite{manly97} on the top 10 runs for manually-assisted and automatic run submissions using a significance threshold of p$<$0.05. 

For automatic runs, the analysis showed that there were no statistical differences between runs 1,2, and 3 of the WasedaMeiseiSoftbank team. Also the same conclusion applies to all RUCMM runs. However, it was shown that ITI\_CERTH run 2 is better than run 1. For manually-assisted runs, the analysis showed that VIREO's team run 3 is better than all other VIREO runs. In addition, run 4 is better than runs 1, 2, and 5. Finally, runs 1 and 2 are better than run 5.

\begin{figure*}[hbtp]
\begin{center}
\includegraphics[height=3.5in,width=6.5in,angle=0]{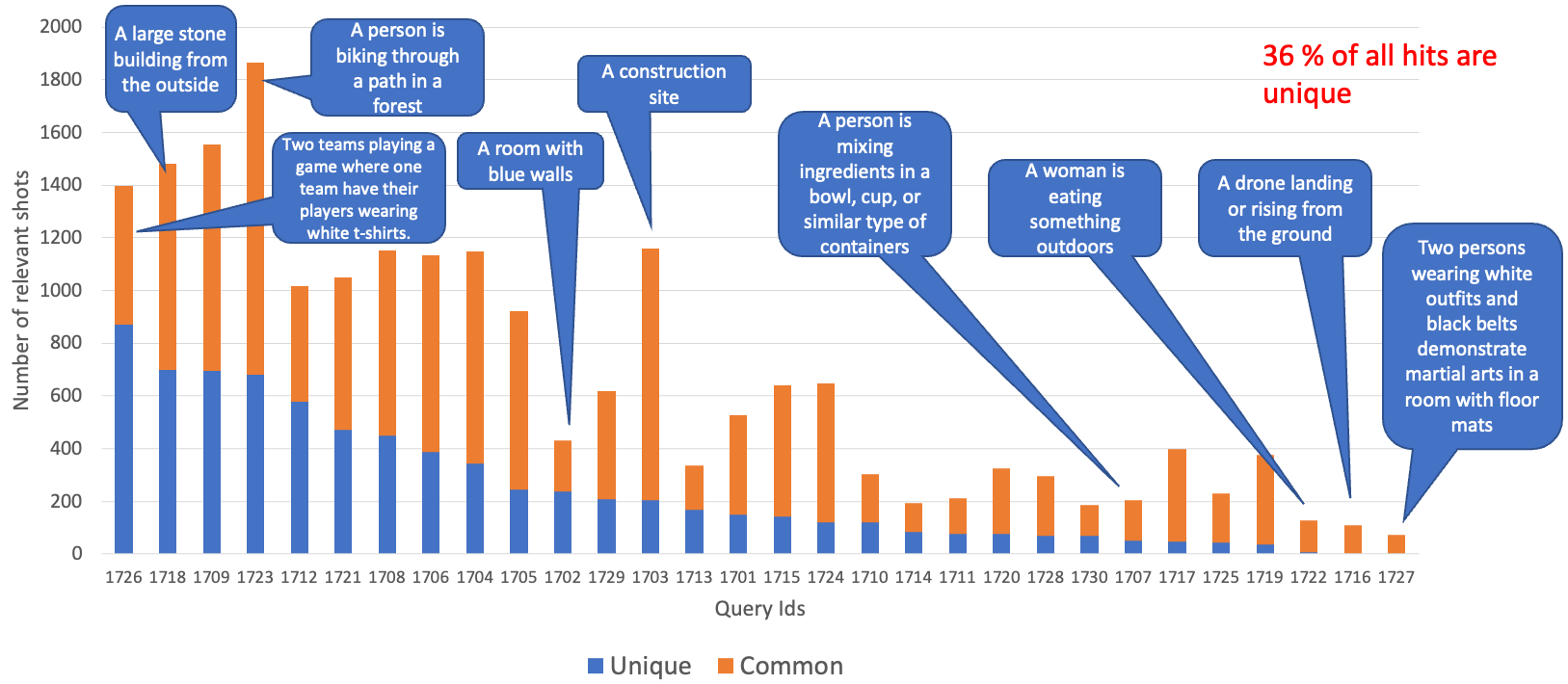}
\caption{AVS: Unique vs overlapping results in main task}
\label{avs.unique.overlapped}
\end{center}
\end{figure*}

\begin{figure}[htbp]
\begin{center}
\includegraphics[height=2.0in,width=3in,angle=0]{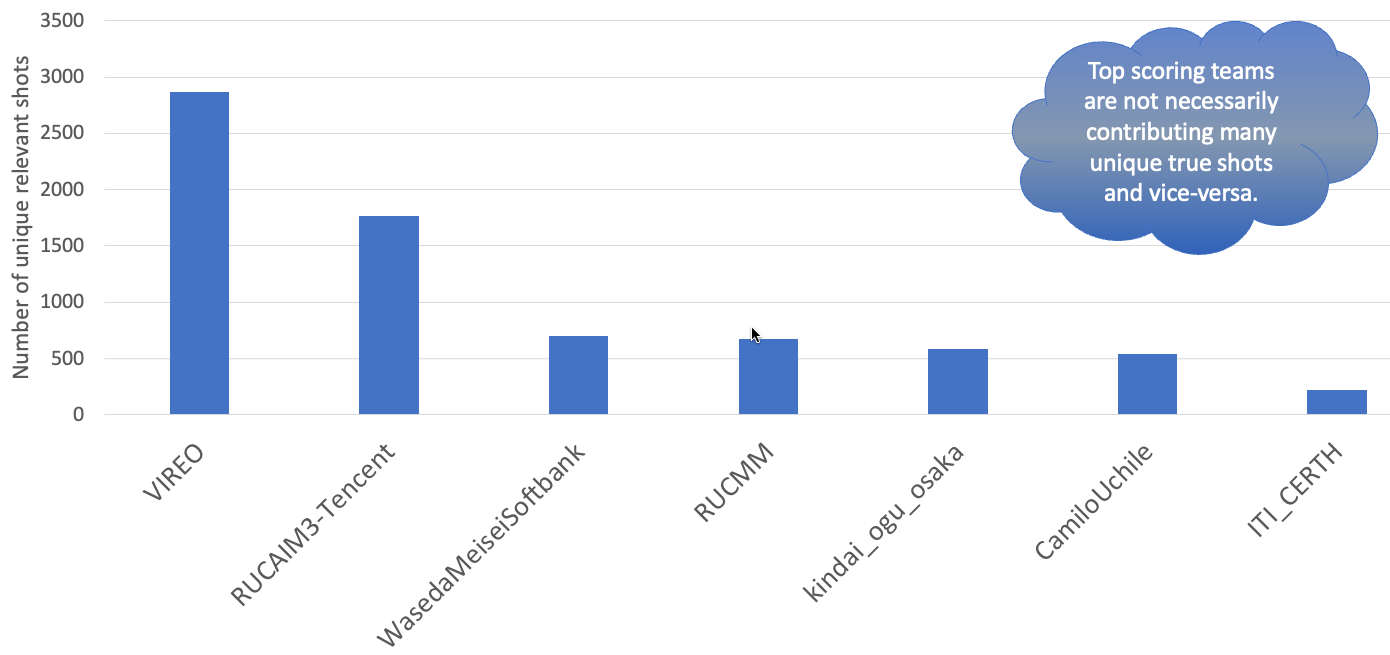}
\caption{AVS: 7336 unique shots contributed by teams in main task}
\label{avs.unique.byteam}
\end{center}
\end{figure}

Figure \ref{avs.unique.overlapped} shows for each topic the number of relevant and unique shots submitted by all teams combined (blue color). On the other hand, the orange bars show the total non-unique true shots submitted by at least 2 or more teams. The chart is sorted by number of unique hits.

The four topics: 1726, 1718, 1709, and 1723 achieved the most unique hits overall while also reporting a high number of hits overall, while the three topics: 1722, 1716, and 1727 reported the lowest unique hits. In general, topics that reported a high number of hits consisted of both unique and non-unique hits, while topics that reported low number of hits mainly only consisted of non-unique hits, representing the difficulty of the query. While it is hard to draw conclusions about why hits vary by topic, there seems to be a correlation with the relative easiness of the query and its components (e.g. more actions/activities in combination with objects or conditions are harder and are being detected less). We should also note here that high/low hits per topic don’t necessarily mean high/low performance in InfAP as a good run must detect and rank results high as well.

Figure \ref{avs.unique.byteam} shows the number of unique clips found by the different participating teams. From this figure and the overall scores in Figures \ref{avs.f.all} and \ref{avs.m.all}, it can be shown that there is no clear relation between teams who found the most unique shots and their total performance. The VIREO team contributed the most unique hits (similar to previous year). Although Waseda, RUCMM, ITI\_CERTH teams performed well, their unique hits contributions were not very high.

\begin{figure*}[htbp]
\begin{center}
\includegraphics[height=3.0in,width=6.5in,angle=0]{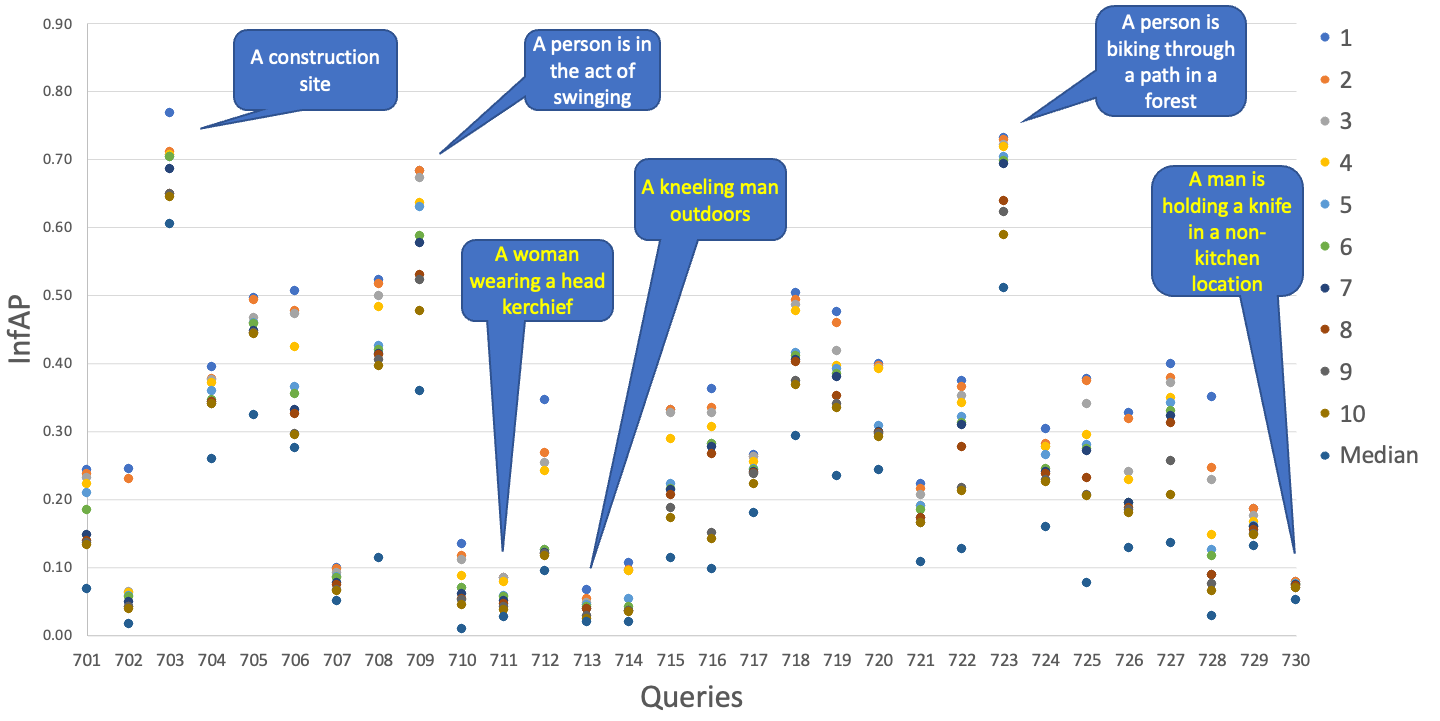}
\caption{AVS: Top 10 runs (xinfAP) per query (fully automatic)}
\label{avs.top10.f}
\end{center}
\end{figure*}

\begin{figure*}[htbp]
\begin{center}
\includegraphics[height=3.0in,width=6.5in,angle=0]{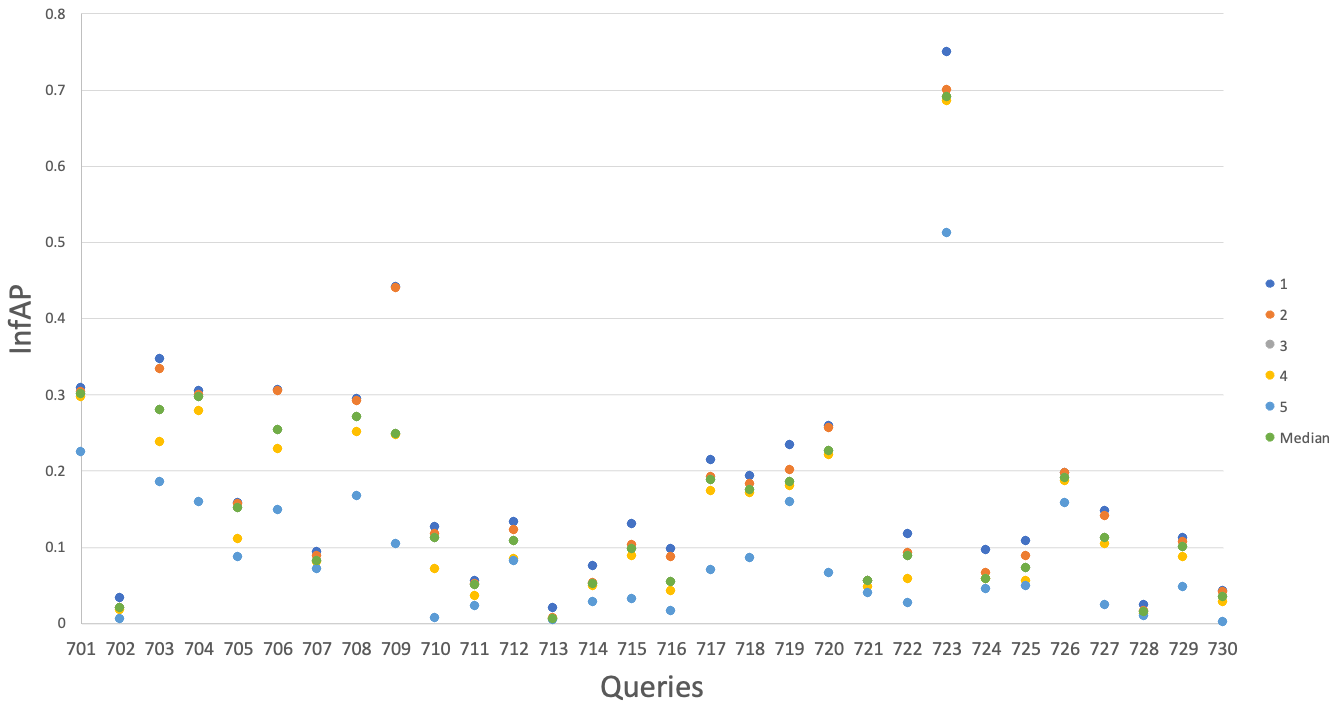}
\caption{AVS: Top runs (xinfAP) per query (manually assisted)}
\label{avs.top10.m}
\end{center}
\end{figure*}

Figures \ref{avs.top10.f} and \ref{avs.top10.m} show the performance of the top 10 runs across the 30 main queries for automatic and manually-assisted runs. Note that each series in this plot represents a rank (from 1 to 10) of the scores, but all scores at a given rank do not necessarily belong to a specific team. A team's scores can rank differently across the 30 queries. Some samples of top and bottom performing queries are highlighted with the query text. From the figures, we can see a high similarity between automatic and manually-assisted systems in terms of query performance relative to each other. Harder queries are those that included non-traditional combinations of concepts (e.g. A man is holding a knife in a non-kitchen location, or a kneeling man outdoors). In general, for automatic systems and for topics not performing well, usually all top 10 runs are condensed together with low spread between their scores, while mid or high performing queries may vary in their range of performance.

The novelty run type encourages submitting unique (hard to find) relevant shots. Systems were asked to label their runs as either novelty type or common type. A new novelty metric was designed to score runs based on how good they are at detecting unique relevant shots. A weight was given to each topic and shot pair such as follows:

\[TopicX\_ShotY_{weight} (x) = 1 - \frac{N}{M}\]

\noindent where N is the number of times Shot Y was retrieved for topic X by any run submission, and M is the number of total runs submitted by all teams. For instance, a unique relevant shot weight will be close to 1.0 while a shot submitted by all runs will be assigned a weight of 0.

\begin{figure}[htbp]
\begin{center}
\includegraphics[height=2.0in,width=3.0in,angle=0]{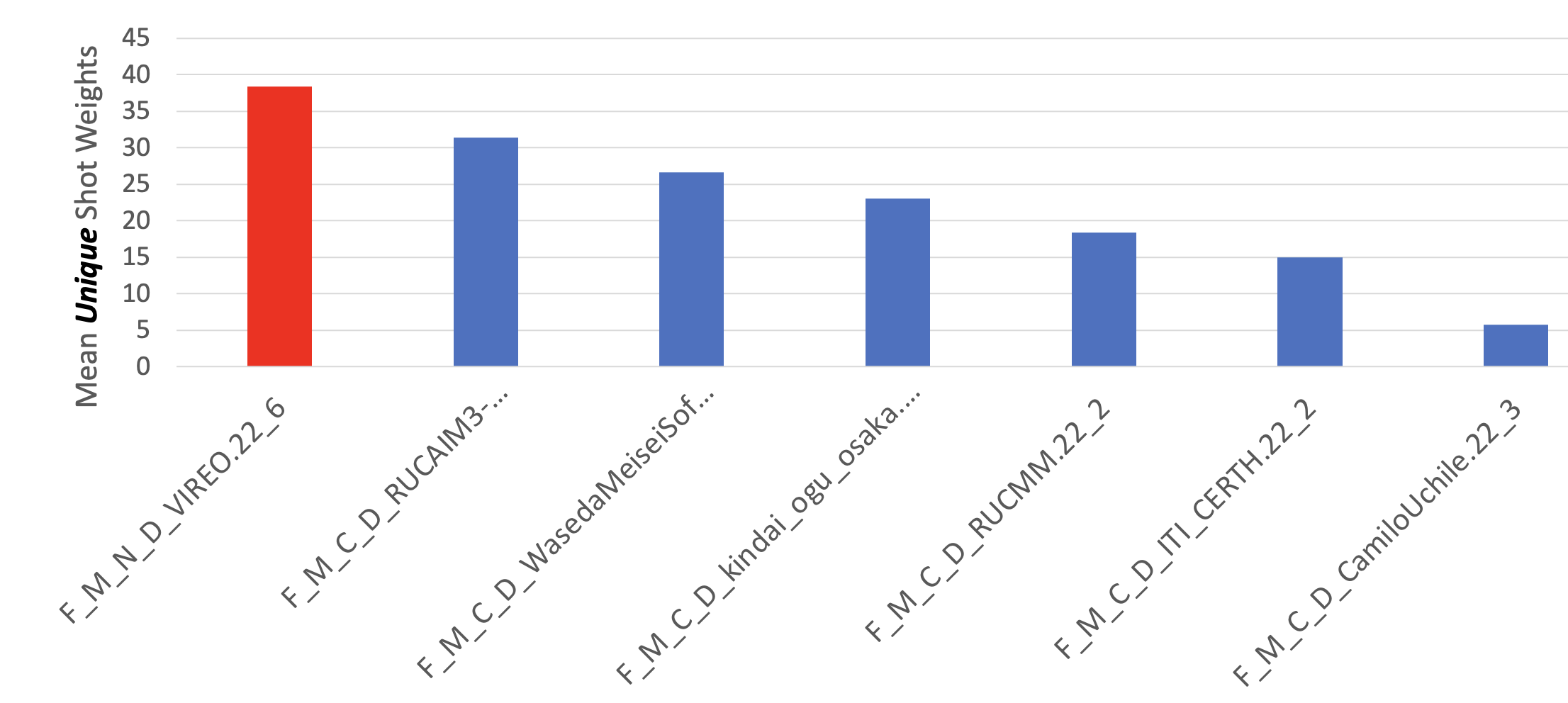}
\caption{AVS: Novelty runs vs best common run from each team}
\label{avs.novelty.scores}
\end{center}
\end{figure}

For Run R and for all topics, we calculate the summation S of all unique shot weights only and the final novelty metric score is the mean score across all evaluated 30 topics. Figure \ref{avs.novelty.scores} shows the novelty metric scores. The red bars indicate the single submitted novelty run. 

We should note here that in running this experiment, for a team that submitted a novelty run, we removed all its other common runs submitted. The reason for doing this was the fact that usually for a given team there would be many overlapping shots within all its submitted runs. For other teams who did not submit novelty runs, we chose the best (top scoring) run for each team for comparison purposes. As shown in the figure, the novelty run (by VIREO team) scored best based on our metric. More runs are needed to conduct a better comparison within novelty systems.

\begin{figure}[htbp]
\begin{center}
\includegraphics[height=2.0in,width=3.0in,angle=0]{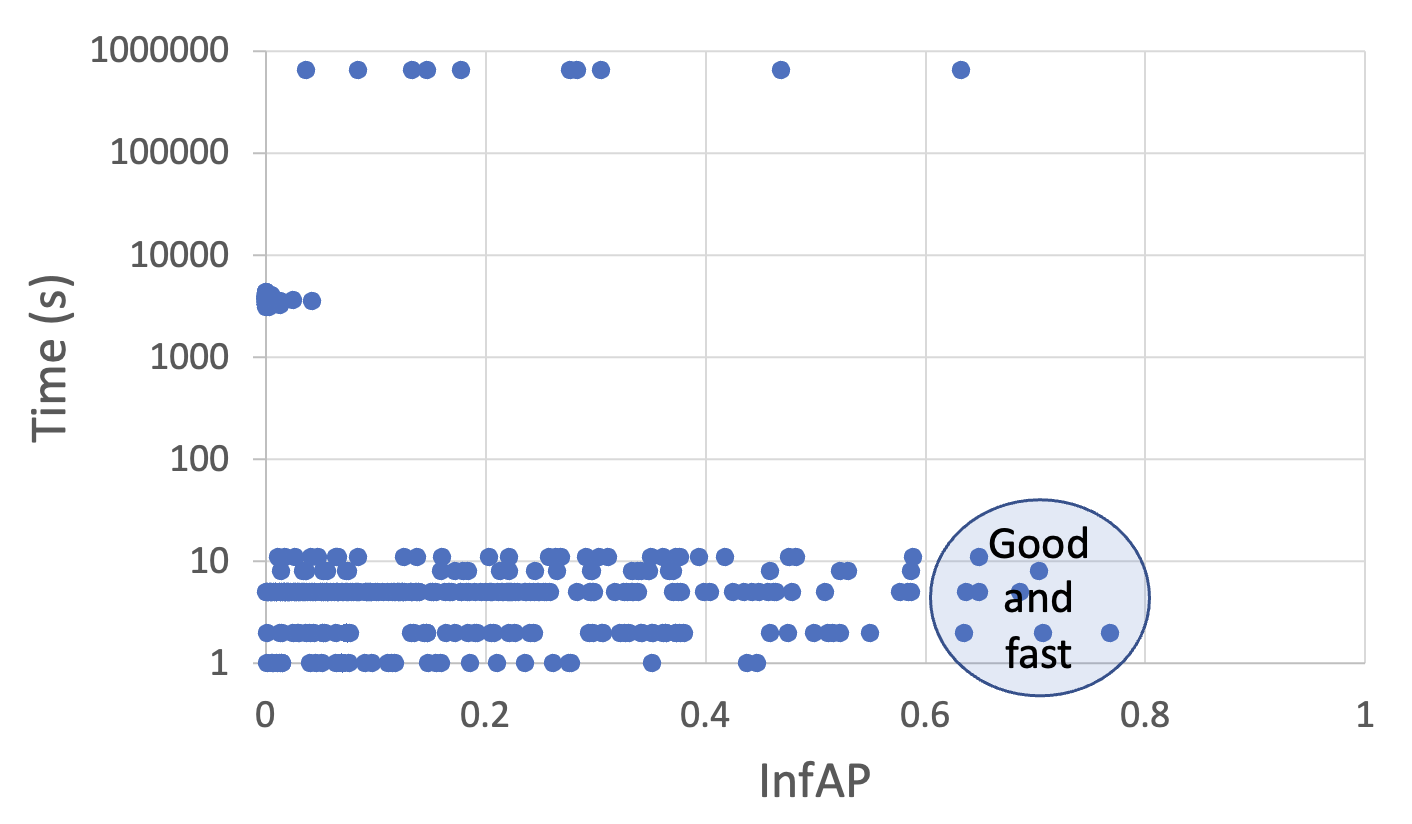}
\caption{AVS: Processing time vs scores (fully automatic)}
\label{avs.f.time.score}
\end{center}
\end{figure}

\begin{figure}[htbp]
\begin{center}
\includegraphics[height=2.0in,width=3.0in,angle=0]{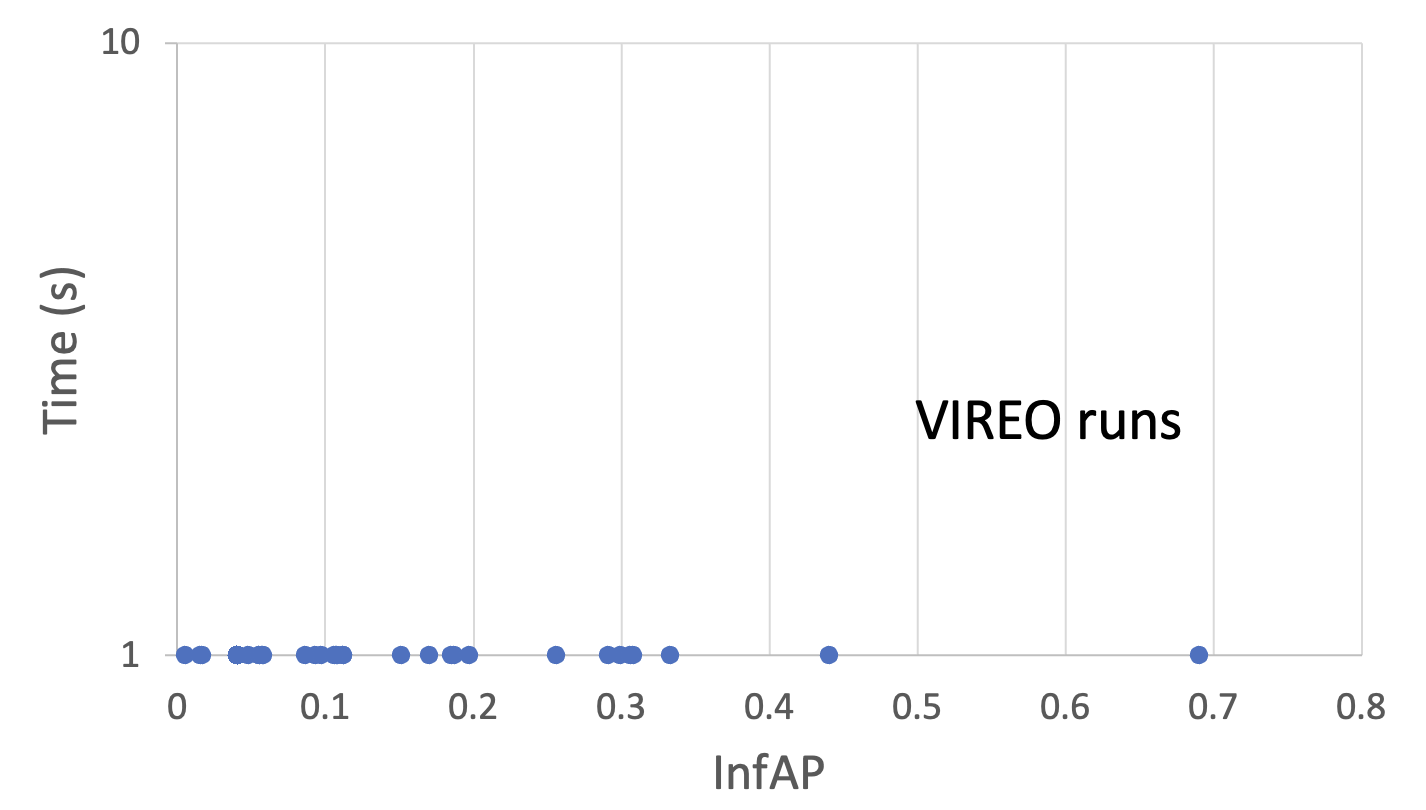}
\caption{AVS: Processing time vs scores (manually assisted)}
\label{avs.m.time.score}
\end{center}
\end{figure}

Among the submission requirements, we asked teams to submit the processing time that was consumed to return the result sets for each query. Figures \ref{avs.f.time.score} and \ref{avs.m.time.score} plot the reported processing times vs the InfAP scores among all run queries for automatic and manually-assisted runs respectively. 

It can be seen that spending more time did not necessarily help in many cases and few queries achieved high scores in less time. There is more work to be done to make systems efficient and effective at the same time. In general, most automatic systems reported processing time below 10 s. Since all manually-assisted runs come from a single team, they all had similar processing time.

To analyze in general which topics were the easiest and most difficult we sorted topics by the number of runs that scored above or below the midpoint score of xInfAP $>$= 0.38 for any given topic and assumed that those runs with 0.38 or above were the easiest topics, while topics with xInfAP $<$ 0.38 were assumed hard topics. 
From this analysis, it can be concluded that the top 5 hard topics were: \say{A kneeling man outdoors}, \say{Two or more persons in a room with a fireplace}, \say{A woman wearing a head kerchief}, \say{A room with blue wall}, and \say{A person wearing a light t-shirt with dark or black writing on it}. On the other hand, the top 5 easiest topics were: \say{A person is biking through a path in a forest}, \say{A construction site}, \say{A person is in the act of swinging}, and \say{A female person bending downwards}, and \say{A type of cloth hanging on a rack, hanger, or line}.

\begin{figure}[htbp]
\begin{center}
\includegraphics[height=1.8in,width=2.8in,angle=0]{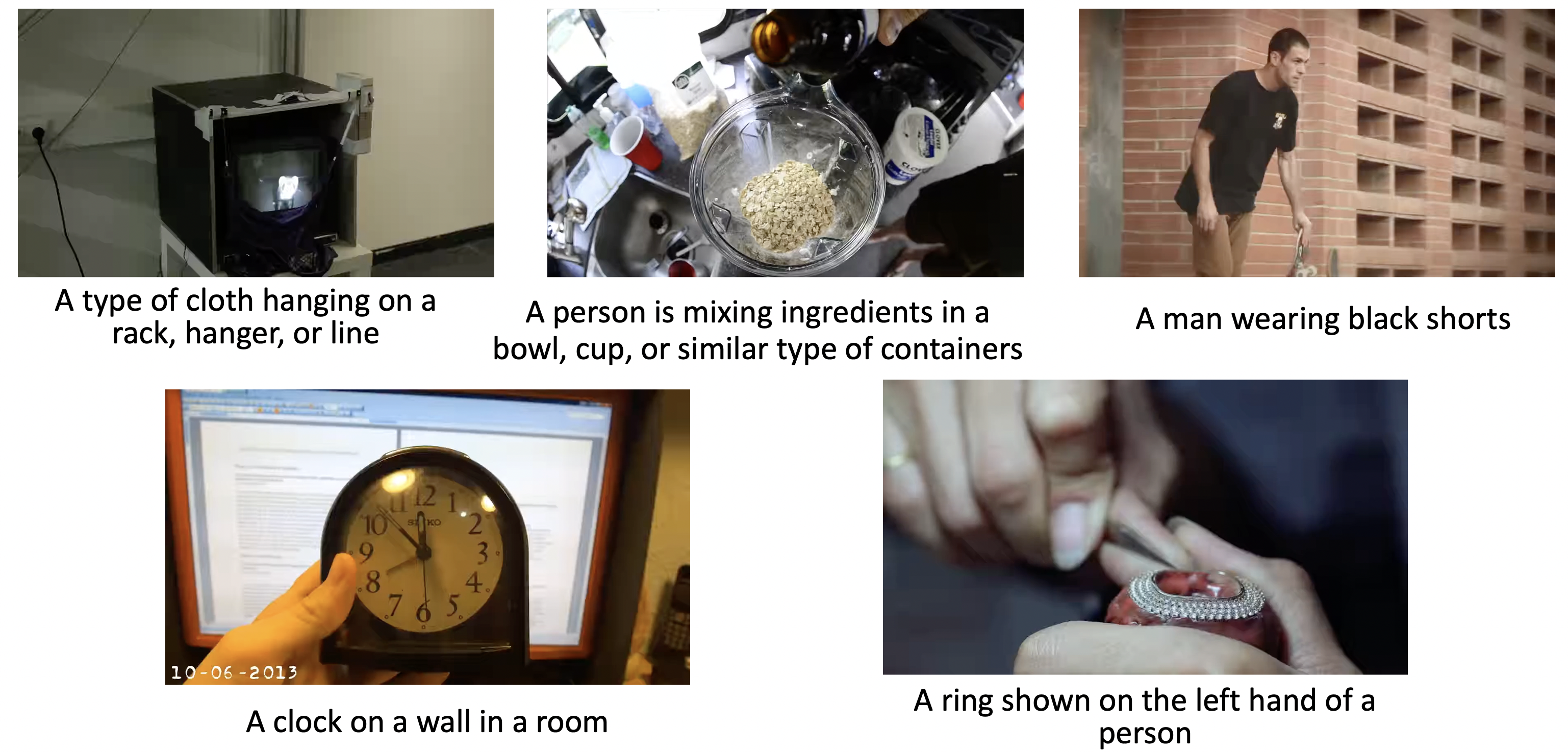}
\caption{AVS: Samples of frequent false positive results}
\label{avs.fp}
\end{center}
\end{figure}

Sample results of frequently submitted false positive shots are demonstrated\footnote{All figures are in the public domain and permissible under RPO \#ITL-17-0025} in Figure \ref{avs.fp}.

\subsubsection{Ad-hoc Observations and Conclusions}

Compared to the semantic indexing task that was conducted to detect single concepts (e.g., airplane, animal, bridge) from 2010 to 2015 it can be seen from running the ad-hoc task the last 6 years that it is still very hard and systems still have a lot of room to research methods that can deal with unpredictable queries composed of one or more concepts including their interactions, relationships and conditions.

From 2016 to 2021 we concluded two cycles of six years running the Ad-hoc task using the Internet Archive (IACC.3) dataset \cite{2016trecvidover} and the Vimeo Creative Commons Collection (V3C1). 

To summarize major observations in 2022 we can see that overall team participation and task completion rates are stable.
Most submitted runs were of training type “D”, and no runs of type “E” or “E” were submitted. One novelty run type was submitted. Overall, 33 systems (28 automatic and 5 manually-assisted) were submitted in the main task including 1 novelty run, while 28 runs were submitted for the progress task. The highest performance this year came lower than the previous two years in general. However, queries are different and meant to search for more fine-grained information. Few automatic systems are good and fast ($<$ 10 sec). There exists a high similarity between automatic and manually-assisted systems in terms of query performance relative to each other. The top scoring teams did not necessarily contribute a lot of unique true shots and vice-versa. About 36\% of all hits are unique, while, 64\% are common hits across the submitted runs. Overall, 13.5\% of all judged shots across all queries are true positives. Hard queries are the ones asked for unusual combinations of facets (compared to well-known concepts commonly found in the available training datasets). For low performance queries, usually all systems are condensed in a small range. While for mid to high performance queries, the top 10 runs vary in their range of performance.

As a general high-level systems overview, we observe the use of multiple text-image and text-video common latent embedding approaches such as VSE++, GSMN, CLIP, and SLIP. In terms of training datasets, multiple text-image and text-video annotated collections have been utilized by teams such as MSR-VTT, TGIF, Flickr8k/30k, MS-COCO, and Conceptual Captions.
Teams adopted multiple visual and textual feature extractors as well as techniques such as triplet loss with margin for embedding space learning. Attention-based methods are popular, while different teams experimented with special methods such as bidirectional Negation Learning (for queries with negative cues) and dual softmax with background queries.
There were no more concept bank approaches as all techniques applied some form of dual task learning (interpretable embeddings).
Finally, it became hard to distinguish between data or feature effects and algorithmic effects.


For detailed information about the approaches and results for individual teams, we refer the reader to the reports \cite{tv22pubs} in the online workshop notebook proceedings.

 \subsection{Deep Video Understanding}
 
Deep video understanding is a challenging task that requires systems to develop a deep analysis and understanding of the relationships between different entities in video, to use known information to reason about other, more hidden information, and to populate a knowledge graph (KG) representation with all acquired information \cite{curtis2020hlvu}. To work on this task, a system should take into consideration all available modalities (speech, image/video, and in some cases text). The aim of this task is to push the limits of multi-modal extraction, fusion, and analysis techniques to address the problem of analyzing long duration videos holistically and extracting useful knowledge to utilize it in solving different types of queries. The target knowledge includes both visual and non-visual elements. As videos and multimedia data are getting more and more popular and usable by users in different domains and contexts, the research, approaches and techniques we aim to be applied in this task will be very relevant in the coming years and near future.

\subsubsection{Dataset}
The Deep Video Understanding Training Set described in Table \ref{trainset} consists of 14 Creative Commons (CC) license movies with a total duration of about 17.5 hours\footnote{https://www-nlpir.nist.gov/projects/trecvid/dvu/\\dvu.development.dataset/}. This training set has been annotated by human assessors and final ground truth, both at the overall movie level (Ontology of relations, entities, actions \& events, Knowledge Graph, and names and images of all main characters), and the individual scene level (Ontology of locations, people/entities, interactions and their order between people, sentiments, and text summary) has been be provided to participating researchers for training and development of their systems. In summary, we hired 5 annotators in addition to a summer student and on average each movie took about 20 hours of work to annotate both movie and scenes. A sample from a scene-level knowledge graph annotation can be seen in Figure \ref{scene.kg}. For more detailed information about the annotation framework please refer to our paper at \cite{erika2022}.

The DVU Test Set described in Table \ref{testset} contains 6 movies licensed from KinoLorberEdu\footnote{\url{https://www.kinolorber.com/}} platform with a total duration of about 8.5 hours. Participants were required to complete a data access form in order to access these movies. The testing set was fully annotated by human annotators to the same degree as the training set. A set of queries, described in more detail in Section 3, were then automatically extracted from human annotations and released to participants, along with the set of movies and annotated images of the movie characters identified during annotation.

Further information about movies' genres and duration are provided below in Tables \ref{trainset} and \ref{testset}.

\begin{center}
\begin{table}
\centering{
 \begin{tabular}{|p{2.6cm} | p{2.7cm} | p{1.5cm}|} 
 \hline
 \textbf{Movie} & \textbf{Genre} & \textbf{Duration}\\ [0.5ex] 
 \hline\hline
 Honey & Romance&86\thinspace min \\
 \hline
 Let's Bring &&\\Back Sophie & Drama& 50\thinspace min\\
 \hline
 Nuclear Family & Drama & 28\thinspace min \\
 \hline
 Shooters & Drama & 41\thinspace min\\
 \hline
 Spiritual Contact - The Movie & Fantasy& 66\thinspace min\\
 \hline
 Super Hero & Fantasy & 18\thinspace min \\
 \hline
 The Adventures of Huckleberry Finn & Adventure & 106\thinspace min \\  
 \hline
 The Big Something & Comedy & 101\thinspace min\\
 \hline
 Time Expired & Comedy / Drama & 92\thinspace min \\
 \hline
 Valkaama & Adventure & 93\thinspace min \\ 
 \hline
 Bagman & Drama / Thriller & 107\thinspace min \\
 \hline
 Manos & Horror & 73\thinspace min \\
 \hline
 Road to Bali & Comedy / Musical & 90\thinspace min \\
 \hline
 The Illusionist & Adventure / Drama & 109\thinspace min \\
 \hline
\end{tabular}
\caption{The full DVU training set}
\label{trainset}
}
\end{table}
\end{center}

\begin{center}
\begin{table}
\centering{
 \begin{tabular}{|p{2.6cm} | p{2.7cm} | p{1.5cm}|} 
 \hline
 \textbf{Movie} & \textbf{Genre} & \textbf{Duration}\\ [0.5ex] 
 \hline\hline
 Calloused Hands & Drama & 92\thinspace min\\
 \hline
 Chained For Life & Comedy / Drama & 88\thinspace min \\
 \hline
 Liberty Kid & Drama & 88\thinspace min \\
 \hline
 Like Me & Horror / Thriller & 79\thinspace min\\
 \hline
 Littlerock & Drama & 82\thinspace min \\
 \hline
 Losing Ground & Comedy / Drama & 81\thinspace min\\
 \hline
\end{tabular}
\caption{The full DVU testing set}
\label{testset}
}
\end{table}
\end{center}

\subsubsection{Annotation}
\begin{figure}[!htbp]
\begin{center}
\includegraphics[height=2.8in,width=2.8in]{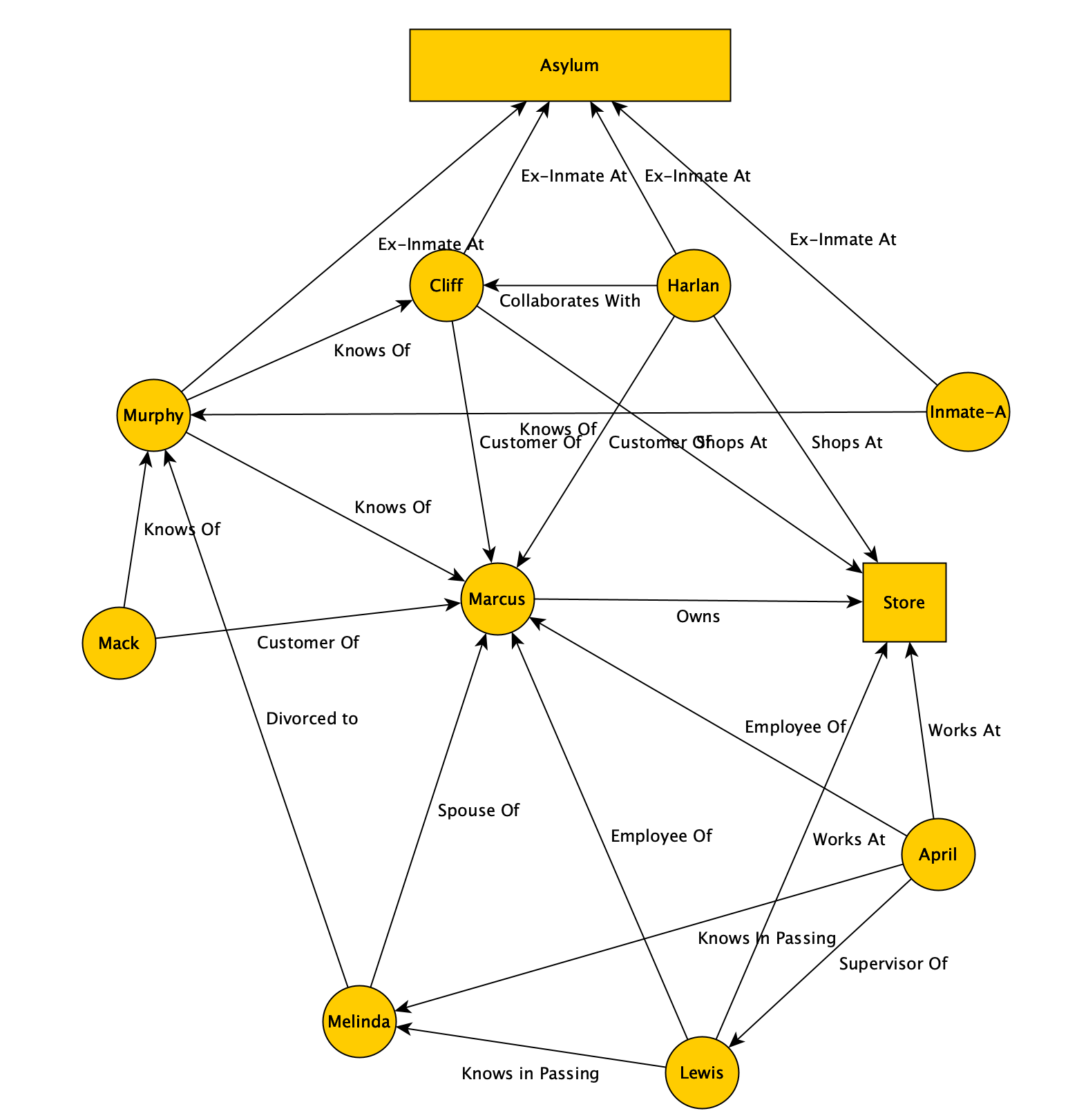}
\caption{Movie-level KG sample}
\label{movie.kg}
\end{center}
\end{figure}

\begin{figure}[!htbp]
\begin{center}
\includegraphics[height=1.8in,width=2.8in]{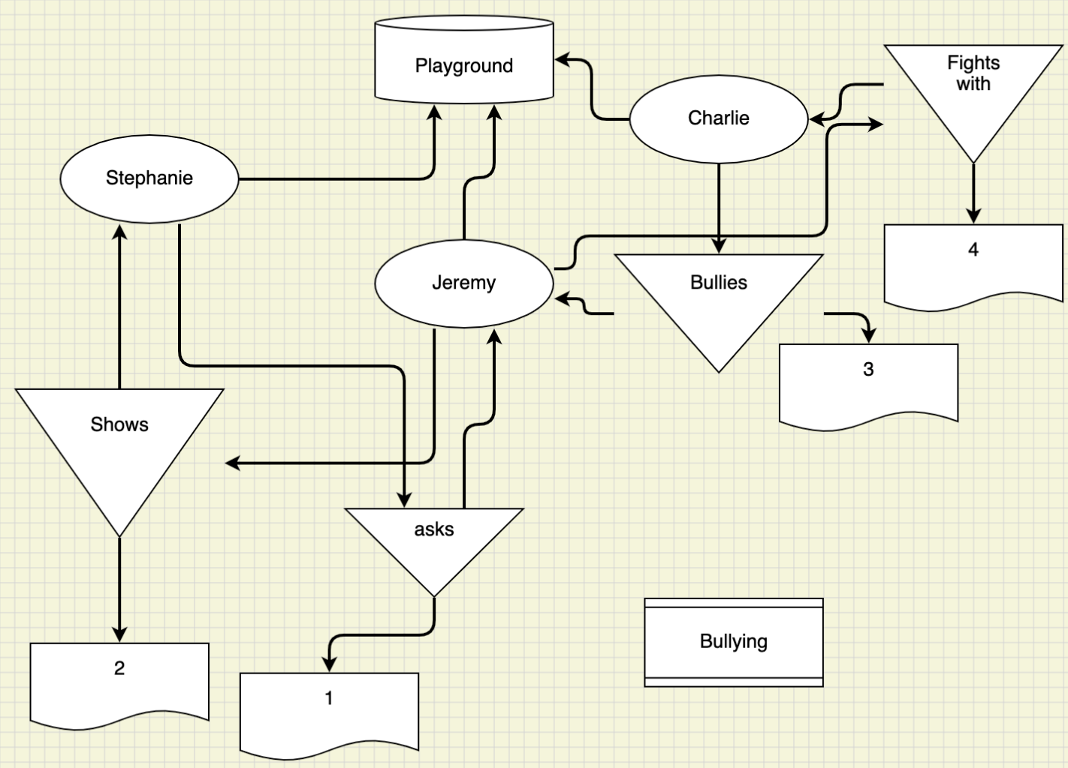}
\caption{Scene-level KG sample}
\label{scene.kg}
\end{center}
\end{figure}

Human assessors annotated each movie of the full DVU dataset. Full movies were annotated to a Knowledge Graph (KG) indicating the relationships and connections between every major character, entity, and concept in the movie. Images of each character and entity were also provided. Figure \ref{movie.kg} shows an example of a movie-level KG. Following this, every scene within each movie was also annotated to a scene-level KG indicating the locations and characters within each scene, at least one sentiment label for that scene, non-neutral mental states of characters, the interactions between characters, and the ordering of the interactions as they happened in that scene. Figure \ref{scene.kg} shows an example of a scene-level KG.

\subsubsection{System task}

The Deep Video Understanding task was as follows: given a whole original \textbf{movie} (e.g. 1.5 - 2hrs long), \textbf{image snapshots} of main entities (persons, locations, and concepts) per movie, and \textbf{ontology} of relationships, interactions, locations, and sentiments used to annotate each movie at global movie-level (relationships between entities) as well as on fine-grained scene-level (scene sentiment, interactions between characters, and locations of scenes), systems were expected to generate a knowledge-base of the main actors and their relations (such as family, work, social, etc) over the whole movie, and of interactions between them over the scene level. This representation would be used to answer a set of queries on the movie-level and/or scene-level (see below details about query types) per movie. The task supported two tracks (subtasks) where teams could join one or both tracks. The Movie track was comprised of queries on the whole movie level, and the Scene track was comprised of queries targeting specific movie scenes.

\subsubsection{Query Topics \& Metrics}

This task has two query types for each of the two subtasks (Tracks). Details are provided below in addition to the metrics used to score submissions.\\

\textbf{Movie-level Track}
\begin{itemize}
  \item \textbf{Question Answering - Required Query Type}
  \\This query type (mandatory in the movie-level track) represents questions on the resulting knowledge base of the movies in the testing dataset. For example, we may ask `How many children does Person A have?', in which case participating researchers should count the `Parent Of' relationships Person A has in the Knowledge Graph. This query type takes a multiple choice questions format.

  \item \textbf{Fill in the Graph Space - Optional Query Type}
  \\Fill in spaces in the Knowledge Graph (KG). Given the listed relationships, events or actions for certain nodes, where some nodes are replaced by variables X, Y, etc., solve for X, Y etc. Example of The Simpsons: X Married To Marge. X Friend Of Lenny. Y Volunteers at Church. Y Neighbor Of X. Solution for X and Y in that case would be: X = Homer, Y = Ned Flanders. 
\end{itemize}

\textbf{Scene-level Track}
\begin{itemize}
    \item \textbf{Find Next or Previous Interaction - Required Query Type}
    \\Given a specific scene and a specific interaction between person X and person Y, participants are asked to return either the previous interaction or the next interaction, in either direction, between person X and person Y. This can be specifically the next or previous interaction within the same scene, or over the entire movie. This query type takes a multiple choice questions format and it is considered a mandatory query in the scene-level track).

    \item \textbf{Find Unique Scene - Optional Query Type}
    \\Given a full, inclusive list of interactions, unique to a specific scene in the movie, teams should find which scene this is.
\end{itemize}

Queries for this task were generated semi-automatically by parsing full annotations over the movie-level and the scene-level and populating a data structure with the full knowledge base. Four different sets of questions and accompanying answers for each query type were automatically generated. The TRECVID team then checked questions by hand, taking care to eliminate any questions which were duplicates or near-duplicates of previous questions, or where the question was considered not of sufficient quality to effectively evaluate systems performance.\\

\textbf{Metrics}
\begin{itemize}
    \item \textbf{Movie-level : Question Answering}
    \\Scores for this query were produced by calculated by the number of Correct Answers / number of Total Questions.

    \item \textbf{Movie-level : Fill in the Graph Space}
    \\Results were treated as a ranked list of result items per each unknown variable, and the Reciprocal Rank score was calculated per unknown variable and Mean Reciprocal Rank (MRR) per query.

    \item \textbf{Scene-level : Find Next or Previous Interaction}
    \\Scores for this query were produced by calculated by the number of Correct Answers / number of Total Questions.

    \item \textbf{Scene-level : Find Unique Scene}
    \\Results were treated as a ranked list of result items per each unknown variable, and the Reciprocal Rank score  was calculated per unknown variable and Mean Reciprocal Rank (MRR) per query.
\end{itemize}

\subsubsection{Evaluation}

The advantage of automatically generating questions for this task was that evaluations could be performed automatically. A system was developed to parse correct answers for each query, as well as submitted answers from each participant team's submission. Answers were then compared and an itemized output was generated allowing participating teams to see the correct answers for each query in addition to their submitted query and assigned scores.

Itemized results per query were provided to each team in addition to a summarized table of final ranked results for each submission listing Total scores, Average scores, and Percentages. Further summarized tables were provided for each submission listing their Total scores, Average scores, and Percentages for each movie of the testing set.

\subsubsection{Results}
\begin{figure}[htbp]
\begin{center}
\includegraphics[height=2.5in,width=3in,angle=0]{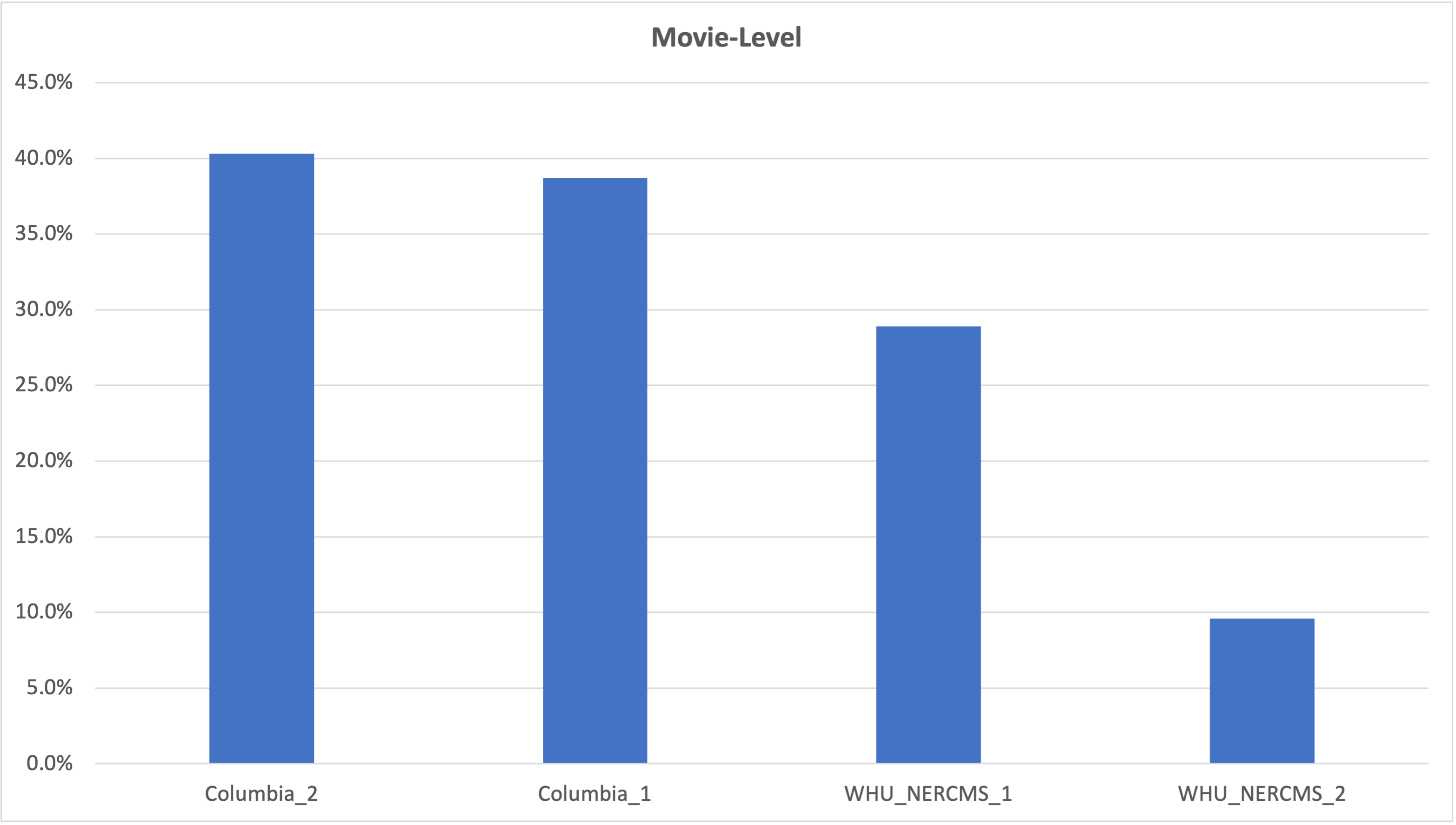}
\caption{DVU: Overall summary scores for movie-level}
\label{dvu.movie.summary}
\end{center}
\end{figure}

\begin{figure}[htbp]
\begin{center}
\includegraphics[height=2.5in,width=3in,angle=0]{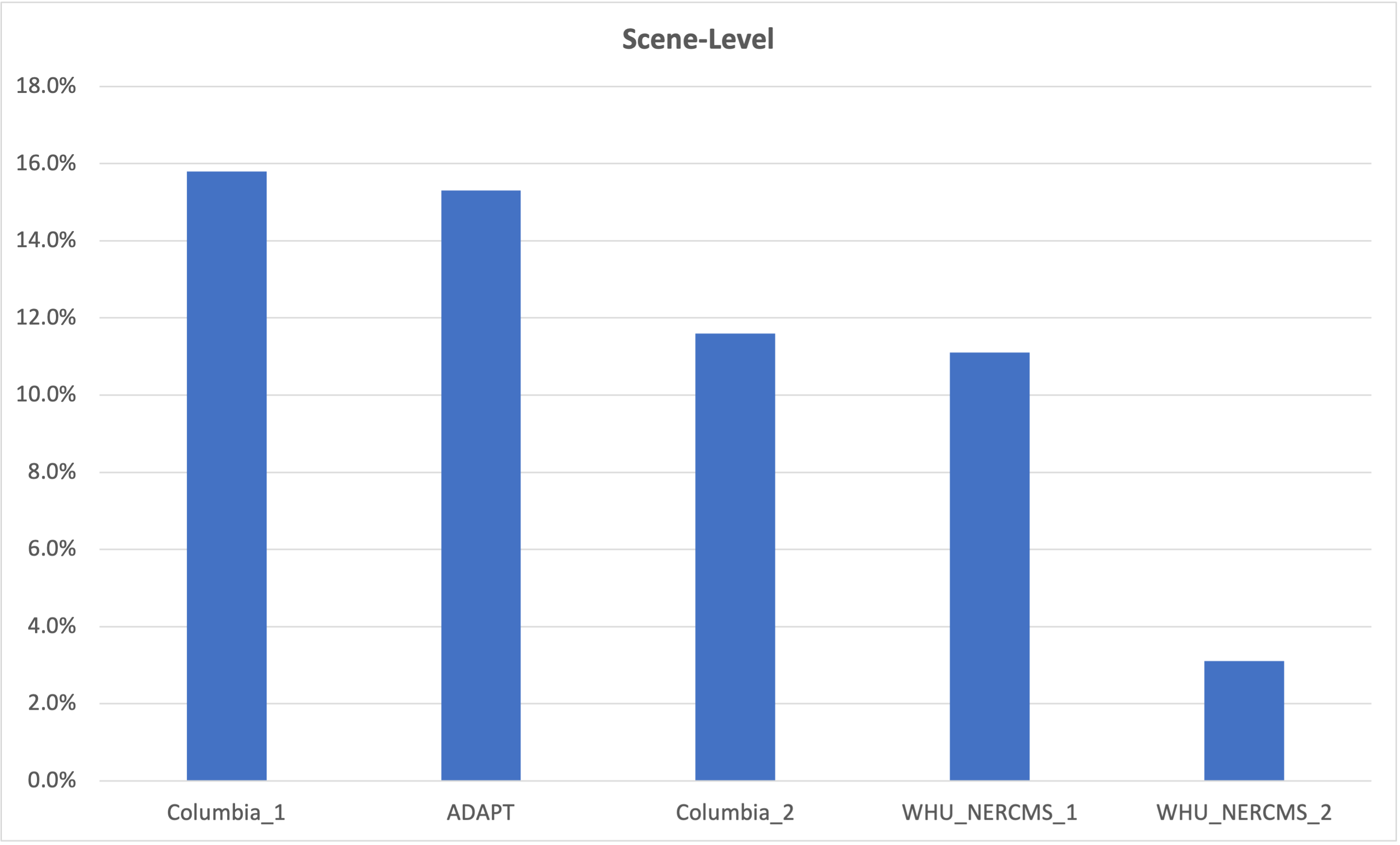}
\caption{DVU: Overall summary scores for scene-level}
\label{dvu.scene.summary}
\end{center}
\end{figure}

Figures \ref{dvu.movie.summary} and \ref{dvu.scene.summary} show the overall summary scores of runs for all teams participating in these tasks. The Columbia team achieved the best results for both movie and scene level queries, with the ADAPT team runner-up for best scores on scene-level queries.

\begin{figure}[htbp]
\begin{center}
\includegraphics[height=2.5in,width=3in,angle=0]{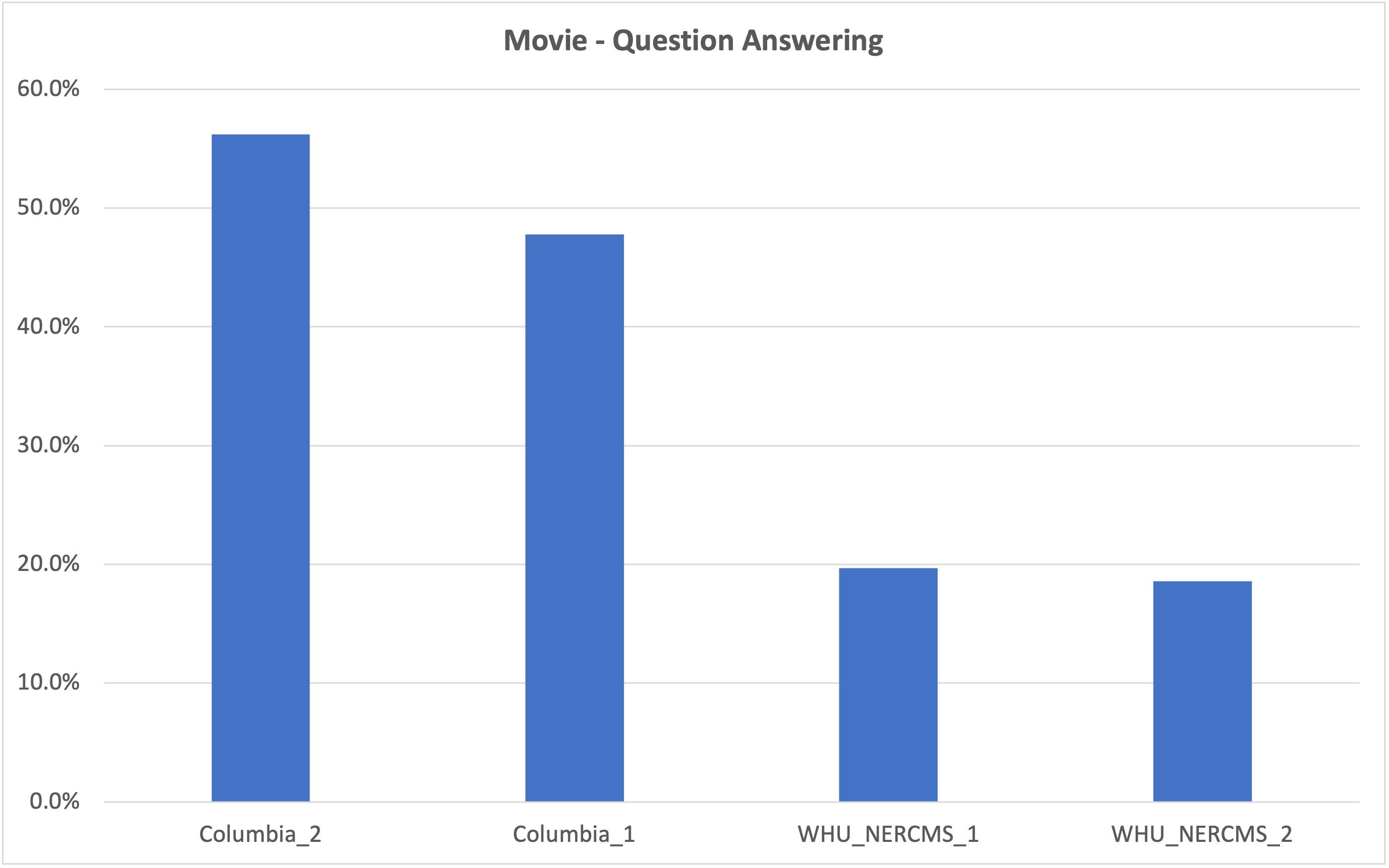}
\caption{DVU: Overall question answering scores for movie-level}
\label{dvu.movie.qa}
\end{center}
\end{figure}

Figure \ref{dvu.movie.qa} shows the overall scores for the Question Answering query over the movie-level. The Columbia team outperformed the other team in the movie queries task with a gap to the other team participating in the movie queries task, WHU\_NERCMS.

\begin{figure}
\begin{center}
\includegraphics[height=2.5in,width=3in,angle=0]{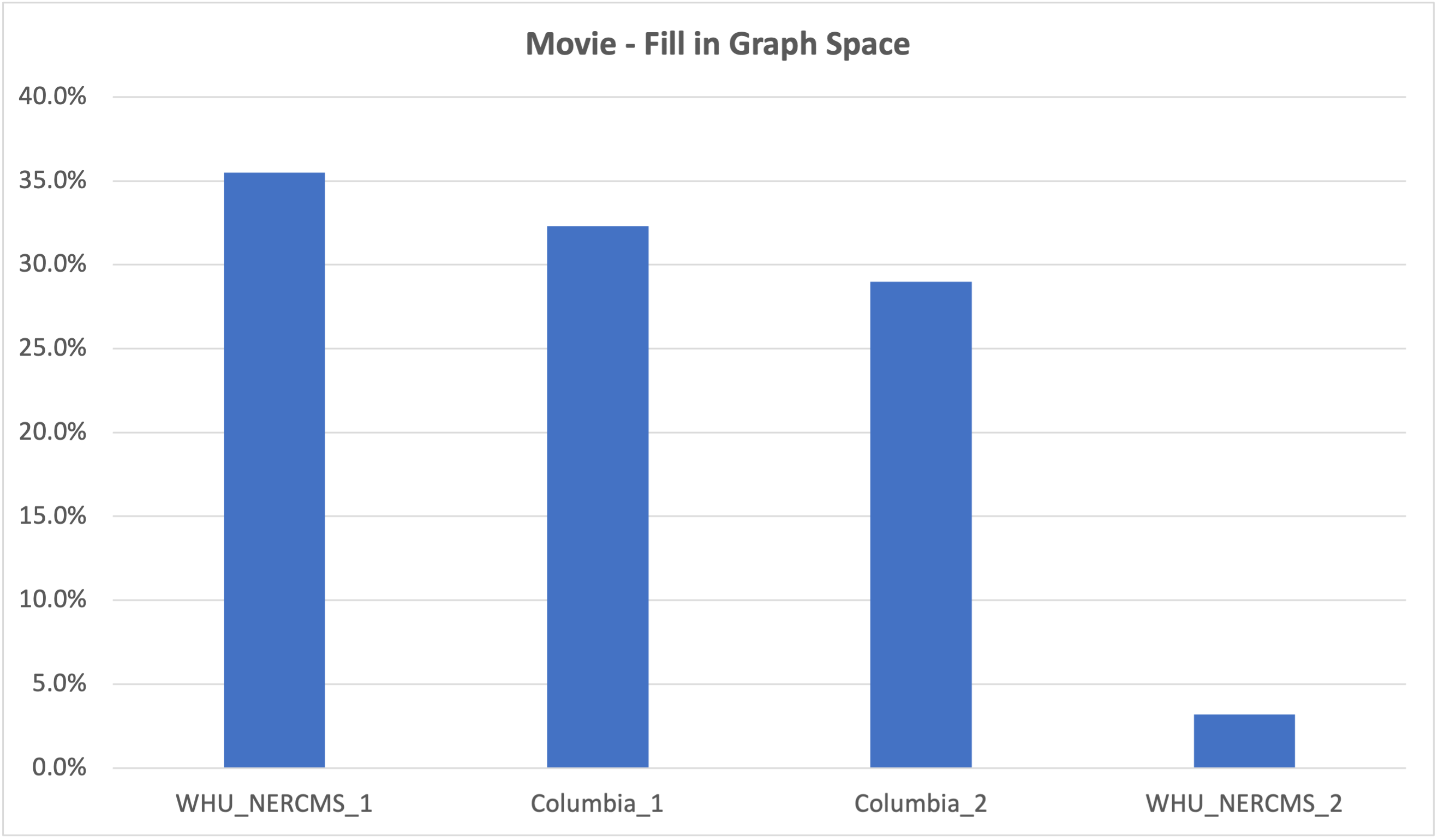}
\caption{DVU: Overall fill in graph space scores for movie-level}
\label{dvu.movie.fill}
\end{center}
\end{figure}

Figure \ref{dvu.movie.fill} shows the overall scores for the Fill in The Graph Space query for the movie-level. WHU\_NERCMS run 1 achieves the best results over Columbia runs 1 and 2. This is then followed by WHU\_NERCMS run 2.

\begin{figure}
\begin{center}
\includegraphics[height=2.5in,width=3in,angle=0]{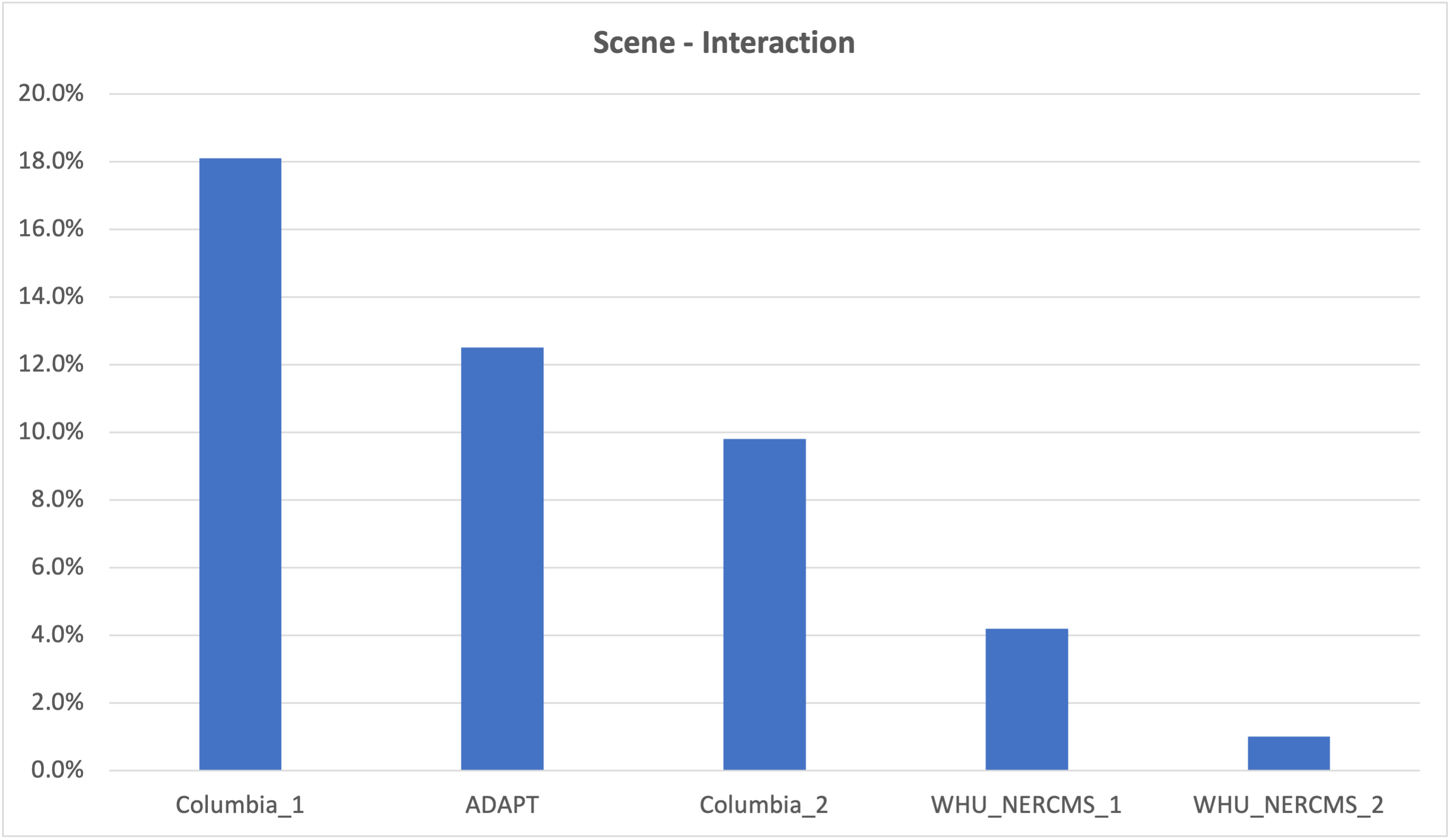}
\caption{DVU: Overall find next / previous interaction results for scene-level}
\label{dvu.scene.inter}
\end{center}
\end{figure}

Figure \ref{dvu.scene.inter} shows the overall scores for Find the Next / Previous Interaction queries over the scene-level. Columbia run 1 achieves the best results, followed by team ADAPT. These are followed by Columbia run 2 and WHU\_NERCMS runs 1 and 2.

\begin{figure}
\begin{center}
\includegraphics[height=2.5in,width=3in,angle=0]{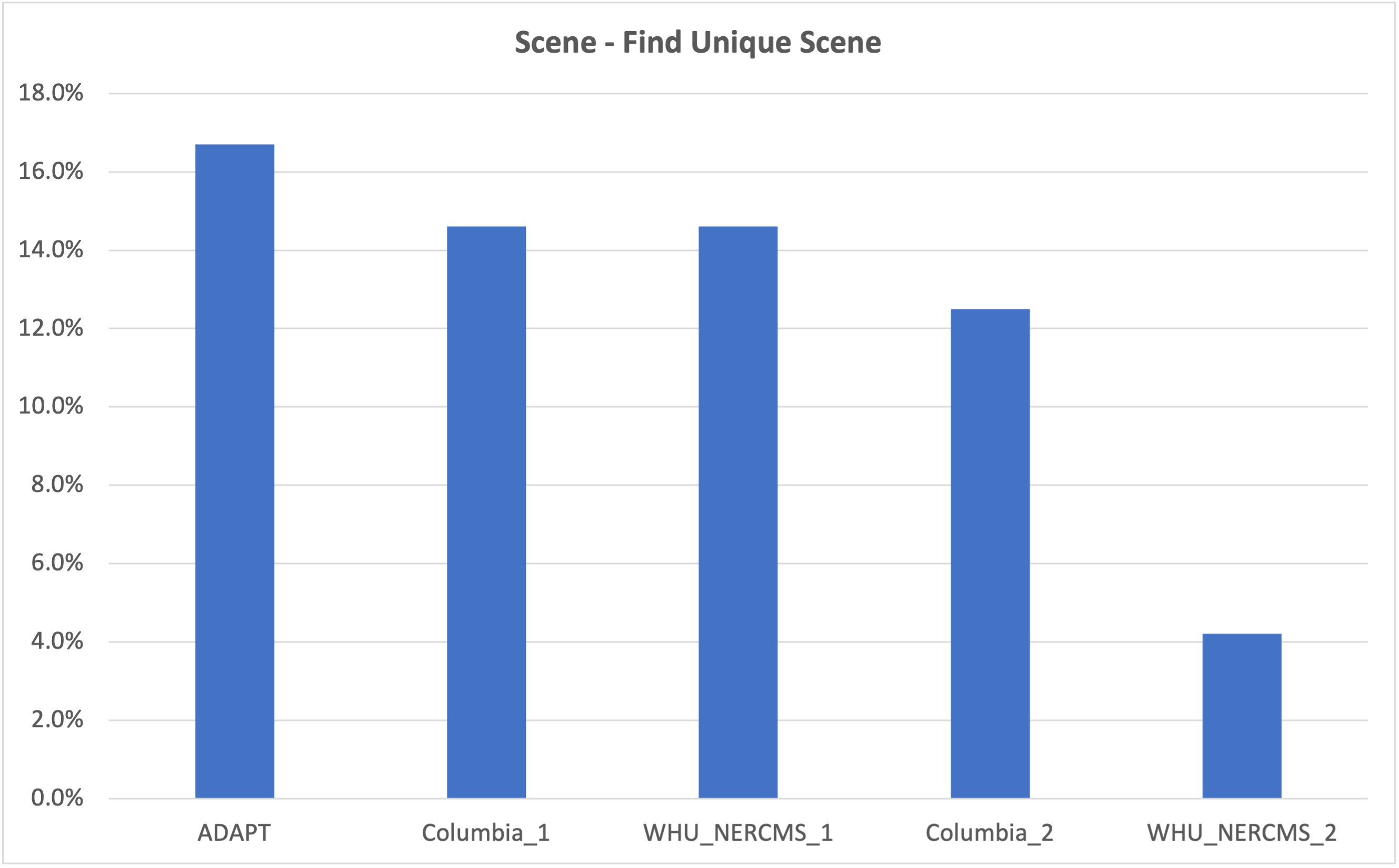}
\caption{DVU: Overall find unique scene results for scene-level}
\label{dvu.scene.findunique}
\end{center}
\end{figure}

Figure \ref{dvu.scene.findunique} shows overall scores for Find the Unique Scene queries for the scene-level. Team ADAPT achieves the best results for this query, followed by Columbia run 1 and WHU\_NERCMS run 1 who achieved equal scores. Following these are Columbia run 2 and WHU\_NERCMS run 2.

\begin{figure*}
\begin{center}
\includegraphics[height=3.5in,width=6.5in,angle=0]{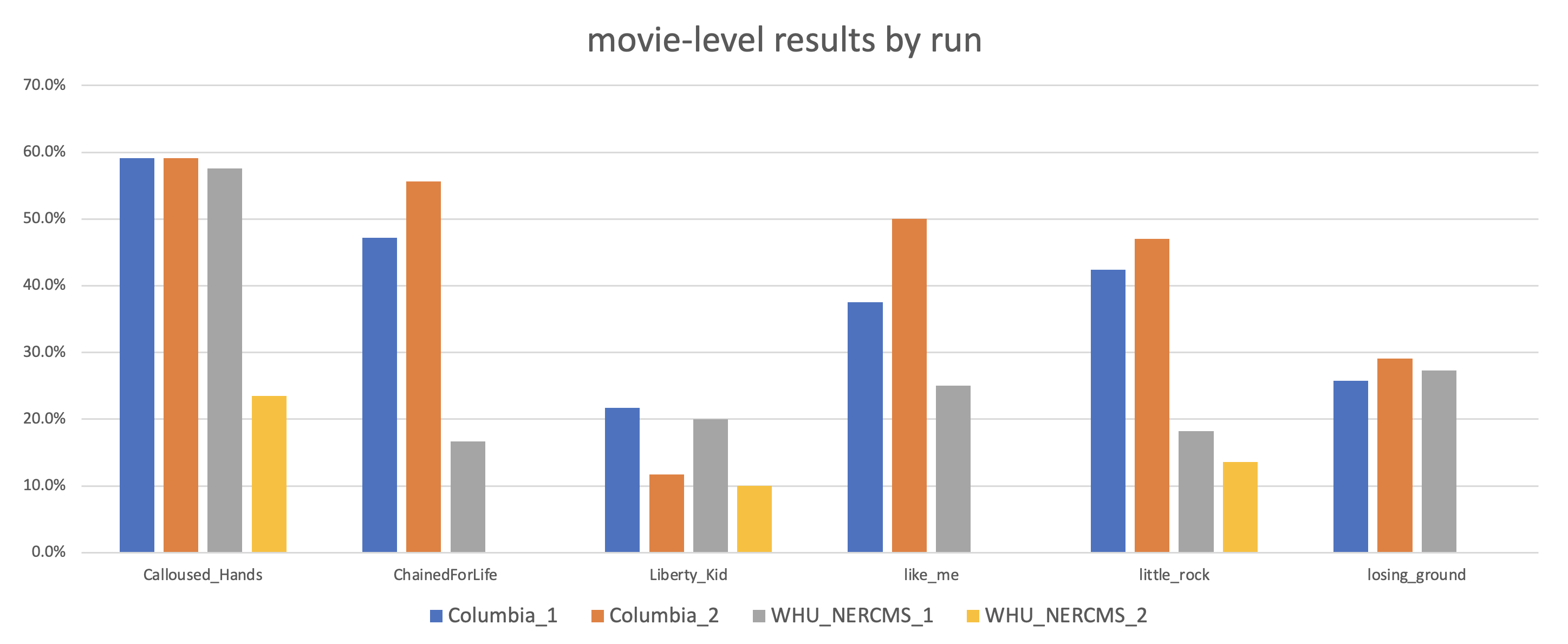}
\caption{DVU: Movie-level results per run}
\label{dvu.movie.results}
\end{center}
\end{figure*}

\begin{figure*}
\begin{center}
\includegraphics[height=3.5in,width=6.5in,angle=0]{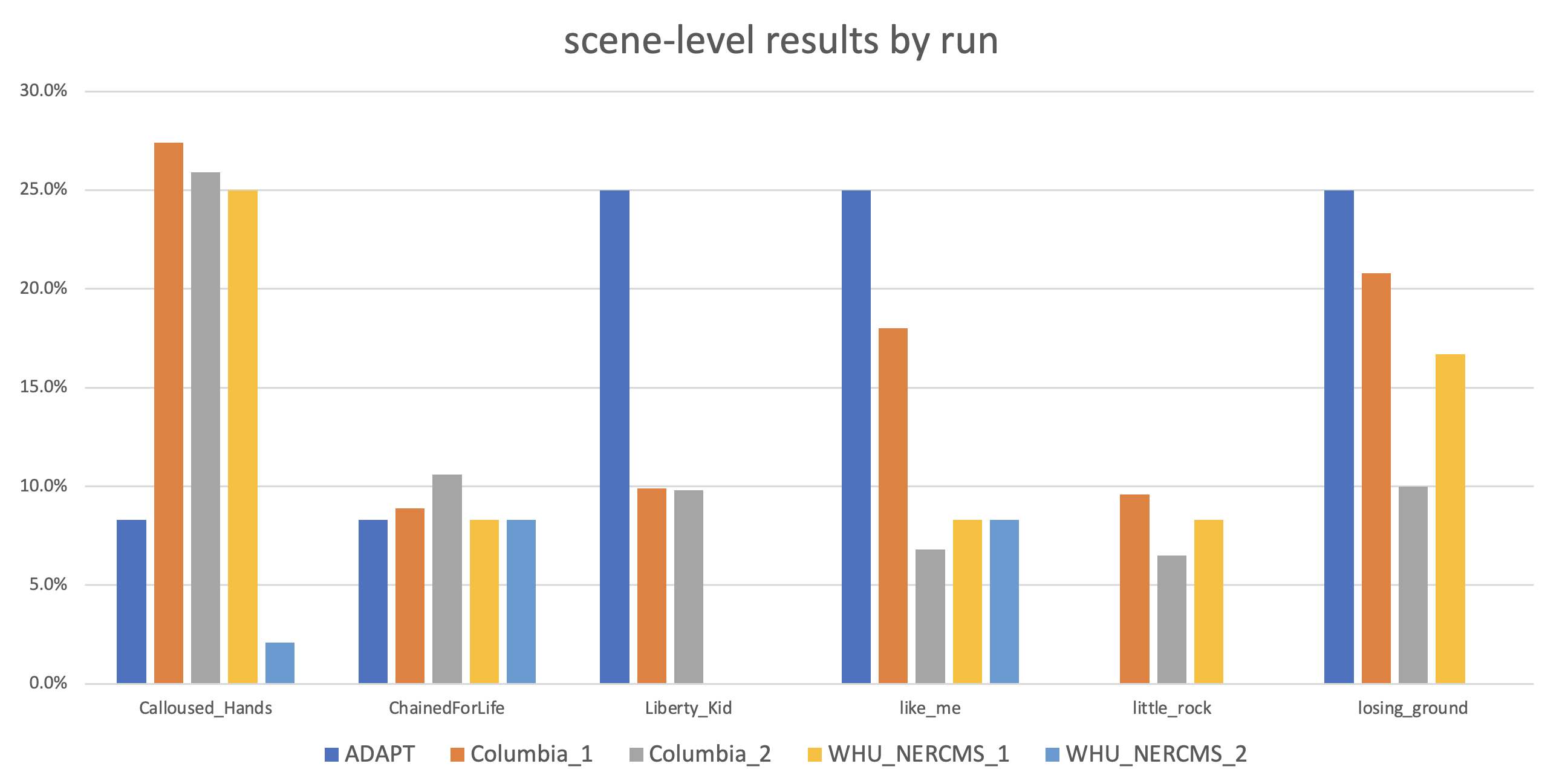}
\caption{DVU: Scene-level results per run}
\label{dvu.scene.results}
\end{center}
\end{figure*}

Figures \ref{dvu.movie.results} and \ref{dvu.scene.results} show results for each run per movie at the movie-level and the scene-level respectively. This shows how team Columbia regularly outperforms other teams for movie-level queries. Team ADAPT performs well for scene-level queries, achieving top results for three movies. The Columbia team performs best for the other three movies in scene-level queries.

\subsubsection{Observations}
This was the first year that Deep Video Understanding was run as a TRECVID task, having been run for the previous three years as an external Grand Challenge. In previous years a set of CC licensed movies were used to run the challenge, while this year a set of 6 movies licensed movies were used for the test set. There were 13 teams who registered for the task, however only three of these submitted runs. We would like to see an increase in this number.

Two of the three finishing teams submitted two runs each for both movie-level queries and scene-level queries, whereas the other finishing team submitted one run for scene-level queries only. Run 2 from Columbia team achieved the best results for scene-level queries, closely followed by team ADAPT. Both runs from Columbia team achieved the two highest results for movie-level queries.

We now summarize the approach taken by teams. ADAPT used a feature extraction algorithm EfficientNetV2 \cite{tan2021efficientnetv2} to generate text descriptions for each video segment. Following this they used PyTorch X3D models for detection of actions by characters. The last step for their scene-level solutions was to search for the coincidences or synonyms using the Gensim library methods \cite{rehurek2011gensim}. Closest synonyms were selected then for the scene-level submissions.

For User-interaction Mapping, Columbia team used a novel approach to multi-entity tracking from their earlier work with face embeddings computed from five facial land-mark points. To find the best approach they experimented with multi-entity tracking only, merge multi-entity tracking and face recognition with the same priority, merge multi-entity tracking and face recognition with multi-entity tracking prioritized, and merge multi-entity tracking and face recognition with face recognition prioritized. For segmenting user stories, they developed a multi-entity-frame method for key-frame extraction. Face recognition and head tracking were equally merged for visual analysis. Locations were then predicted for each prompt by using SIFT features. Their experimental analysis showed that this approach improved the performance. 

WHU\_NERCMS adopted a four-stage method. The first of which was the video structuralization module, which included auto speech recognition (ASR) using YouTube API and clip segmentation. The second part was the instance search module, which included person recognition and track using SCRFD \cite{chen2018scrfd}, ArcFace \cite{deng2019arcface}, faster RCNN \cite{ren2015faster} and Deepsort \cite{wojke2017simple}, as well as location recognition using ResNet. The third module was the interaction and relation recognition module for recognizing the interactions between people and location at the scene-level. This was also used for recognition of the relationships between characters at the movie-level. The fourth module was the knowledge graph module, which adapted the results from the interaction and relation recognition module to generate knowledge graphs for each movie to answer the queries.

\subsubsection{Conclusions}
This was the first year of the new Deep Video Understanding task. Prior to this, Deep Video Understanding has been organized as an external Grand Challenge at ACM Multimedia for three years running. The Grand Challenge version of this task had been using a more complete set of questions for evaluation of systems. The data-set used for the Grand Challenge versions of this task has grown year over year with more movies been added with each iteration. This started with CC licensed movies, however 6 specially licensed movies were used for the test set for this year's Grand Challenge and this TRECVID task. There were 13 teams who registered for the task, however only three of them submitted runs. A greater number of people participated in this year's Grand Challenge, and we would wish to encourage more Grand Challenge participants to also take part in the TRECVID task.

Of the three finishing teams, two submitted two runs each for both movie-level queries and scene-level queries, whereas the other finishing team submitted one run for scene-level queries only. Run 2 from Columbia team achieved the best results for scene-level queries, closely followed by ADAPT team. Both runs from Columbia team achieved the two highest results for movie-level queries.

Submissions for movie-level queries scored much higher than for scene-level queries, indicating that the movie-level queries were easier. Movie-level `Question Answering' queries scored higher than `Fill in the Graph Space' queries, also indicating that these were slightly easier. Scene-level `Find Next or Previous Interaction' queries scored very similar results to `Find the Unique Scene' queries. Queries for the movie \textit{Calloused Hands} scored higher than any other movie. \textit{Liberty Kid} was among the lowest scoring.

\subsection{Disaster Scene Description and Indexing}
\begin{table*} \begin{center}

\begin{tabular}[htbp]{|c|c|c|c|c|}
\hline
Damage & Environment & Infrastructure & Vehicles & Water\\ \hline\hline
Misc. Damage & Dirt & Bridge & Aircraft & Flooding\\ \hline
Flooding/Water Damage& Grass & Building & Boat & Lake/Pond\\ \hline
Landslide& Lava & Dam/Levee & Car & Ocean\\ \hline
Road Washout& Rocks & Pipes & Truck & Puddle\\ \hline
Rubble/Debris& Sand & Utility or Power Lines/Electric Towers & & River/Stream\\ \hline
Smoke/Fire& Shrubs & Railway & &\\ \hline
& Snow/Ice & Wireless/Radio Communication Towers & &\\ \hline
& Trees & Water Tower & &\\ \hline
& & Road & &\\ \hline
\end{tabular}
\caption{DSDI: The test dataset has 5 coarse categories, each divided into 4-9 more specific labels.}
\label{tab:dsdi.categories}
\end{center}
\end{table*}

Computer vision capabilities have rapidly been advancing and are expected to become an important component for incident and disaster response. Having prior knowledge about affected areas can be very helpful for the first responders. Communication systems often go down in major disasters, which makes it very difficult to get any information regarding the damage. Automated systems, such as robots or low flying drones, can therefore be used to gather information before rescue workers enter the area.

With the popularity of deep learning, computer vision research groups have access to very large image and video datasets for various tasks and the performances of systems have dramatically improved. However, the majority of computer vision capabilities are not meeting public safety’s needs, such as support for search and rescue, due to the lack of appropriate training data and requirements. Most current datasets do not have public safety hazard labels due to which state-of-the-art systems trained on these datasets fail to provide helpful labels in disaster scenes.

In response, the New Jersey Office of Homeland Security and MIT Lincoln Laboratory developed a dataset of images collected by the Civil Air Patrol of various natural disasters. The Low Altitude Disaster Imagery (LADI) dataset was developed as part of a larger NIST Public Safety Innovator Accelerator Program (PSIAP) grant. Two key properties of the dataset are as follows:

\begin{enumerate}
    \item Low altitude
    \item Oblique perspective of the imagery and disaster-related features.
\end{enumerate}

These are rarely featured in computer vision benchmarks and datasets. The LADI dataset acted as a starting point to help label a new video dataset with disaster-related features. The image dataset could be used for the training and development of systems for the DSDI task.

DSDI task was introduced in TRECVID in 2020, and this is the third iteration of the task.

\subsubsection{Datasets}
\paragraph {\textbf{Training Dataset}}
The training dataset is based on the LADI dataset hosted as part of the AWS Public Dataset program along with the DSDI video test dataset used in 2021. 

The LADI dataset consists of 20\,000+ human annotated images and about 500\,000 machine annotated images. The images are from locations with FEMA major disaster declarations for a hurricane, earthquake, or flooding\footnote{https://www.fema.gov/disaster/declarations}. The lower altitude criterion distinguishes the LADI dataset from satellite datasets to support the development of computer vision capabilities with small drones operating at low altitudes. A minimum image size (4MB) was selected to maximize the efficiency of the crowdsource workers, since lower resolution images are harder to annotate. 

The ground truth for the DSDI test set for 2021 was made public after completion of the task and is available to be used as training dataset. It consisted of about 6.7 hours of video that were segmented into small video clips (or shots) of a maximum duration of 20.85 seconds. The videos were from earthquakes, flooding, fire, and erosion affected areas. They have been collected from both domestic and international sources. There are a total of 2801 shots with a median length of 8.34 seconds, location metadata which included the start and end coordinates, and the path of the aircraft.

\paragraph {\textbf{Test Dataset}}
The test dataset for the task this year consists of about 5.98 hours of video. 
The dataset raw data was mostly collected by Federal Emergency Management Agency (FEMA) as jpeg images by flying over the affected areas after the natural disaster events. A pre-processing step was conducted to select a sufficiently diverse set of images across a mix of different events, then the sequential jpeg images were stitched to generate the videos. Finally, video speed was subjectively set to balance the total number of clips human annotators can finish within their time while giving them better experience creating the ground truth (e.g., avoiding too fast or too slow videos). In total, a set of 2157 shots with a maximum duration of 16.7 seconds and mean of 10 seconds were generated and distributed to teams.

\begin{figure*}[htbp]
    \centering
    \includegraphics[width=1.0 \linewidth]{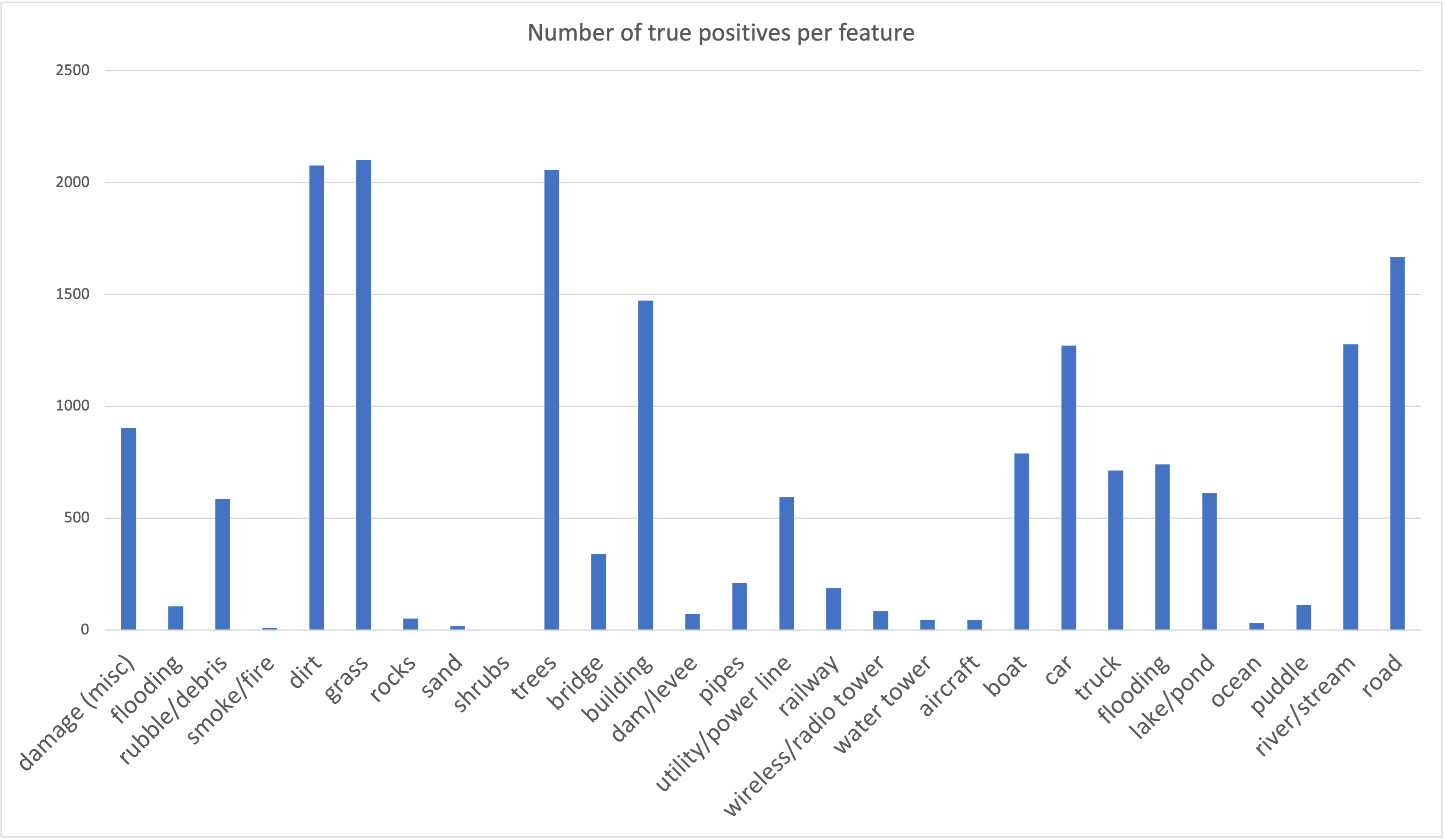}
    \caption{DSDI: Number of shots containing each feature}
    \label{fig:dsdi.features.positive}
\end{figure*}

\paragraph{\textbf{Categories}}
The categories used for the test dataset are the same as those used for the LADI training dataset~\cite{jliu2019ladi}. Five coarse categories were selected based on their importance for the task, and each of these categories is divided into 4-9 more specific labels. The hierarchical labeling scheme is shown in Table~\ref{tab:dsdi.categories}.

As can be expected from a real-world dataset, features appear with varied frequency within the videos. Some features such as grass, trees, buildings, roads, etc. appear much more frequently than others. On the other hand, features such as smoke and sand are rare. This year we excluded four features due to their rare positive shots. These were lava, landslide, road washout, and snow/ice. Figure~\ref{fig:dsdi.features.positive} shows the number of shots that contain each feature. 

\paragraph{\textbf{Annotation}}
The video annotation was done using full-time annotators instead of crowdsourcing. It is essential that the annotators become familiar with the task and the labels before they start a category. For this reason, we created a practice page for each category with multiple examples for each label within that category. The annotators were given 2 videos as a test to mark the labels visible to them, and the answers were compared to ours. We also had regular discussions with the annotators to understand their process and clarify any confusion during the labeling of the dataset. 

Two full-time annotators labeled the testing dataset. Both annotators had worked on the task previously in the last two years and were familiar with it. The annotators worked independently on each category by watching each clip and recording if any of the labels in the given category exists anywhere in the clip. To create the final ground truth, for each shot, the union of the labels was used.

\subsubsection{System Task}

Systems were required to return a ranked list of up to 1000 shots for each of the 32 features. Each submitted run specified its training type:
\begin{itemize}
    \item LADI-based (L): The run only used the supplied LADI dataset for development of its system.
    \item Non-LADI (N): The run did not use the LADI dataset, but only trained using other dataset(s).
    \item LADI + Others (O): The run used the LADI dataset in addition to any other dataset(s) for training purposes.
\end{itemize}

\subsubsection{Evaluation and Metrics}
The evaluation metric used for the task is mean average precision (MAP). The average precision is calculated for each feature, and the mean average precision is reported for each submission. Furthermore, the true positive, true negative, false positive, and false negative rates are also reported. Teams self reported the clock time per inference to compare the speeds of the various systems. 

\subsubsection{Results}
This year 8 teams signed up to join the task and finally 2 teams submitted runs. In total, we received 10 runs including 5 LADI+Others (O) runs and 5 LADI-based (L) runs. For detailed information about the approaches and results for individual teams' performances and runs, we refer the reader to the site reports \cite{tv22pubs} in the online workshop notebook proceedings. We present the overall results in this section.

The testing dataset had very rare occurrences of lava, snow/ice, landslide, and road washout features, so these four features were removed from all result calculations.

\begin{figure}[htbp]
    \centering
    \includegraphics[width=1.0 \linewidth]{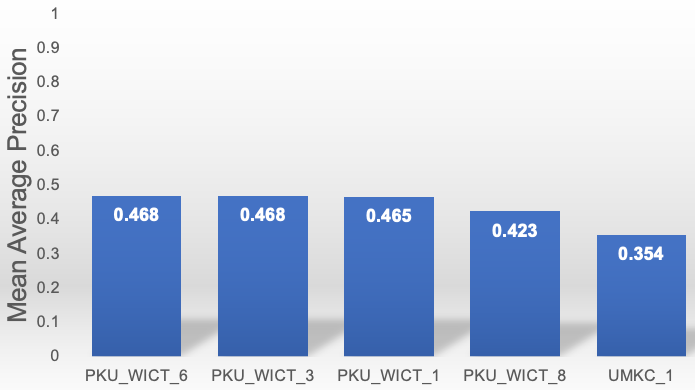}
    \caption{DSDI: Mean average precision score for each run with training type L.}
    \label{fig:dsdi.results.teams.ladi}
\end{figure}

\begin{figure}[htbp]
    \centering
    \includegraphics[width=1.0 \linewidth]{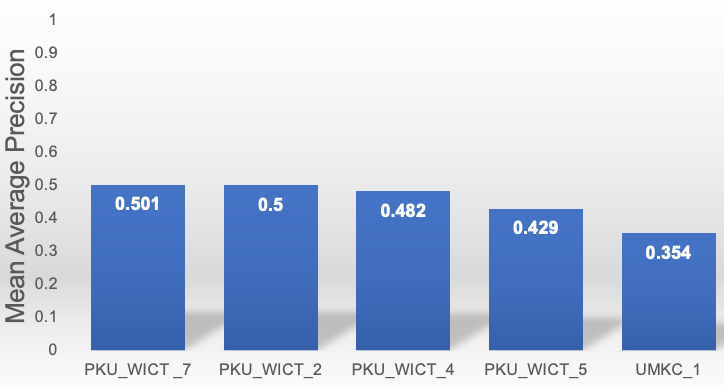}
    \caption{DSDI: Mean average precision score for each run with training type O.}
    \label{fig:dsdi.results.teams.ladi.others}
\end{figure}

Figures~\ref{fig:dsdi.results.teams.ladi} and~\ref{fig:dsdi.results.teams.ladi.others} show the mean average precision score for each run with training types L and O respectively. 
As previously noticed in the last two years, runs from the category O performed higher than those which was trained only using the LADI dataset.

\begin{figure}[htbp]
    \centering
    \includegraphics[width=1.0 \linewidth]{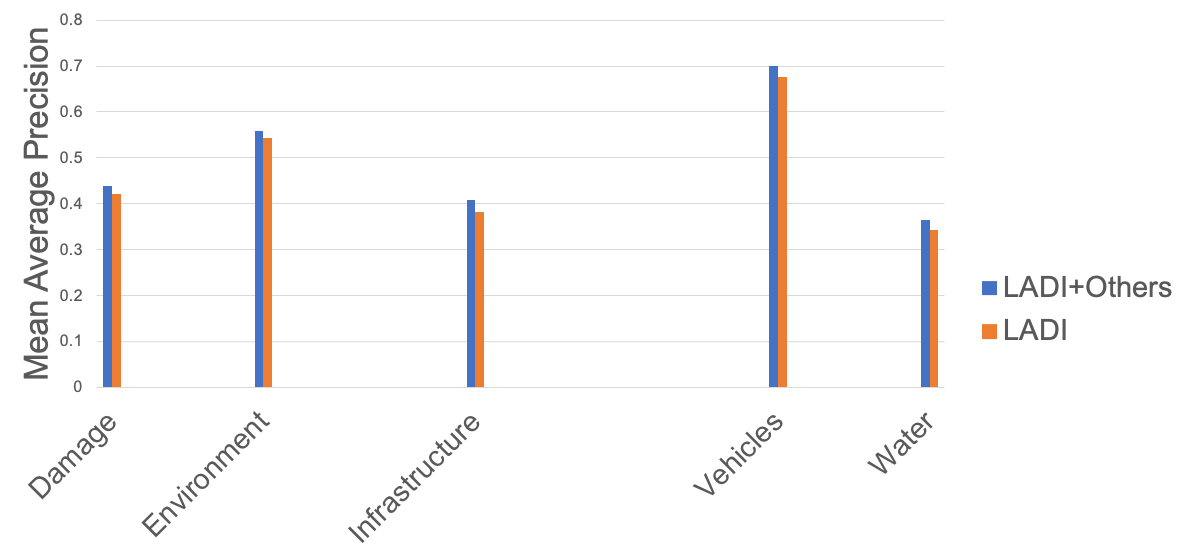}
    \caption{DSDI: Mean average precision by categories for both training types}
    \label{fig:dsdi.results.categories}
\end{figure}

Figure~\ref{fig:dsdi.results.categories} shows the mean average precision values organized by categories for run types L and O. The chart shows how the systems perform on features within each category. All categories with runs of L+O performed slightly higher than runs of type L.
The vehicles category scored the highest, while water scored the lowest overall.
  
\begin{figure*}[htbp]
    \centering
    \includegraphics[width=0.7 \linewidth]{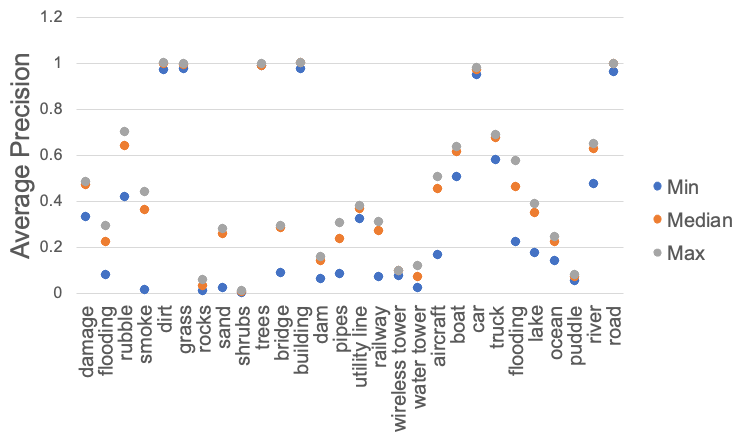}
    \caption{DSDI: Plot of average precision values for each feature for systems with training type L.}
    \label{fig:dsdi.avg.prec.ladi}
\end{figure*}

\begin{figure*}[htbp]
    \centering
    \includegraphics[width=0.7 \linewidth]{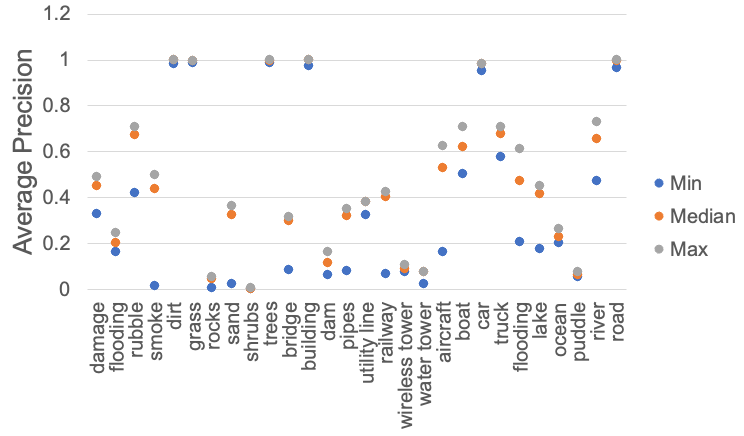}
    \caption{DSDI: Plot of average precision values for each feature for systems with training type O.}
    \label{fig:dsdi.avg.prec.ladi.others}
\end{figure*}

Figures~\ref{fig:dsdi.avg.prec.ladi} and~\ref{fig:dsdi.avg.prec.ladi.others} show the plot of average precision min, max and median scores for each feature for systems with run types L and O respectively. Systems tend to perform well on features that are commonly seen in training data, such as grass, trees, buildings, etc. However, features such as rocks, shrubs, dams, wireless towers, water towers, and puddles performed the lowest. In general, both sets of runs show the same pattern of performance across features.

\begin{figure*}[htbp]
    \centering
    \includegraphics[width=0.7 \linewidth]{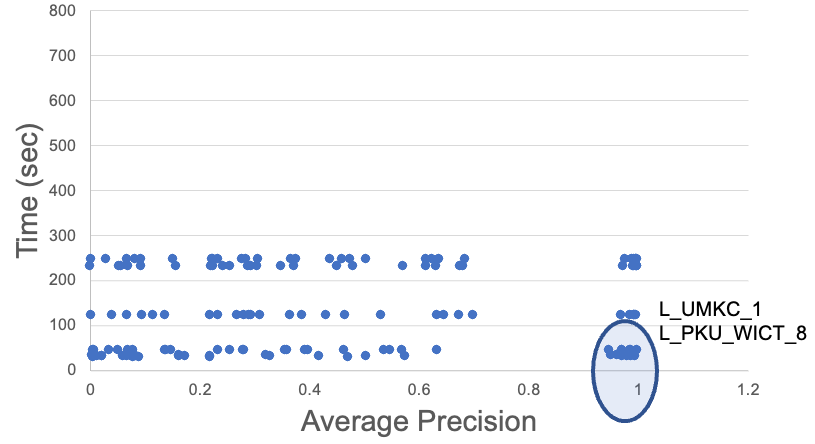}
    \caption{DSDI: Plot of average precision values against processing time for each feature for systems with training type L.}
    \label{fig:dsdi.time.ladi}
\end{figure*}

\begin{figure*}[htbp]
    \centering
    \includegraphics[width=0.7 \linewidth]{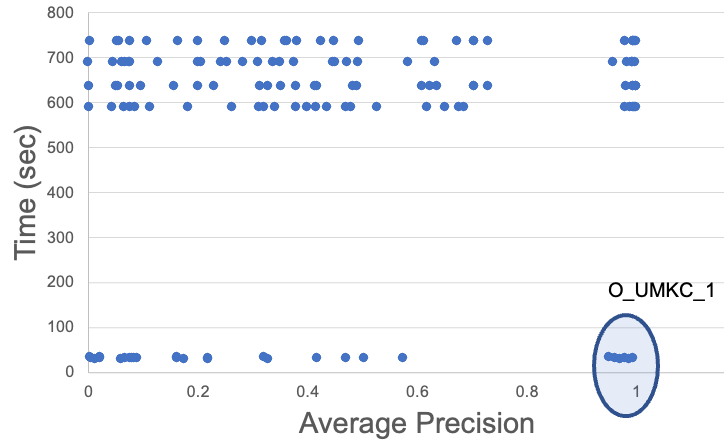}
    \caption{DSDI: Plot of average precision values against processing time for each feature for systems with training type O.}
    \label{fig:dsdi.time.ladi.others}
\end{figure*}

Figures~\ref{fig:dsdi.time.ladi} and ~\ref{fig:dsdi.time.ladi.others} show the submitted processing time reported by each run and feature for both L and O run types. Overall, LADI-based systems reported less processing time. While the majority of systems consumed more time, they did not gain too much in performance. The fastest system reported ~30 sec at max performance.

\begin{figure*}[htbp]
    \centering
    \includegraphics[width=1.0 \linewidth]{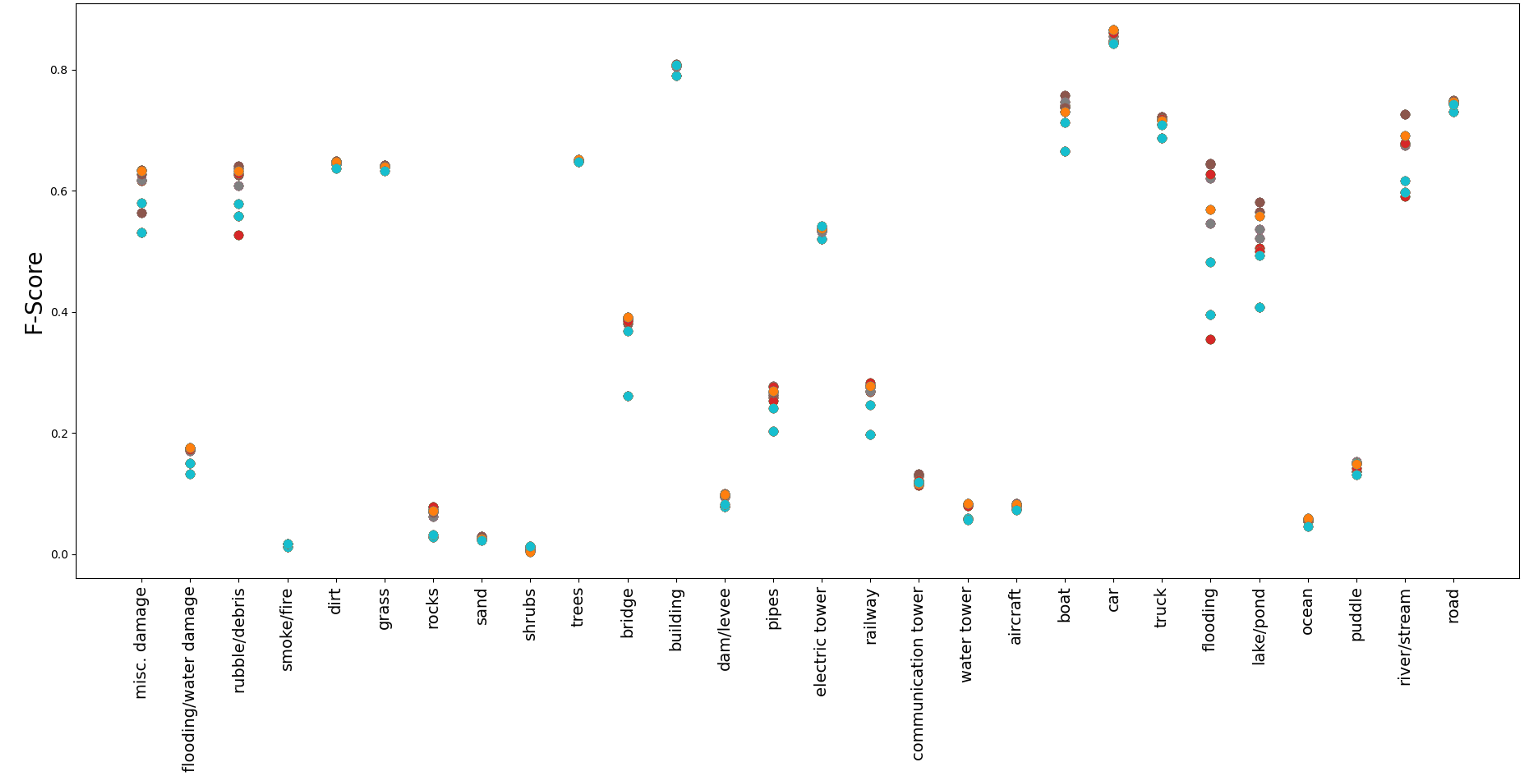}
    \caption{DSDI: F-measure for all the runs.}
    \label{fig:dsdi.F.measure}
\end{figure*}

We also reported the true positives, true negatives, false positives, and false negatives for each run. The F-measure using these values is shown in Figure~\ref{fig:dsdi.F.measure}. About half the features, for all runs, achieved 0.5 or more F-score. Some features show a spread in scores across runs (e.g. flooding, debris, lake, river). While other features are condensed (e.g. smoke, dirt, grass, sand, shrubs, trees, aircraft, car, ocean, road). This may be due to they are either easy features to recognize (e.g. car) or hard (e.g. sand, smoke). In addition, since some features are not frequent in the ground truth as others, this affects the retrieval performance (e.g. sand, smoke, shrubs).

\subsubsection{Conclusion and Future Work}

This was the third iteration of the DSDI task. While the participation in the task decreased from last year, teams performed reasonably well. A new test dataset from various event sources was employed representing more diversity. Performance varied by feature. The O run type performed higher than L-based runs. In general, few runs/features achieve high performance and high efficiency at the same time.
The task experienced some challenges such as small datasets and limited resources for annotation.
The training and testing dataset should be from the same distribution. However, it is hard to do with different nature of calamities. Some known issues with the training data are:

\begin{enumerate}
    \item The LADI dataset labels can be noisy due to crowd-sourced annotation.
    \item There is a class imbalance as certain labels are far more prevalent than others.
    \item The datasets are mostly limited to a certain types of disasters. It is not simple to have representation for all disaster labels since data acquisition requires multiple sources.
\end{enumerate}

The DSDI test dataset was labeled by dedicated annotators, which resulted in cleaner annotation. All annotations will be added to the existing DSDI dataset resources. As the task had little participation compared to the last 2 years. It was decided to discontinue the task in 2023.

\subsection{Video to Text}
\begin{table*} \begin{center}

\begin{tabular}[htbp]{|c|c|}
\hline
&Number of runs\\ \hline\hline
Kslab&  4\\ \hline
MLVC\_HDU&  4\\ \hline
RUCAIM3-Tencent& 4\\ \hline
VIDION& 4\\ \hline
WasedaMeiseiSoftbank & 4\\ \hline
ELT\_01&  4\\ \hline
\end{tabular}
\caption{VTT: List of teams participating and their submitted runs.}
\label{tab:vtt.participants}
\end{center}
\end{table*}

Automatic annotation of videos using natural language text descriptions has been a long-standing goal of computer vision. The task involves understanding many concepts such as objects, actions, scenes, person-object relations, the temporal order of events throughout the video, to mention a few. In recent years there have been major advances in computer vision techniques that enabled researchers to start practical work on solving the challenges posed in automatic video captioning. 

There are many use-case application scenarios that can greatly benefit from the technology, such as video summarization in the form of natural language,  facilitating the searching and browsing of video archives using such descriptions, describing videos as an assistive technology, etc. In addition, learning video interpretation and temporal relations among events in a video will likely contribute to other computer vision tasks, such as the prediction of future events from the video. 

The Video to Text (VTT) task was introduced in TRECVID 2016. Since then, there have been substantial improvements in the dataset and evaluation. 
Essentially, each year's testing dataset is being appended to previous year's development dataset. In addition, since 2021, a subset of videos is being dedicated to a progress subtask for which the ground truth is withheld and participants submit results from 2021 to 2023. 
They will then be able to compare their systems across the three years to measure improvement over the years on the same set of videos.

\subsubsection{System Task}
For each video, automatically generate a text description of 1 sentence independently and without taking into consideration the existence of any annotated descriptions for the videos. Up to 4 runs are allowed per team.

For this year, 6 teams participated in the VTT task. The 6 teams submitted a total of 24 runs. 
A summary of participating teams is shown in Table~\ref{tab:vtt.participants}.

\subsubsection{Data}

During 2020 and 2021, the VTT data was selected from the V3C2 data collection. In previous years, the VTT testing dataset consisted of Twitter Vine videos, which generally had a duration of 6 seconds. In 2019, we supplemented the dataset with videos from Flickr. The V3C dataset \cite{rossetto2019v3c} is a large collection of videos from Vimeo. It also provides us with the advantage that we can distribute the videos rather than links, which may not be available in the future. This year, the testing dataset was selected from the V3C1 collection which is very similar to V3C2 in its characteristics.

For the purpose of this task, we only selected video segments with lengths between 3 and 10 seconds. A total of 2008 video segments were annotated manually by multiple annotators for this year's task. Since we have selected 300 videos for our progress set in 2021, our results will be reported for 2008 new videos (non-progress) and the 300 videos in progress set.

\begin{figure}[htbp]
  \centering
  \includegraphics[width=1.0\linewidth]{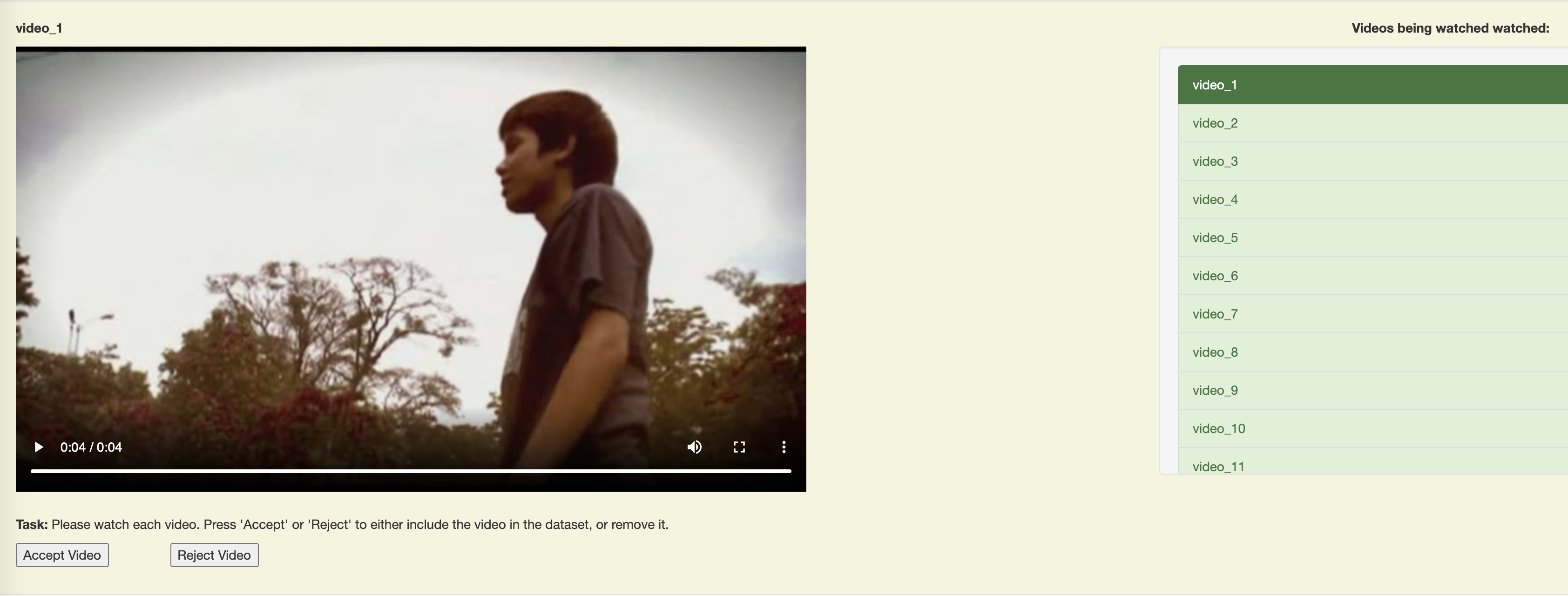}
  \caption{VTT: Screenshot of video selection tool.}
  \label{fig:vtt.video.selection}
\end{figure}

It is important for a good dataset to have a diverse set of videos. We reviewed around 7639 videos and selected 2008 videos. Figure \ref{fig:vtt.video.selection} shows a screenshot of the video selection tool that was used to decide whether a video was to be selected or not. We tried to ensure that the videos covered a large set of topics. If we came across a large number of videos that looked similar to previously selected clips, they were rejected. We also removed the following types of videos:
\begin{itemize}
    \item Videos with multiple, unrelated segments that are hard to describe, even for humans.
    \item Any animated videos.
    \item Other videos that may be considered inappropriate or offensive.
\end{itemize}

\begin{table}[htbp]
    \centering
    
    \begin{tabular}{|c|c|c|}
    \hline
    Annotator  & Avg. Length & Total Videos Watched\\
    \hline\hline
      1  & 24.84 & 2008\\
      \hline
      2  & 18.06 & 2008\\
      \hline
      3  & 21.93 & 2008\\
      \hline
      4  & 26.09 & 2008\\
      \hline
      5  & 21.50 & 2008\\
      \hline
    \end{tabular}
    \caption{VTT: Average number of words per sentence for all the annotators. The table also shows the number of videos watched by each annotator.}
    \label{tab:avg_gt_length}
\end{table}

\paragraph {\textbf{Annotation Process}}

The videos were divided among 5 annotators, with each video being annotated once by each to create 5 annotations per video. 

The annotators were asked to include and combine into 1 sentence, if appropriate and available, four facets of the video they are describing:

\begin{itemize}
\item{\textbf{Who} is the video showing (e.g., concrete objects and beings, kinds of persons, animals, or things)}?
\item{\textbf{What} are the objects and beings doing (generic actions, conditions/state or events)}?
\item{\textbf{Where} was the video taken (e.g., locale, site, place, geographic location, architectural)}?
\item{\textbf{When} was the video taken (e.g., time of day, season)}?
\end{itemize}

Different annotators provide varying amounts of detail when describing videos. Some people try to incorporate as much information as possible about the video, whereas others may write more compact sentences. Table~\ref{tab:avg_gt_length} shows the average number of words per sentence for each of the annotators. The average sentence length varies from 18 words to 26 words, emphasizing the difference in descriptions provided by the annotators. The overall average sentence length for the dataset is 24.46 words.

Furthermore, the annotators were also asked the following questions for each video:
\begin{itemize}
\item Please rate how difficult it was to describe the video.
\begin{enumerate}
    \item Very Easy
    \item Easy
    \item Medium
    \item Hard
    \item Very Hard
\end{enumerate}
\item How likely is it that other assessors will write similar descriptions for the video?
\begin{enumerate}
    \item Not Likely
    \item Somewhat Likely
    \item Very Likely
\end{enumerate}
\end{itemize}

The average score for the first question was 2.52 (on a scale of 1 to 5), showing that the annotators thought the videos were close to medium level of difficulty on average. The average score for the second question was 2.37 (on a scale of 1 to 3), meaning that they thought that other people would write a similar description as them for most videos. The two scores are negatively correlated as annotators are more likely to think that other people will come up with similar descriptions for easier videos. The Pearson correlation coefficient between the two questions is -0.61.

\subsubsection{Submissions}

Systems were required to specify the run types based on the types of training data and features used. 

The list of training data types is as follows:
\begin{itemize}
    \item `I': Training using image captioning datasets only.
    \item `V': Training using video captioning datasets only.
    \item `B': Training using both image and video captioning datasets.
\end{itemize}

The feature types can be one of the following:
\begin{itemize}
    \item `V': Only visual features are used.
    \item `A': Both audio and visual features are used.
\end{itemize}

In total, 5 runs were of type ``BV" (used only visual features from both image and video datasets), 5 runs used ``IV" (used image datasets with visual features), and 12 runs are of type ``VV" (used video datasets with visual only features), and 2 runs are of type ``VA" (used video datasets with visual and audio features).

Teams were also asked to specify the loss function used for their runs. Loss functions reported were mainly based on cross-entropy (16 runs). Four runs reported KLDivloss (Kullback-Leibler divergence), while 4 other runs applied self-critical reinforcement learning loss.

\subsubsection{Evaluation and Metrics}

The description generation task scoring was done automatically using different metrics. We also used a human evaluation metric on selected runs to compare with the automatic metrics. 

METEOR (Metric for Evaluation of Translation with Explicit ORdering) \cite{banerjee2005meteor} and BLEU (BiLingual Evaluation Understudy) \cite{papineni2002bleu} are standard metrics in machine translation (MT). BLEU was one of the first metrics to achieve a high correlation with human judgments of quality. It is known to perform poorly if it is used to evaluate the quality of individual sentence variations rather than sentence variations at a corpus level. In the VTT task the videos are independent and there is no corpus to work from. Thus, our expectations are lowered when it comes to evaluation by BLEU.  METEOR is based on the harmonic mean of unigram or n-gram precision and recall in terms of overlap between two input sentences. It redresses some of the shortfalls of BLEU such as better matching synonyms and stemming, though the two measures seem to be used together in evaluating MT.

The CIDEr (Consensus-based Image Description Evaluation) metric \cite{vedantam2015cider} is borrowed from image captioning. It computes TF-IDF (term frequency inverse document frequency) for each n-gram to give a sentence similarity score. The CIDEr metric has been reported to show high agreement with consensus as assessed by humans. We also report scores using CIDEr-D, which is a modification of CIDEr to prevent ``gaming the system''. 

The SPICE (Semantic Propositional Image Caption Evaluation) metric \cite{spice2016} is another metric that has gained popularity in image captioning evaluation. The metric uses scene graph similarity between generated captions and the ground truth instead of n-grams.

The STS (Semantic Textual Similarity) metric \cite{han2013umbc} was also applied to the results, as in the previous years of this task. This metric measures how semantically similar the submitted description is to one of the ground truth descriptions.

In addition to automatic metrics, the description  generation task includes human evaluation of the quality of automatically generated captions.
Recent developments in Machine Translation evaluation have seen the emergence of DA (Direct Assessment), a method shown to produce highly reliable human evaluation results for MT and Natural Language Generation \cite{DA,msr20}. 
DA now constitutes the official method of ranking in main MT benchmark evaluations \cite{WMT17,barrault-EtAl:2020:WMT1}. 

With respect to DA for evaluation of video captions (as opposed to MT output), human assessors are presented with a video and a single caption. After watching the video,  assessors rate how well the caption describes what took place in the video on a 0--100 rating scale \cite{graham2018evaluation}. Large numbers of ratings are collected for captions before ratings are combined into an overall average system rating (ranging from 0 to 100\,\%). Human assessors are recruited via Amazon's Mechanical Turk (AMT), with quality control measures applied to filter out or downgrade the weightings from workers unable to demonstrate the ability to rate good captions higher than lower quality captions. This is achieved by deliberately ``polluting'' some of the manual (and correct) captions with linguistic substitutions to generate captions whose semantics are questionable. For instance, we might substitute a noun for another noun and turn the manual caption ``A man and a woman are dancing on a table" into ``A {\em horse} and a woman are dancing on a table'', where ``horse'' has been substituted for ``man''.  We expect such automatically-polluted captions to be rated poorly and when an AMT worker correctly does this, the ratings for that worker are improved.

DA was first used as an evaluation metric in TRECVID 2017. This metric has been used every year since then to rate each team's primary run.

\subsubsection{Results}

The description generation task scoring was done using popular automatic metrics that compare the system generation captions with ground truth captions as provided by human assessors. We also continued the use of Direct Assessment, which was introduced in TRECVID 2017, to compare the submitted runs. 

\begin{figure}[htbp]
  \centering
  \includegraphics[width=1.0\linewidth]{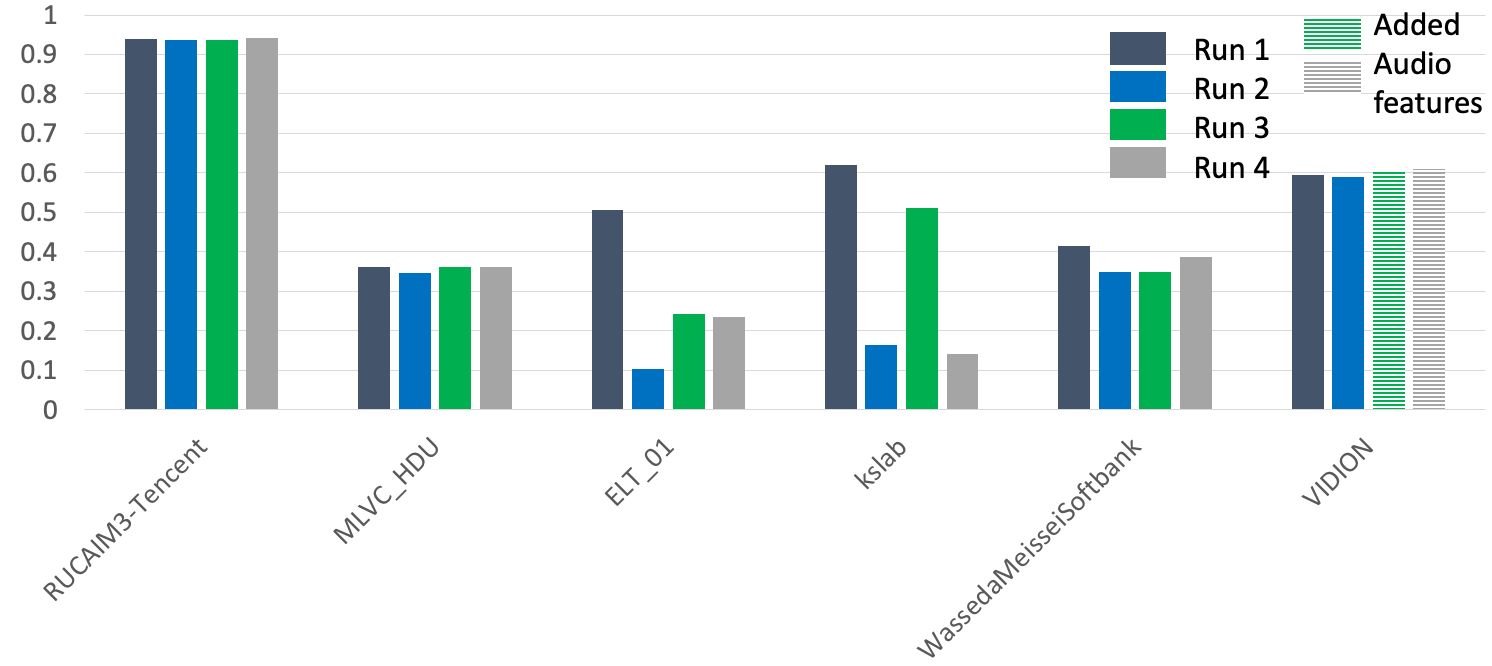}
  \caption{VTT: Comparison of all runs using the CIDEr metric.}
  \label{fig:vtt.cider.results}
\end{figure}

\begin{figure}[htbp]
  \centering
  \includegraphics[width=1.0\linewidth]{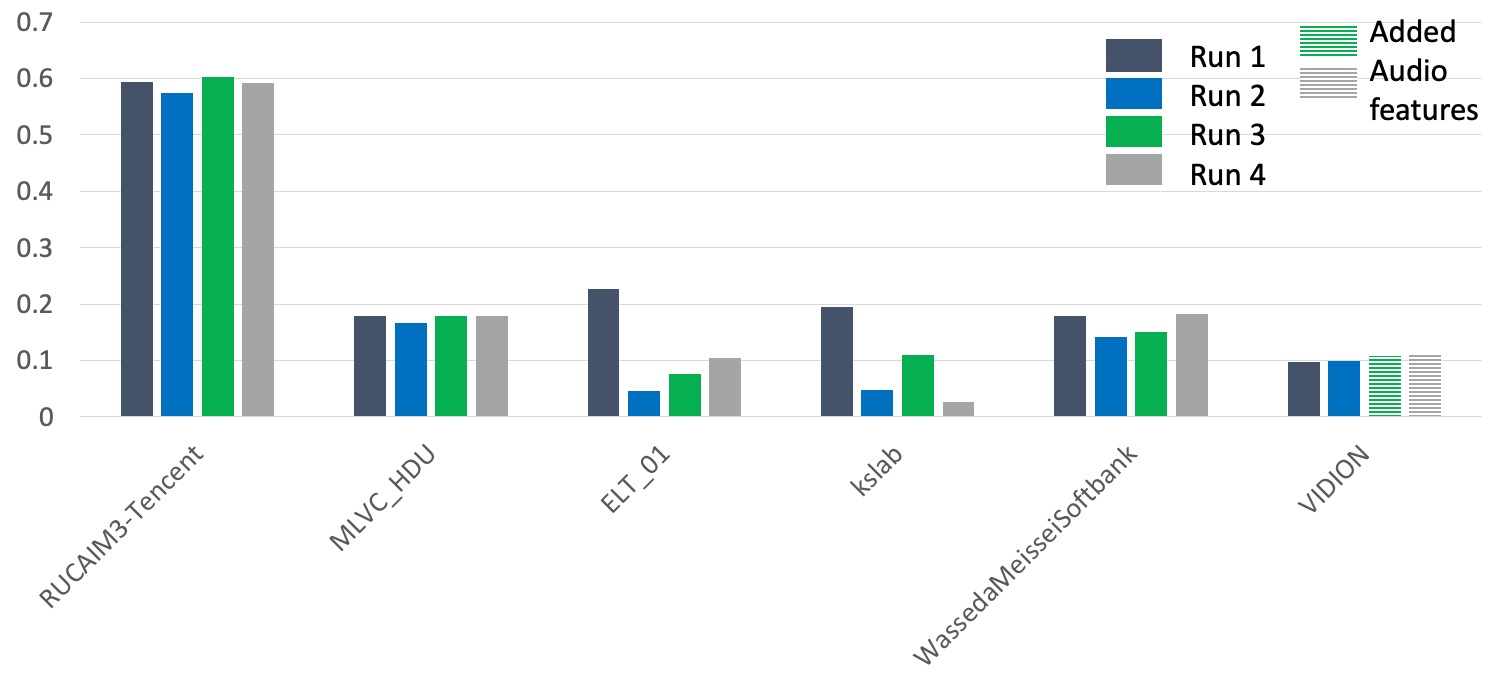}
  \caption{VTT: Comparison of all runs using the CIDEr-D metric.}
  \label{fig:vtt.ciderd.results}
\end{figure}

\begin{figure}[htbp]
  \centering
  \includegraphics[width=1.0\linewidth]{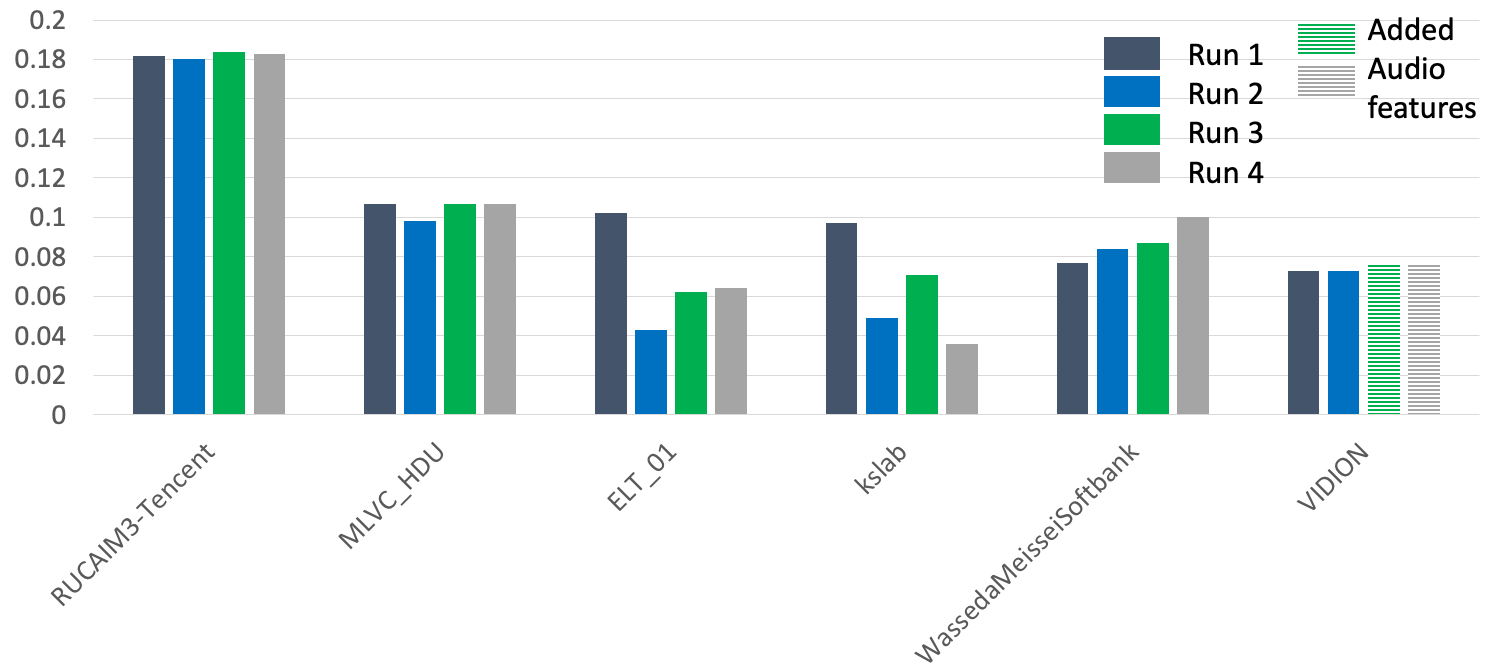}
  \caption{VTT: Comparison of all runs using the SPICE metric.}
  \label{fig:vtt.spice.results}
\end{figure}

\begin{figure}[htbp]
  \centering
  \includegraphics[width=1.0\linewidth]{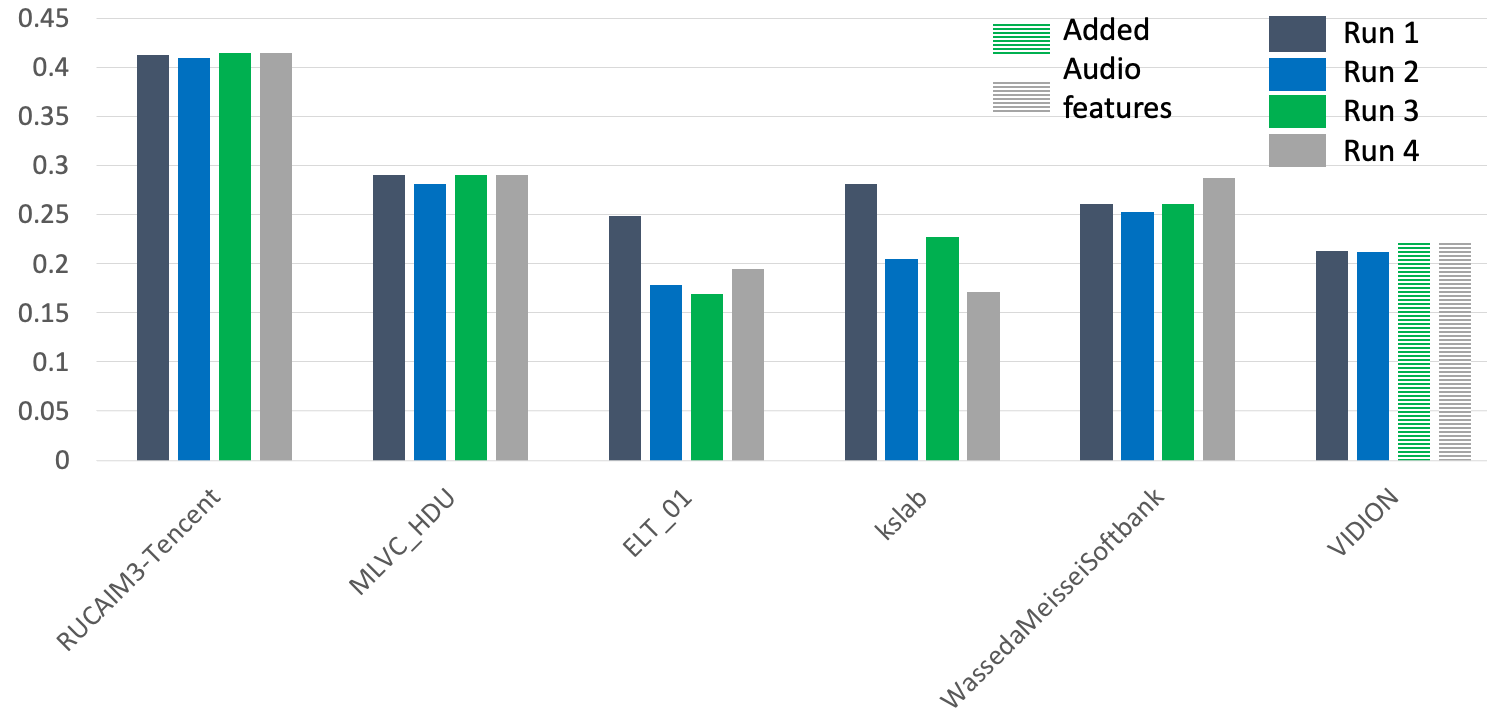}
  \caption{VTT: Comparison of all runs using the METEOR metric.}
  \label{fig:vtt.meteor.results}
\end{figure}

\begin{figure}[htbp]
  \centering
  \includegraphics[width=1.0\linewidth]{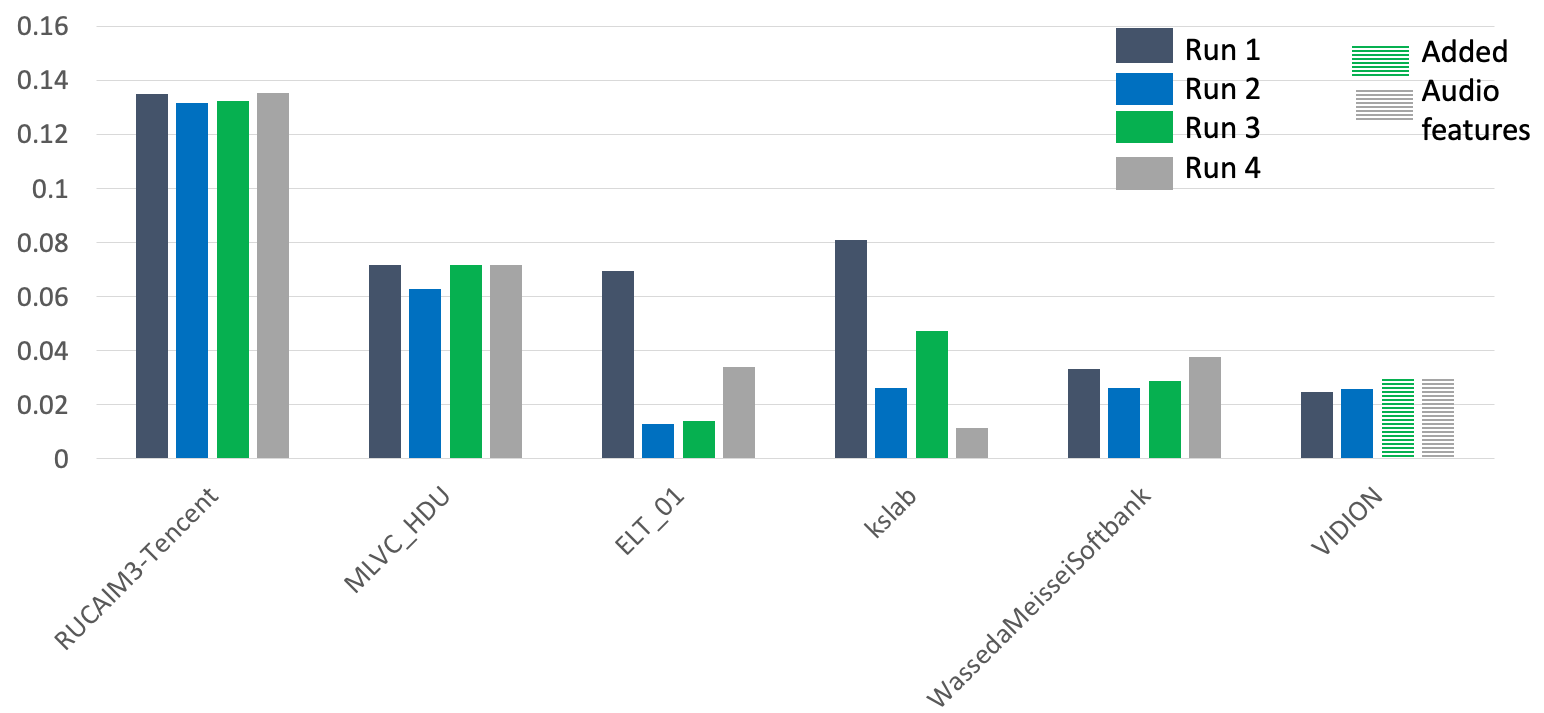}
  \caption{VTT: Comparison of all runs using the BLEU metric.}
  \label{fig:vtt.bleu.results}
\end{figure}

\begin{figure}[htbp]
  \centering
  \includegraphics[width=1.0\linewidth]{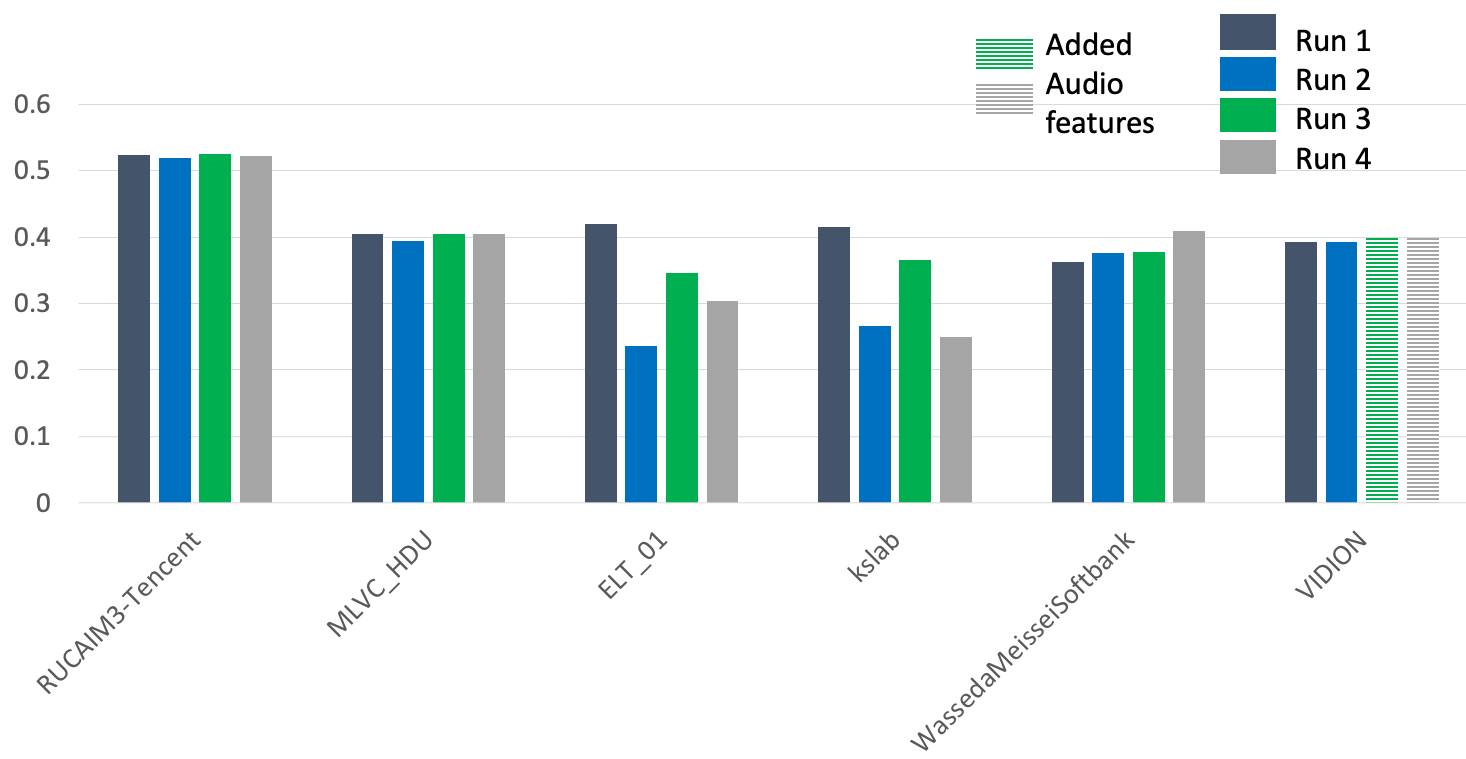}
  \caption{VTT: Comparison of all runs using the STS metric.}
  \label{fig:vtt.sts.results}
\end{figure}

The metric score for each run is calculated as the average of the metric scores for all the descriptions within that run.  Figure~\ref{fig:vtt.cider.results} shows the performance comparison of all teams using the CIDEr metric. All runs submitted by each team are shown in the graph. Figure~\ref{fig:vtt.ciderd.results} shows the scores for the CIDEr-D metric, which is a modification of CIDEr. Figure~\ref{fig:vtt.spice.results} shows the SPICE metric scores. Figures~\ref{fig:vtt.meteor.results} and~\ref{fig:vtt.bleu.results} show the scores for METEOR and BLEU metrics respectively. The STS metric allows the comparison between two sentences. For this reason, the captions are compared to a single ground truth description at a time, resulting in 5 STS scores. We will report the average of these scores as the STS score, and Figure~\ref{fig:vtt.sts.results} shows how the runs compare on this metric. 
In general, we can see that RUCAIM3 team achieved top performance across all automatic metrics. While performance across metrics for each team is relatively comparable, the STS metric shows the top performance run from each team is very close to each other.

Table~\ref{tab:vtt.auto.metric.corr.run} shows the correlation between the different metric scores for all the runs. The metrics correlate very well, which shows that they agree on the overall scoring of the runs. The correlation scores between all other metrics range between 0.8 and 0.98. 

\begin{table*}
\centering
\begin{tabular}{lrrrrrr}
\toprule
{} &  BLEU  & METEOR & CIDER & CIDER-D &  SPICE & STS \\
\midrule
BLEU    &        1.000 &         0.95 &        0.80 &         0.94 &       0.96 &  0.86 \\
METEOR  &        0.95 &          1.000 &       0.80 &         0.96 &       0.98 &  0.89 \\
CIDER   &        0.80 &          0.80 &        1.000 &         0.85 &       0.85 &  0.91 \\
CIDER-D &        0.94 &          0.96 &        0.85 &         1.000 &       0.98 &  0.87 \\
SPICE   &        0.96 &          0.98 &        0.85 &         0.98 &       1.000 &  0.94 \\
STS     &        0.86 &          0.89 &        0.91 &         0.87 &       0.94 &  1.000 \\
\bottomrule
\end{tabular}

\caption{VTT: Correlation between overall run scores for automatic metrics.}
\label{tab:vtt.auto.metric.corr.run}
\end{table*}

\begin{figure}[htbp]
  \centering
  \includegraphics[width=1.0\linewidth]{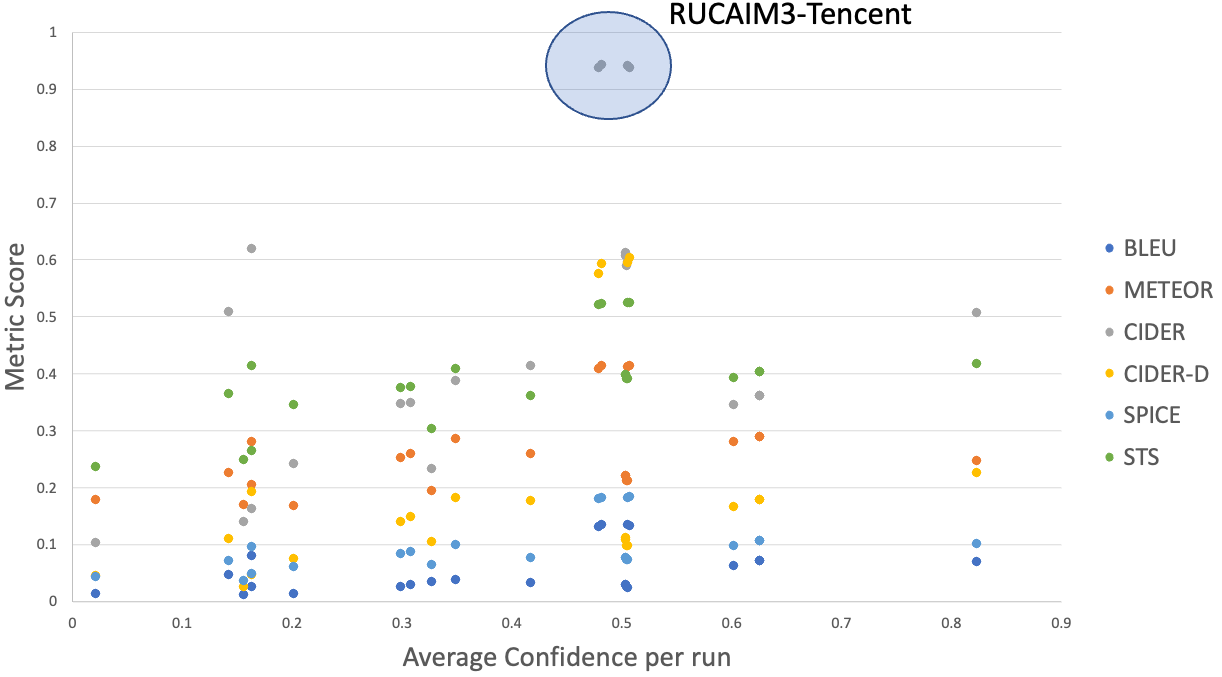}
  \caption{VTT: Plot of the system reported sentence confidence scores against the various metric scores.}
  \label{fig:vtt.conf.metric.corr}
\end{figure}

Teams were asked to provide a confidence score for each generated sentence. Figure~\ref{fig:vtt.conf.metric.corr} shows the submitted average confidence scores for each run against each metric score. There seems to be some correlation (not very strong) between confidence and metric scores. In addition, it can be shown that the RUCAIM3 team had few runs with very high CIDER score at mid level confidence. Overall STS, CIDER, and CIDER-D metrics correlated the highest with confidence scores.

\begin{figure}[htbp]
  \centering
  \includegraphics[width=1.0\linewidth]{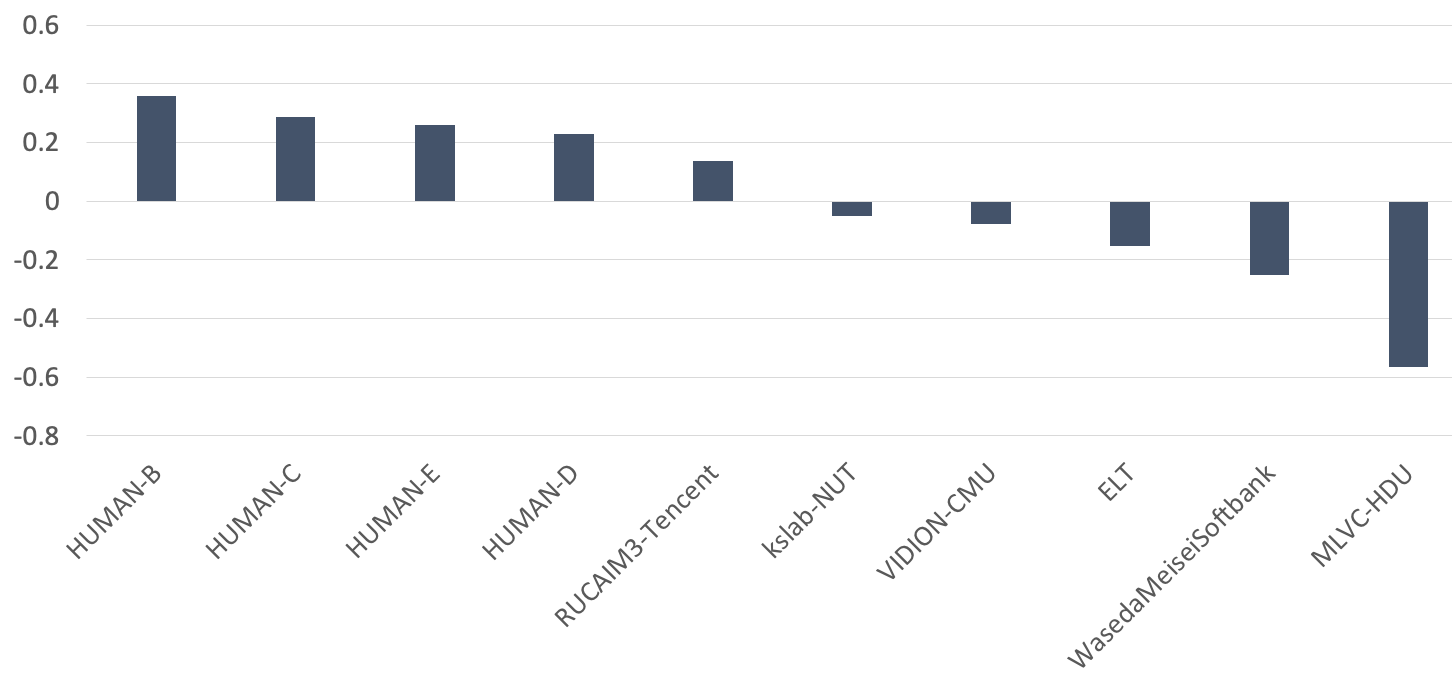}
  \caption{VTT: Average DA score per system after standardization per individual worker's mean and standard deviation score.}
  \label{fig:vtt.da.z.results}
\end{figure}

Figure~\ref{fig:vtt.da.z.results} shows the average DA score per system after it is standardized per individual AMT worker's mean and standard deviation score. The DA raw scores are micro-averaged per caption, and then averaged over all videos. The HUMAN systems represent manual captions provided by assessors. As expected, captions written by assessors outperform the automatic systems. The top system based on automatic metrics still outperforms other systems based on the DA experiments.

\begin{figure}[htbp]
  \centering
  \includegraphics[width=1.0\linewidth]{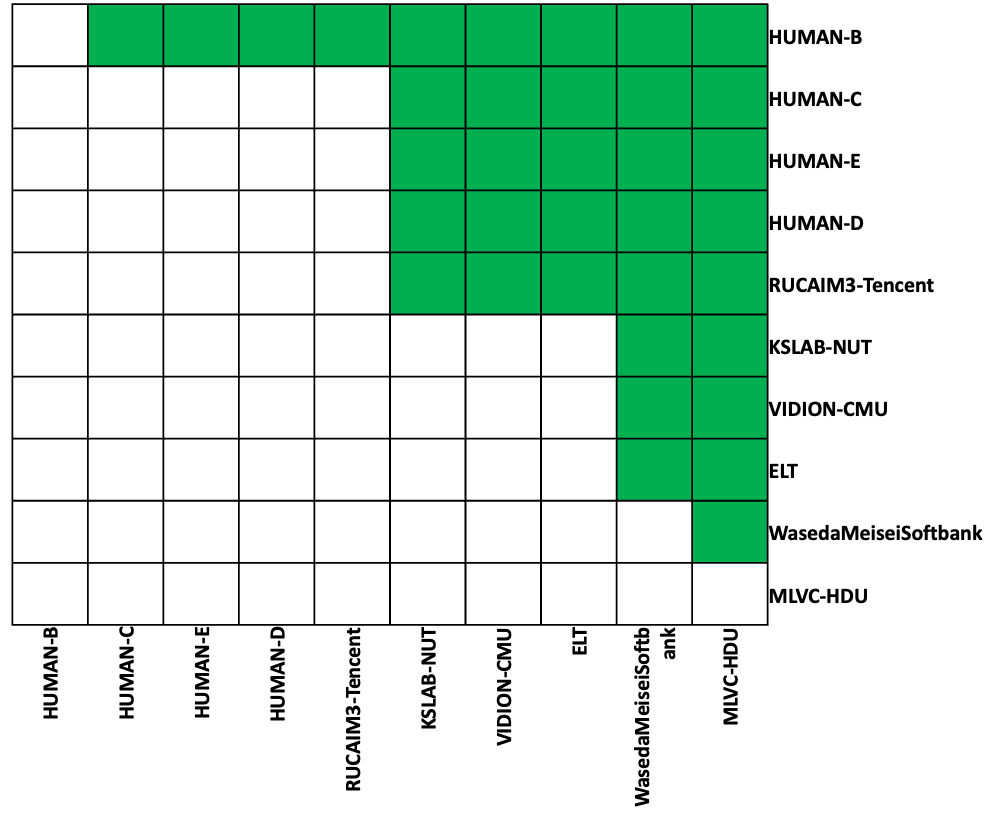}
  \caption{VTT: Comparison of the primary runs of each team with respect to the DA score. The `HUMAN' system is ground truth captions. Green squares indicate a significantly better result for the row over the column. }
  \label{fig:vtt.da.significance}
\end{figure}

Figure~\ref{fig:vtt.da.significance} shows how the systems compare according to DA. The green squares indicate that the system in the row is significantly better than the system shown in the column (p \textless 0.05). The figure shows that no system reaches the level of human performance. Among the systems, RUCAIM3-Tencent is significantly better than the other systems.  

Table~\ref{tab:vtt.da.metric.corr} shows the correlation between different overall metric scores for the primary runs of all teams. The `DA\_Z' metric is the score generated by humans. The score correlates positively with all metrics. The correlation ranged between 0.41 to 0.89 with CIDER and STS achieving the highest correlation with DA.   

\begin{table*}
\centering
\begin{tabular}{lrrrrrrr}
\toprule
 {} &  CIDER &  CIDER-D &  SPICE &  METEOR &  BLEU & STS\\
\midrule
 DA\_Z & 0.89 & 0.59 &  0.47 &  0.41 &  0.44 &  0.62\\
\bottomrule

\end{tabular}
\caption{VTT: Correlation between DA and automatics metrics for the primary runs only}
\label{tab:vtt.da.metric.corr}
\end{table*}

\begin{figure}[htbp]
  \centering
  \includegraphics[width=1.0\linewidth]{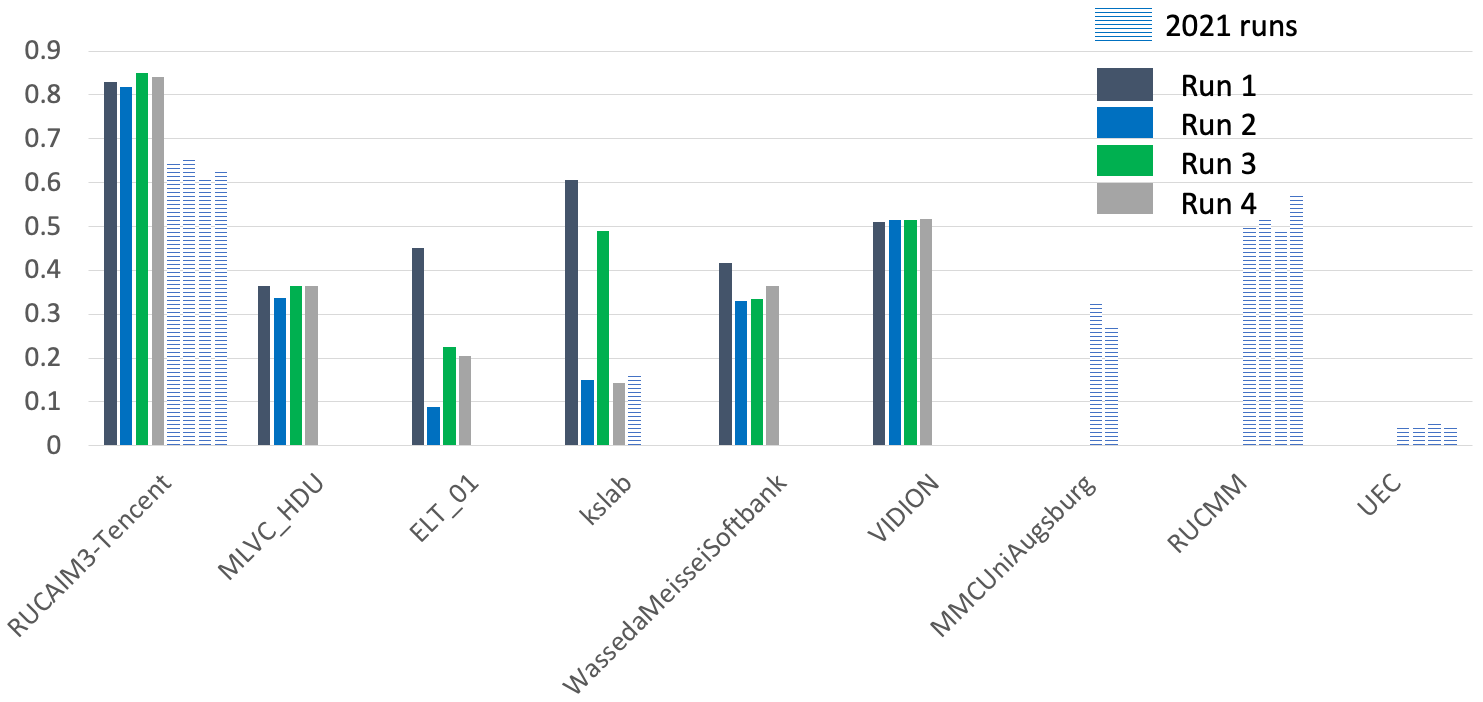}
  \caption{VTT: Comparison of all progress runs (submitted in 2021 - 2022) using the CIDEr metric.}
  \label{fig:vtt.cider.results.p}
\end{figure}

\begin{figure}[htbp]
  \centering
  \includegraphics[width=1.0\linewidth]{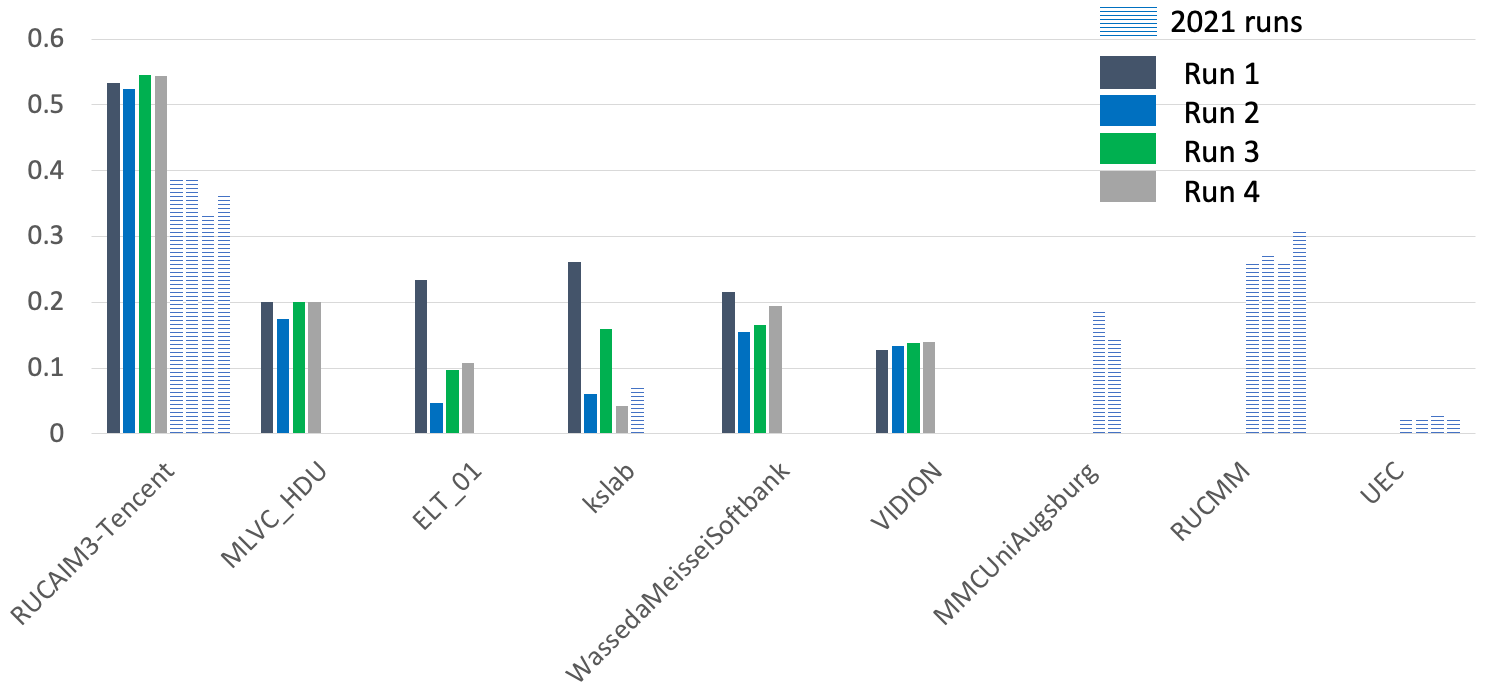}
  \caption{VTT: Comparison of all progress runs (submitted in 2021 - 2022) using the CIDEr-D metric.}
  \label{fig:vtt.ciderd.results.p}
\end{figure}

\begin{figure}[htbp]
  \centering
  \includegraphics[width=1.0\linewidth]{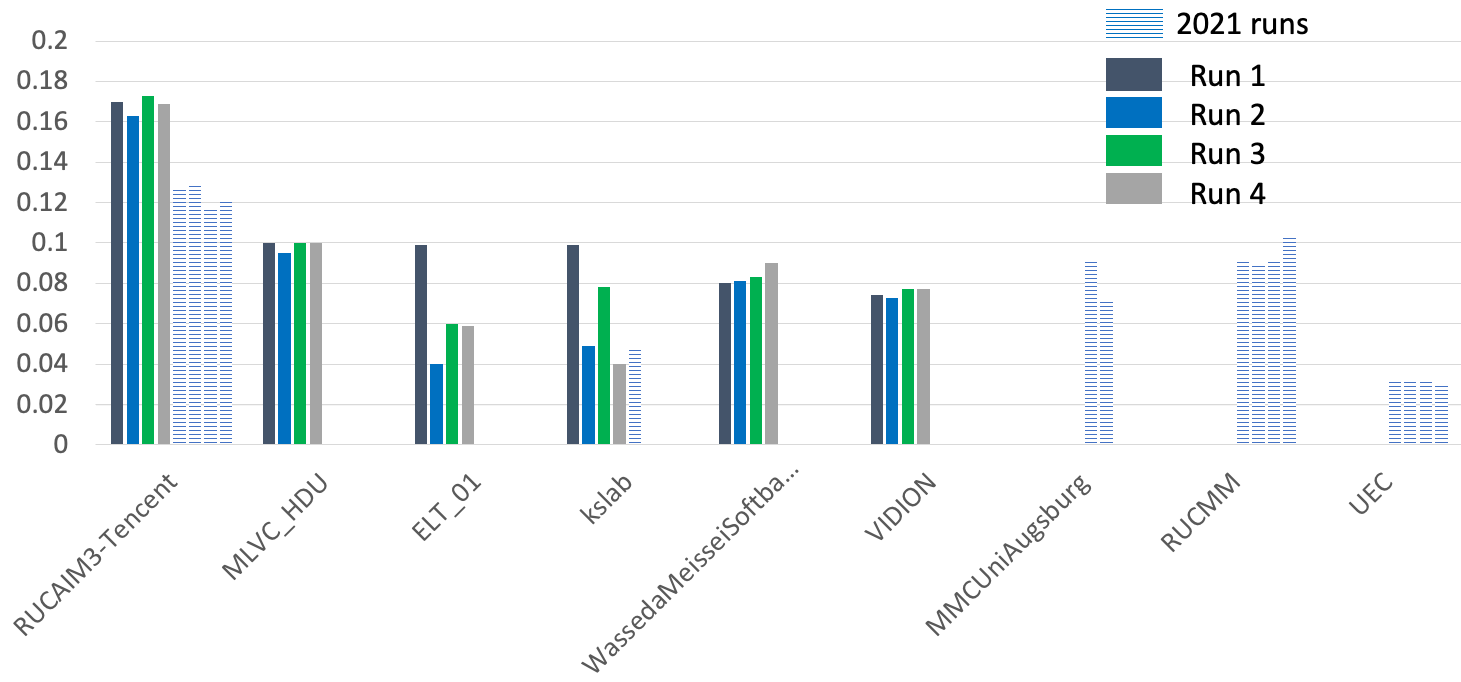}
  \caption{VTT: Comparison of all progress runs (submitted in 2021 - 2022) using the SPICE metric.}
  \label{fig:vtt.spice.results.p}
\end{figure}

\begin{figure}[htbp]
  \centering
  \includegraphics[width=1.0\linewidth]{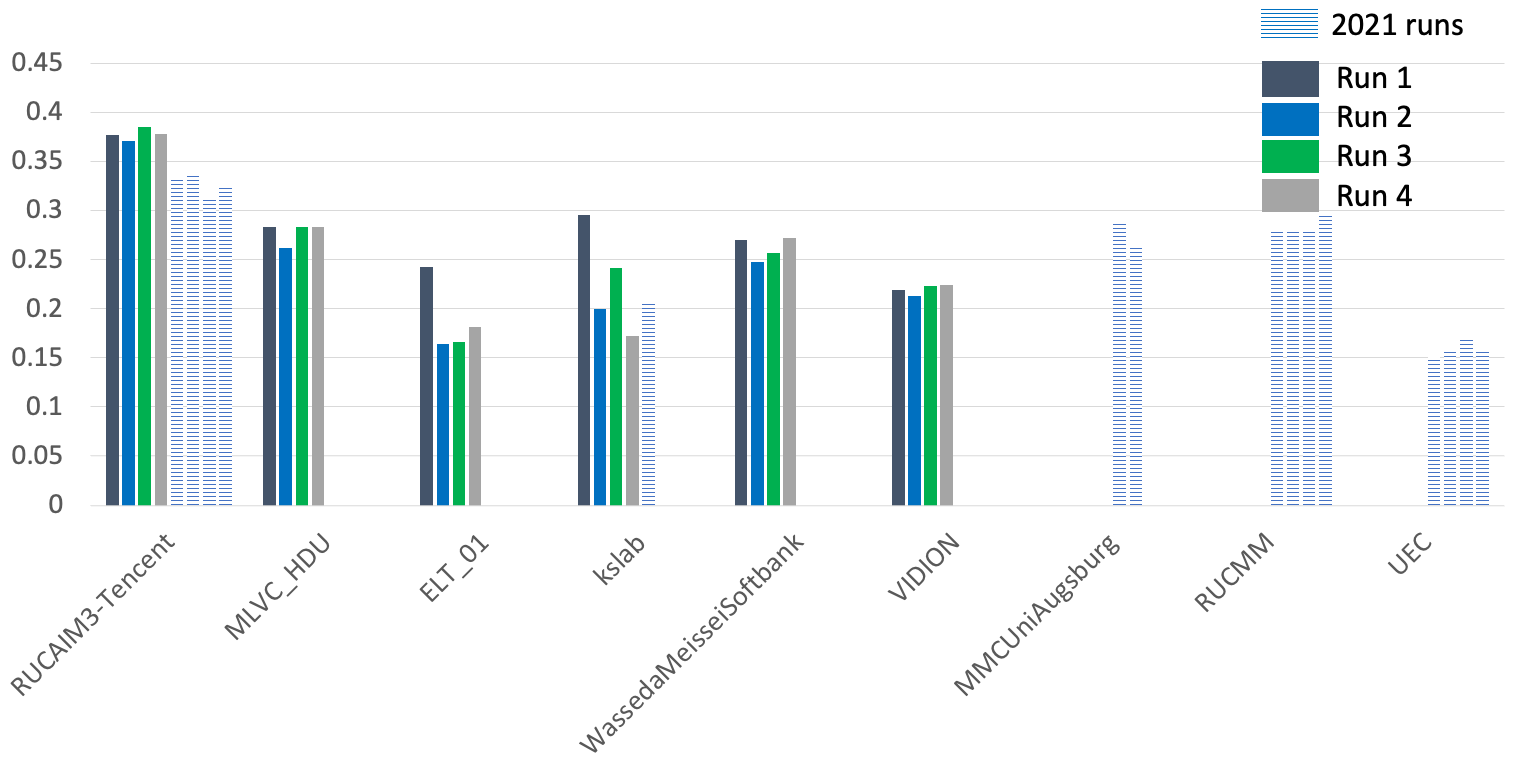}
  \caption{VTT: Comparison of all progress runs (submitted in 2021 - 2022) using the METEOR metric.}
  \label{fig:vtt.meteor.results.p}
\end{figure}

\begin{figure}[htbp]
  \centering
  \includegraphics[width=1.0\linewidth]{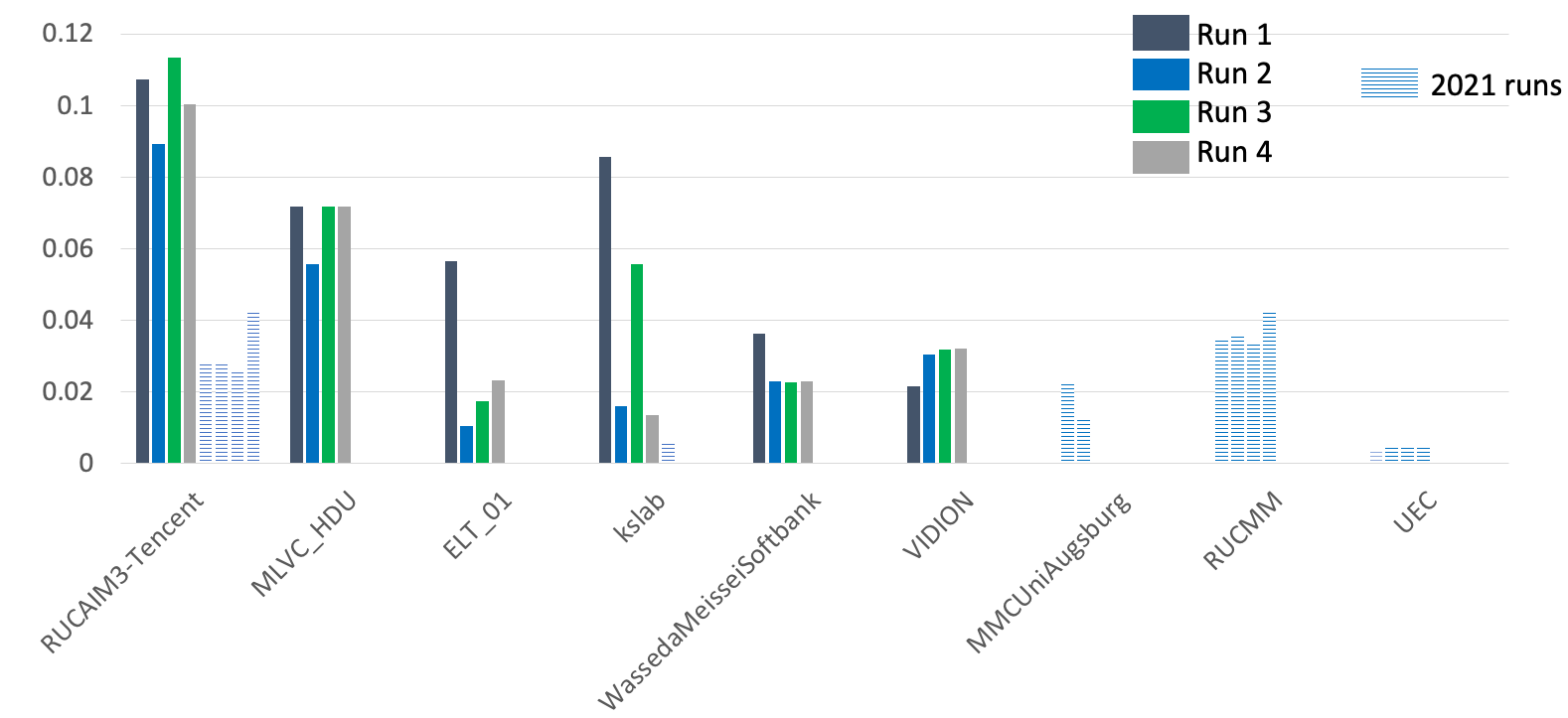}
  \caption{VTT: Comparison of all progress runs (submitted in 2021 - 2022) using the BLEU metric.}
  \label{fig:vtt.bleu.results.p}
\end{figure}

\begin{figure}[htbp]
  \centering
  \includegraphics[width=1.0\linewidth]{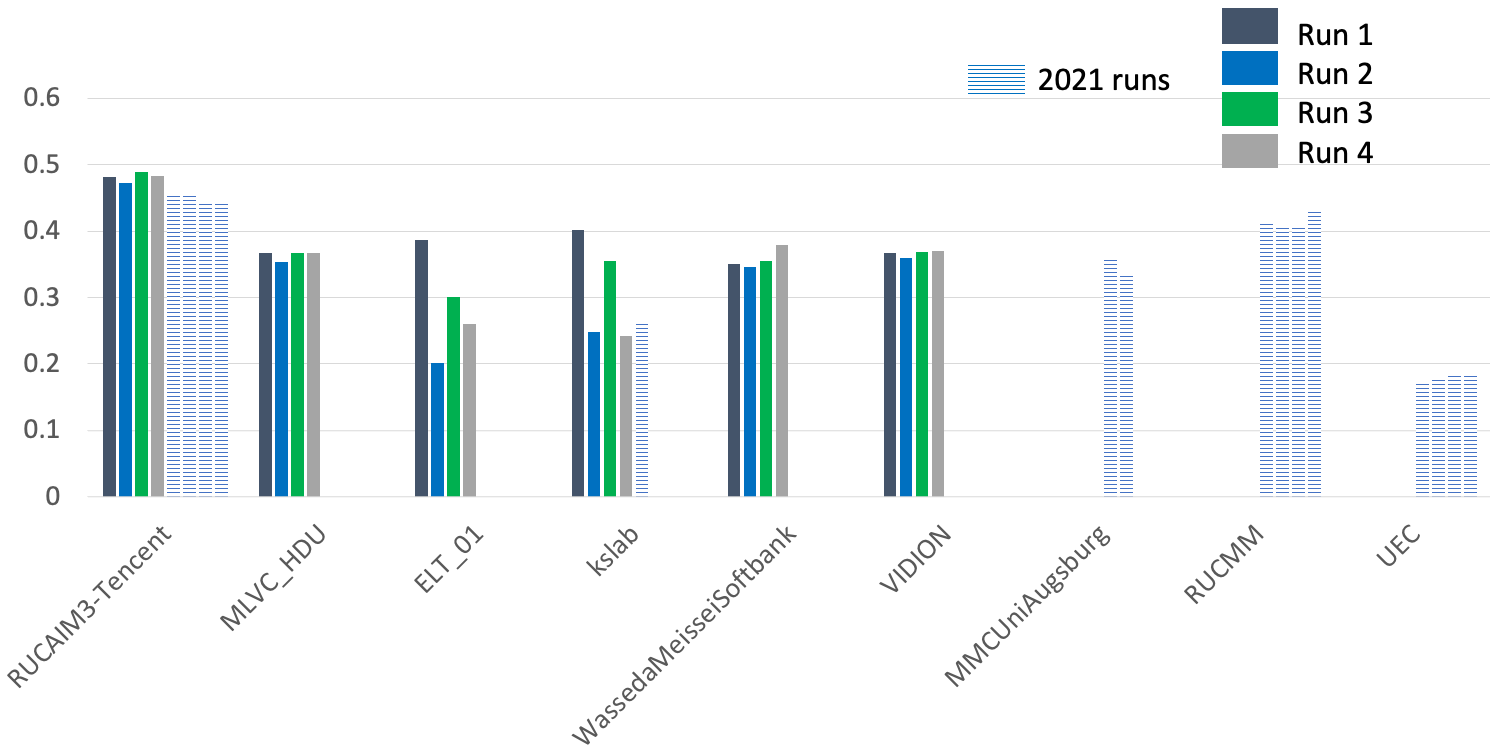}
  \caption{VTT: Comparison of all progress runs (submitted in 2021 - 2022) using the STS metric.}
  \label{fig:vtt.sts.results.p}
\end{figure}

\begin{figure}[htbp]
  \centering
  \includegraphics[width=1.0\linewidth]{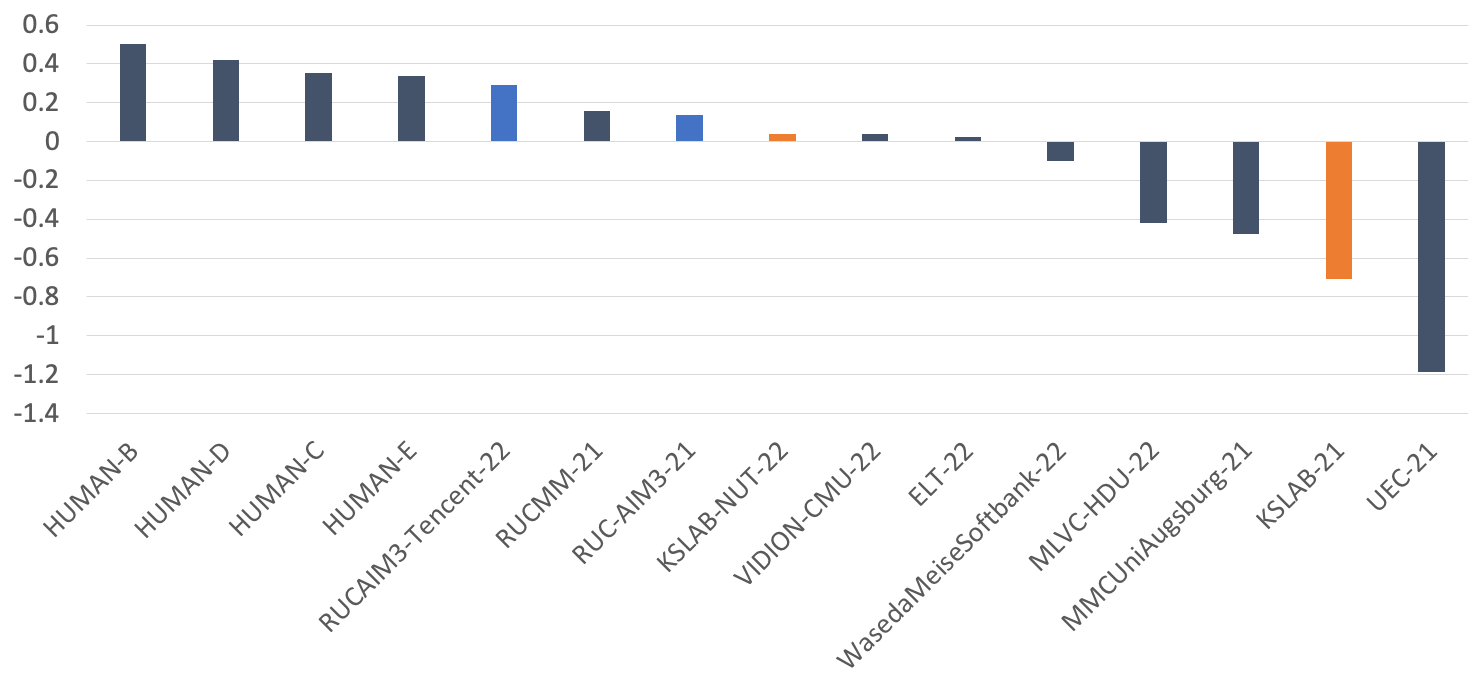}
  \caption{VTT: Average DA score per system, for progress task, after standardization per individual worker's mean and standard deviation score. }
  \label{fig:vtt.progress.da}
\end{figure}

Figures~\ref{fig:vtt.cider.results.p}, ~\ref{fig:vtt.ciderd.results.p}, ~\ref{fig:vtt.spice.results.p}, ~\ref{fig:vtt.meteor.results.p}, ~\ref{fig:vtt.bleu.results.p}, and ~\ref{fig:vtt.sts.results.p} show the automatic metrics scores for the progress subtask which evaluated runs on 300 fixed videos in 2021 and 2022. Two teams participated in both years (RUCAIM3-Tencent and kslab). In general, the 2022 systems are better than their 2021 versions. With respect to the three teams who only joined this progress task in 2021, we can see that the two teams namely RUCMM and MMCUniAugsburg 2021 systems are competitive compared to other 2022 systems. Similar to the main task, the STS metric here shows close performance of the top run from each team. Finally, the DA experiment was also conducted on the progress subtask videos for the primary runs submitted in 2021 and this year. Figure~\ref{fig:vtt.progress.da} shows the results where it can be seen that RUCAIM3-Tencent and kslab teams who participated in both years performed better in 2022 based on the human evaluation in DA metric.

\subsubsection{Task observations and conclusions}

The VTT task continues to have healthy participation. Given the challenging nature of the task, and the increasing interest in video captioning in the computer vision community, we hope the dataset resources generated from the task as well as algorithms by teams inspire more improvements for the task in the future. 

This was the first year using the V3C1 test data following two years of V3C2. The progress subtask concludes that this year's systems are better than previous year's. We hope more existing teams submit again next year to measure progress over 3 years. High correlation exists between all automatic metrics. Audio features were used by only two runs which also proved to be beneficial. Based on the DA evaluation, human captions are still better than the best automatic system. In general, the metrics reported higher scores compared to 2021 (caution: different testing dataset, but same domain). With increasing interest in video captioning, participants have a number of open datasets available to train their systems.

As a general high-level overview of the participant systems, we can summarize each system as follows: 
the RUCAIM3-Tencent team leveraged a vision-language pre-training model pre-trained on large-scale image-text datasets for video captioning such as BLIP (Bootstrapping Language-Image Pre-training). They employed an effective pseudo-label-based data augmentation method to expand the fine tuning data and designed re-ranking strategies to automatically select better descriptions from the candidates.

The Elyadata team explored different transformer combinations and types of data. They compared the use of an image captioning model versus the use of multiple frames using spatial features only. In addition, the introduction of spatiotemporal transformer was explored as an alternative to the spatial frame encoder. All models are based on the BLIP model.

The kslab team built a system consisting of three phases: frame extraction from the video, captioning for each frame, and aggregation of the captions. They adopted the NIC (Neural Image Caption generator) and OFA (One For All) models for the still image captioning phase, and sentence-based and word-by-word Lexrank methods for the sentence aggregation phase.

The MLVC\_HDU team adopted a Semantic Alignment Network (SAN), which attempts to establish a mapping relation between generated words and video frames by attention mechanism and then to decode these video frames in predicting the next word. SAN learns to capture the most discriminative phrase of the partially decoded caption and also the mapping that aligns each phrase with the relevant frames.

The VIDION team leveraged a fine-tuned pretrained BLIP model trained on COCO, Visual Genome, three web datasets, Conceptual 12M, SBU Captions, and LAION. They combined BLIP with a model that they trained on AudioSet (human-labeled dataset for audio events) to account for the audio modality of the data. Their runs based on audio modality performed slightly better than runs without audio models incorporated.

The Waseda Team conducted a pretraining on the VATEX dataset based on SwinBERT which is a model comprising Video
SwinTransformer for video feature extraction and Transformer Encoder for the decoder. Finally, they applied a fine-tuning step using the TRECVID-VTT dataset to use it for their official run generation.

For detailed information about the approaches and results for individual teams' performance and runs, we refer the reader to the site reports \cite{tv22pubs} in the online workshop notebook proceedings.

\subsection{Activities in Extended Video}
The Activities in Extended Video (ActEV) evaluation series are designed to accelerate the development of robust, multi-camera, automatic human activity detection systems for forensic and real-time alerting applications. In this evaluation, an activity is defined as \say {one or more people performing a specified movement or interacting with an object or group of objects (including driving)}, while an instance indicates an occurrence (time span of the start and end frames) associated with the activity. 
This year's ActEV Self-Reported Leaderboard (SRL) Challenge is based on the Multiview Extended Video with Activities (MEVA) Known Facility (KF) dataset \cite{MEVAdata}. The large-scale MEVA dataset is designed for activity detection in multi-camera environments. Previous ActEV task evaluations in 2021 and 2020 used the VIRAT dataset which had 35 target activities \cite{oh2011large}. The NIST TRECVID ActEV series were initiated in 2018 to support the Intelligence Advanced Research Projects Activity (IARPA) Deep Intermodal Video Analytics (DIVA) Program.

\par The TRECVID 2018 ActEV (ActEV18) evaluated system detection performance on 12 activities for the self-reported evaluation and 19 activities for the leaderboard evaluation using the VIRAT V1 and V2 datasets \cite{TrecVIDActev18}. 
For the self-reported evaluation, the participants ran their software on their hardware and configurations and submitted the system outputs with the defined format to the NIST scoring server. 
\par The ActEV18 evaluation addressed two different tasks: 1) identify a target activity along with the time span of the activity (AD: activity detection), 2) detect objects associated with the activity occurrence (AOD: activity and object detection). 
\par For the TRECVID 2019 ActEV (ActEV19) evaluation, we primarily focused on  18 activities and increased the number of instances for each activity. ActEV19 included the test set from both VIRAT V1 and V2 datasets and the systems were evaluated on the activity detection (AD) task only. 
\par The TRECVID 2020 ActEV (ActEV20) SRL is based on the VIRAT V1 and V2 datasets  with 35 activities with updated names to make it easier to use the MEVA dataset to train systems for TRECVID ActEV leaderboard.
The TRECVID 2021 ActEV (ActEV21) was based on the same 35 activities as ActEV20 and on the VIRAT V1 and V2 datasets and systems are evaluated on the activity detection (AD) task only.
\par
\begin{figure}[htbp]
\begin{centering}
\includegraphics[width=3.16667in,height=2.10000in]{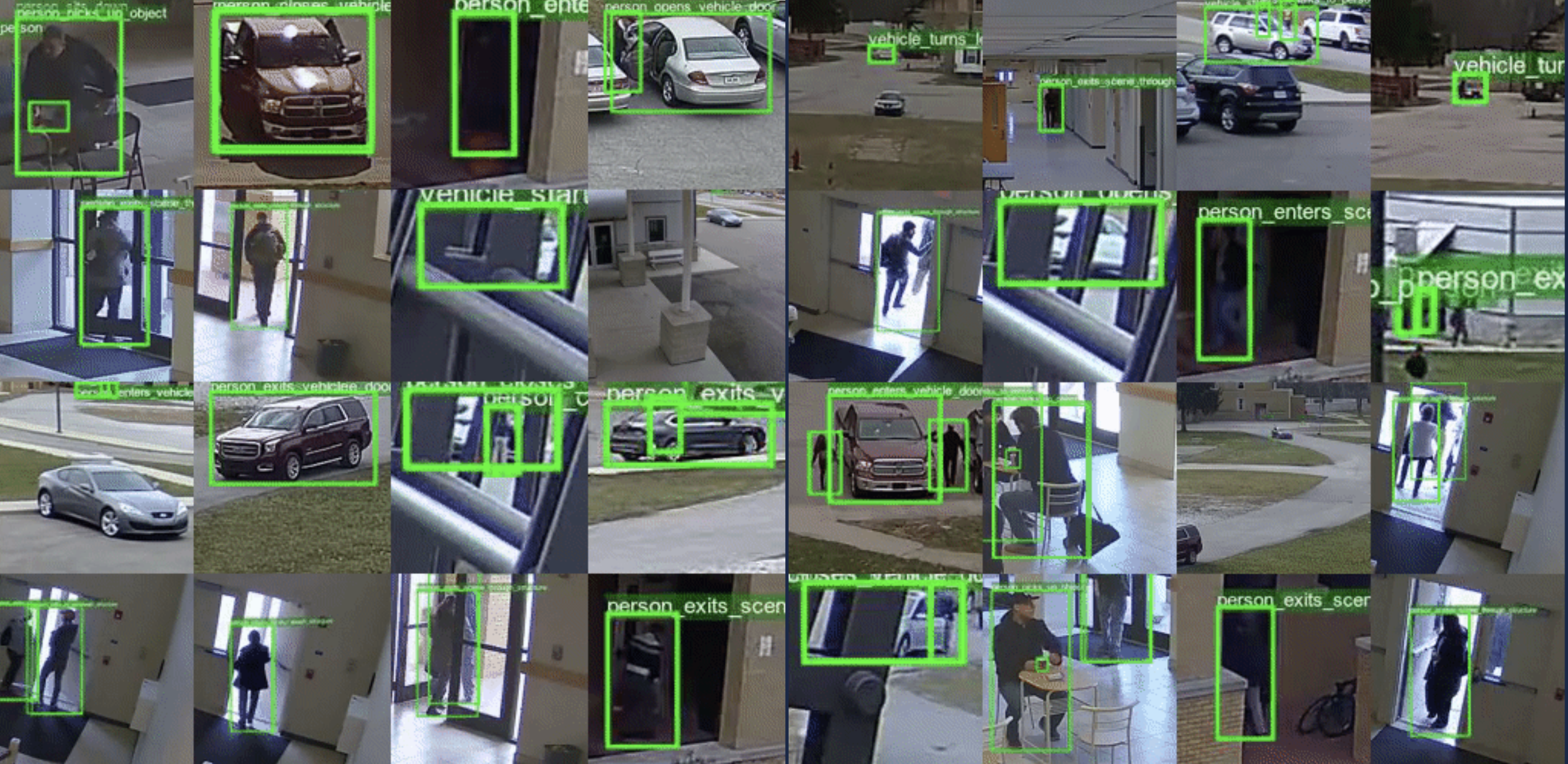}
\caption{Example of activities for MEVA dataset used ActEV SRL evaluation. IRB (Institutional Review Board): ITL-00000755}
\label{fig_actev:7}
\end{centering}
\end{figure}

 Figure \ref{fig_actev:7} illustrates an example of representative activities that were used in the TRECVID 2022 ActEV SRL based on the MEVA dataset. 

All these evaluations are primarily targeted for the forensic analysis that processes an entire corpus prior to returning a list of detected activity instances. 

\par In this section, we first discuss the task and datasets used and introduce the metrics to evaluate algorithm performance. In addition, we present the results for the TRECVID ActEV22 submissions and discuss observations and conclusions.

\subsubsection{Task and Dataset}
\par In the TRECVID'22 ActEV SRL evaluation, there are two tasks for systems; the primary task is Activity and Object Detection (AOD) and the secondary task is Activity Detection (AD)
\par \textbf{Task1:} for the AOD task, given the predefined activity classes, the objective is to automatically detect the presence of the target activity, spatio\-temporally localize all instances of the activity, and provide a confidence score indicating the strength of evidence that the activity is present. This task requires spatio\-temporal localization of objects involved in the activity (as one bounding box per frame that encompasses people, vehicles, and other objects). For a system-identified activity instance to be evaluated as correct, the activity class must be correct and the spatio\-temporal overlap must fall within a minimal requirement. The evaluation tool, ActEV\_Scorer, transforms the localization bounding boxes of both the system and reference files on the fly so that developers have the flexibility to spatially localize individual objects or a single encompassing box. 
\par \textbf{Task2:} for the AD task, given the predefined activity classes, the objective is to automatically detect the presence of the target activity, temporally localize all instances, and provide a presence confidence score indicating the strength of evidence that the activity is present. This task does not require spatio\-temporal localization of objects. For a system-identified activity instance to be evaluated as correct, the activity class must be correct and the temporal overlap must fall within a minimal requirement.

\begin{table*}[htbp]
  \centering
  \caption{A list of activity names for TRECVID ActEV SRL evaluation, there were 20 activities based on the MEVA dataset.}
 
    \begin{tabular}{ll}
    person\_closes\_vehicle\_door  & person\_closes\_vehicle\_door \\
    person\_enters\_scene\_through\_structure  & person\_enters\_scene\_through\_structure \\
    person\_enters\_vehicle  & person\_enters\_vehicle \\
    person\_exits\_scene\_through\_structure  & person\_exits\_scene\_through\_structure \\
    person\_exits\_vehicle  & person\_exits\_vehicle \\
    person\_interacts\_with\_laptop  & person\_interacts\_with\_laptop \\
    person\_opens\_facility\_door  & person\_opens\_facility\_door \\
    person\_opens\_vehicle\_door  & person\_opens\_vehicle\_door \\
    person\_picks\_up\_object  & person\_picks\_up\_object \\
    person\_puts\_down\_object & person\_puts\_down\_object \\
    \end{tabular}%
  \label{table_actev:1}%
\end{table*}%

\par The ActEV SRL evaluation is based on the Known Facilities (KF) data from the Multiview Extended Video with Activities (MEVA) dataset. The KF data was collected at the Muscatatuck Urban Training Center (MUTC) with a team of over 100 actors performing in various scenarios. The KF dataset has two parts: (1) the public training and development data and (2) SRL test dataset.
\par 
For this evaluation, we used 20 activities from the MEVA dataset and the activities were annotated by Kitware, Inc. The CVPR’22 ActivityNet ActEV SRL test dataset is a 16-hour collection of videos which only consists of Electro-Optics (EO) camera modalities from public cameras. The  ActEV SRL test dataset is the same as the one used for WACV’22 HADCV workshop ActEV SRL challenge and for the CVPR ActivityNet 2022 ActEV SRL challenge. The detailed definition of each activity and evaluation requirements are described in the evaluation plan \cite{TrecVIDActev20}. 
\par Table \ref{table_actev:1}  lists the 20 activity names for TRECVID ActEV SRL evaluation, based on the MEVA dataset. 

\subsubsection{Performance Measures}

 ActEV is not a discrete detection task unlike speaker recognition \cite{greenberg2020two} and fingerprint identification \cite{karu1996fingerprint},  it is a streaming detection task where multiple activity instances can overlap temporally or spatially and is similar to keyword spotting in audio \cite{le2014developing}. From a metrology perspective, the difference between discrete and streaming detection tasks is that non-target trials (i.e., test probes not belonging to the class) are not countable for streaming detection because the number of unique temporal/spatial instances are near infinite.  To account for this difference, the ActEV evaluations used two methods to normalize the measured false alarm performance.  The first, \say{Rate of False Alarms} ($R_{fa}$), is an instance-based false alarm measure that uses the number of video minutes as an estimate of the number of non-target trials as the false alarm denominator.  The second, \say{Time-based False Alarms} ($T_{fa}$), is a time-based false alarm measure that used the sum of non-target time as the denominator.  The two variations correspond to two views concerning the impact false alarms have on a user reviewing detections. The former is instance-based which implies the user effort would scale linearly with the detected instances and the latter time-based which implies the user effort would scale linearly with the duration of video reviewed.

 For both the AOD (primary) and AD (secondary) tasks for TRECVID'22 ActEV SRL, the submitted results are measured by Probability of Missed Detection (Pmiss) at a Rate of Fixed False Alarm ($R_{fa}$) of 0.1 (denoted Pmiss@0.1RFA). RateFA is the average number of false alarms activity instances per minute. Pmiss is the portion of activity instances where the system did not detect the activity within the required temporal (AD) and spatio-temporal (AOD) overlap requirements. Submitted results are scored for Pmiss and RateFA at multiple thresholds (based on confidence scores produced by the systems), creating a detection error tradeoff (DET) curve. 
 \par
The primary measure of performance for TRECVID ActEV21 was the normalized, partial Area Under the DET Curve ($nAUDC$) from 0 to a fixed, Rate of False Alarms ($R_{fa}$) nAUDC $R_{FA}$ value $a$ , denoted $nAUDC_a$, which is the same different than the metric used for the TRECVID ActEV20 and ActEV19 evaluations which used $T_{fa}$.  The switch to $R_{fa}$ coincided with a new experimental finding. $T_{fa}$-optimized systems tend to hyper-segment detections to maximize performance on the metrics. When evaluators reviewed the detections of top systems, the number of detections to review overwhelmed the reviewer.  Consequently, changing the primary metric to use $R_{fa}$ greatly penalized hyper fragmentation and produced systems with fewer, higher quality detections. All ActEV performance measurements were on a per-activity basis and then performance was aggregated by averaging over activities.  While presence confidence scores were used to compute performance, cross-activity presence confidence score normalization was not required nor evaluated.

\begin{figure}[!htbp]
\begin{centering}
\includegraphics[width=3.21667in,height=1.50000in]{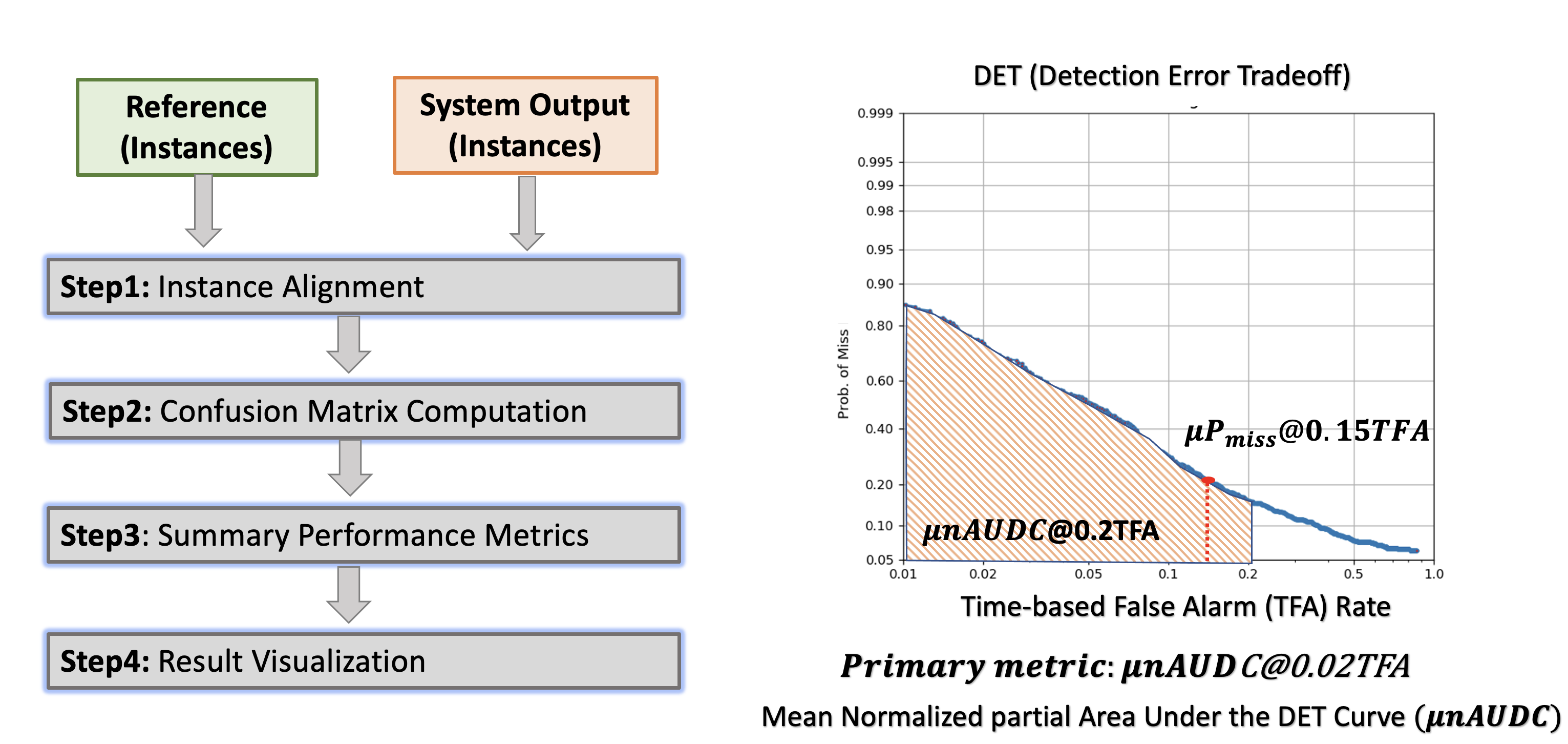}
\caption{Performance measure calculation and Detection Error Tradeoff (DET) curves}
\label{fig_actev:3}
\end{centering}
\end{figure}

Figure \ref{fig_actev:3} shows a summary of performance metric calculation. For given reference annotation and system output, the steps are 1) Align the reference activity instance with each relevant system’s instance; 2)Compute detection confusion matrix; 3)Compute summary performance metrics; and 4) Visualize the results such as DET curve shown here, which the x-axis is Time-based False Alarm (TFA) Rate and y-axis is probability of missed detection. For both the AOD (primary) and AD tasks, the submitted results are measured by Probability of Missed Detection (Pmiss) at a Rate of Fixed False Alarm (RateFA) of 0.1 (Pmiss@0.1RFA). RateFA is the average number of false alarms activity instances per minute. Pmiss is the portion of activity instances where the system did not detect the activity within the required temporal (AD) and spatio-temporal (AOD) overlap requirements. For ActEV'22 ActEV SRL evaluation primary metric was the AOD mean Normalized partial Area Under the DET Curve  \(\mu nAUDC\).

 \begin{figure*}[htbp]
\begin{centering}
\includegraphics[width=5.56667in,height=1.2000in]{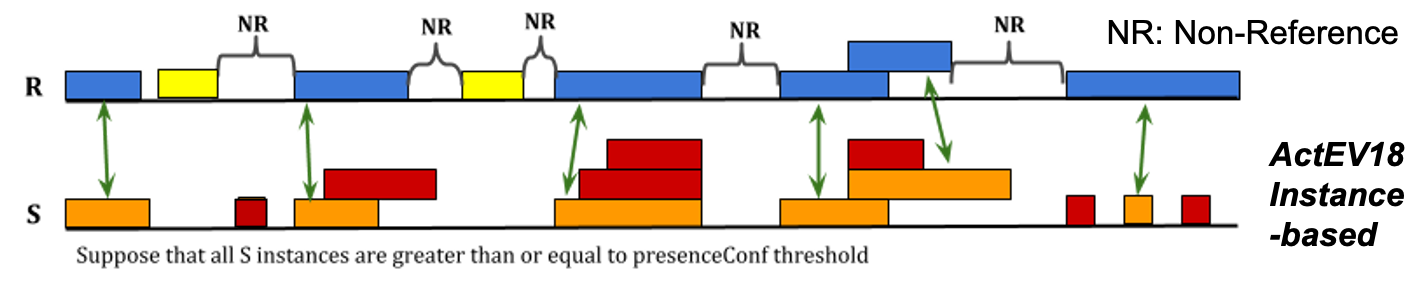}
\caption{Illustration of activity instance alignment. \(R\) is the set of reference instances and \(S\) is the set of the system instances. Green arrows connect \(R\) and \(S\) instances that are determined to be aligned and thus labelled correct detections.}
\label{fig_actev:1}
\end{centering}
\end{figure*}

 As shown in Figure \ref{fig_actev:1}, the detection confusion matrix is calculated with an alignment between reference and system output instances per target activity; Correct Detection (\(CD\)) indicates that the reference and system output instances are correctly mapped (instances marked in blue). Missed Detection (\(MD\)) indicates that an instance in the reference has no correspondence in the system output (instances marked in yellow) while False Alarm (\(FA\)) indicates that an instance in the system output has no correspondence in the reference (instances marked in red). After calculating the confusion matrix, we summarize system performance: for each instance, a system output provides a confidence score that indicates how likely the instance is associated with the target activity. The confidence scores are not used as a decision threshold. Rather, a decision threshold is applied on the scores to determine the error counts ($N_{FA}$ and $N_{miss}$).

 \par In the ActEV22 evaluation, a probability of missed detections (\(P_{\text{miss}}\)) and a rate of false alarms (\(R_{\text{FA}})\) were used and computed at a given decision threshold:
\[P_{\text{miss}}(\tau)\  = \frac{N_{\text{MD}}(\tau)}{N_{\text{TrueInstance}}}\]
\[\text{R}_{\text{FA}}(\tau)\  = \frac{N_{\text{FA}}(\tau)}{\text{VideoDurInMinutes}}\]

\noindent where \(N_{\text{MD\ }}(\tau)\) is the number of missed detections at the threshold   
\(\tau\), \(N_{\text{FA}}(\tau)\) is the number of false alarms, and \emph{VideoDurInMinutes} is the video duration in minutes. \(N_{\text{TrueInstance}}\) is the number of reference instances annotated in the sequence per activity. Lastly, the Detection Error Tradeoff (DET) curve \cite{martinDET} is used to visualize system performance. For the TRECVID ActEV18 challenge, we evaluated algorithm performance for one operating point, 
\(P_{\text{miss}}\text{\ at\ }R_{\text{FA}} = 0.1\).

To understand system performance better and to be more relevant to the use cases,  we used the normalized, partial area under the DET curve (\(nAUDC\)) from 0 to a fixed (\(R_{fa}\)) to evaluate algorithm performance.
The partial area under DET curve is computed separately for each activity over all videos in the test collection and then is normalized to the range [0, 1] by dividing by the maximum partial area. \(nAUDC_a=0\) is a perfect score. The \(nAUDC_a\) is defined as:

\[nAUDC_{a} = \frac{1}{a}\int_{x=0}^{a} P_{miss}(x)dx,  x=R_{fa}\]

\noindent where \(x\) is integrated over the set of \(R_{fa}\) and \(P_{miss}\) as defined above.

In the AOD task, a system not only detects and temporally localizes the target activity, but also spatio-temporally localizes the objects that are associated with a given activity by providing the coordinates of object bounding boxes and object presence confidence scores.

The primary metric is similar to AD, however, the instance alignment step uses an additional alignment term for object detection congruence to optimally map reference and system output instances---this is covered
in further detail in the evaluation plan \cite{TrecVIDActev22}.

For the object detection (secondary) metric, we employed the Normalized Multiple Object Detection Error (N\_MODE) described in \cite{kasturi2009framework} and \cite{bernardin2008evaluating}. N\_MODE evaluates the relative number of false alarms and missed detections for all objects per activity instance. Note that the metric is applied only to the frames where the system overlaps with the reference. The metric also uses the Hungarian algorithm to align objects between the reference and system output at the frame level. The confusion matrix for each frame \emph{t} is calculated from the confidence scores of the objects' bounding boxes, referred to as the object presence confidence threshold \(\tau\). \(\text{CD}_{t}(\tau)\) is the count of reference and system output object bounding boxes that are correctly mapped for frame t at threshold
\(\tau\). \(\text{MD}_{t}(\tau)\) is the count of reference bounding boxes not mapped to a system object bounding box at threshold \(\tau\). \(\text{FA}_{t}(\tau)\) is the count of system bounding boxes that are
not aligned to reference bounding boxes. The equation for N\_MODE follows:

\small{
\[N_{\text{MODE}\left( \tau \right)} = \sum_{t = 1}^{N_{\text{frames}}}\frac{\left( C_{\text{MD}}\times MD_{t}\left( \tau \right) + C_{\text{FA}}\times FA_{t}\left( \tau \right) \right)}{\sum_{t = 1}^{N_{\text{frames}}}N_{R}^{t}}\]}

\(N_{\text{frames}}\) is the number of frames in the sequence for the reference instance and \(N_{R}^{t}\) is the number of reference objects in frame t. For each instance-pair, the minimum N\_MODE value (minMODE)
is calculated for object detection performance and \(P_{\text{Miss}}\ \)at \(R_{\text{FA}}\) points are reported for both activity-level and object-level detections. For the activity-level detection, we used the same operating points \(P_{\text{miss}}\) at \(R_{\text{FA}} = 0.1\) and \(P_{\text{miss}}\) at
\(R_{\text{FA}} = .2\) while \(P_{\text{miss}}\) at \(R_{\text{FA}} = 0.1\) was used for the object-level detection. We used 1- minMODE for the object detection congruence term to align the instances for the target activity detection. In this evaluation, the spatial object localization (that is, how precisely systems can localize the objects) is not addressed.

\subsubsection{ActEV Results}

A total of 6 teams from academia and industry from 4 countries participated in the ActEV22 evaluation. Each participant was allowed to submit multiple system outputs and a total of 38 submissions were received. Table \ref{table_actev:4}  lists the participating teams along with results ordered by $nAUDC@0.2RFA$ scores for the top performing system per team along with $mean\_P_{miss}@.1{RFA}$ values. The top $nAUDC@0.2RFA$ performance on activity detection is by BUPT-MCPRL at 67.05\% followed by UMD at 83.0\% and Mlvc\_hdu is third at 99.22\%.

\begin{table*}[htbp]
\small
  \centering
  \caption{Summary of participants information and and results ordered by AOD, $\mu nAUDC$ values. The AOD values of $mean\_P_{miss}@.1RFA$ values along with the $nMODE@.1RFA$ are also presented. We also present the AD values of $nAUDC@.2RFA$ and $mean\_P_{miss}@.1RFA$. Each team was allowed to have multiple submissions.}
    \begin{tabular}{|l|l|r|c|c|c|c|}
    \toprule
    \multicolumn{1}{|c|}{\multirow{3}[6]{*}{Team Names}} & \multicolumn{1}{c|}{\multirow{3}[6]{*}{Organization}} & \multicolumn{3}{p{13.7em}|}{Primary Task: Activity and Object Detection} & \multicolumn{2}{p{9.5em}|}{Secondary Task: Activity Detection} \\
\cmidrule{3-7}          &       & \multicolumn{3}{p{12.215em}|}{\begin{center}(AOD)\end{center}} & \multicolumn{2}{p{9.83em}|}{\begin{center}(AD)\end{center}} \\
\cmidrule{3-7}          &       & \multicolumn{1}{p{3.66em}|}{Pmiss @0.1RFA} & \multicolumn{1}{p{3.66em}|}{nMODE @0.1RFA} & \multicolumn{1}{p{3.66em}|}{nAUDC @0.2RFA} & \multicolumn{1}{p{3.66em}|}{Pmiss @0.1RFA} & \multicolumn{1}{p{3.66em}|}{nAUDC @0.2RFA} \\
    \midrule
    \multicolumn{1}{|p{7.em}|}{BUPT\_MCPRL} & \multicolumn{1}{p{16.em}|}{Beijing University of Posts and Telecommunications, China} & \multicolumn{1}{c|}{0.6309} & 0.0538 & 0.6705 & 0.5805 & 0.6231 \\
    \midrule
    \multicolumn{1}{|p{7.em}|}{UMD} & \multicolumn{1}{p{16.em}|}{University of Maryland, USA } & \multicolumn{1}{c|}{0.8131} & 0.1620 & 0.8300  & 0.7789 & 0.7995 \\
    \midrule
    \multicolumn{1}{|p{7.em}|}{MLVC\_hdu} & \multicolumn{1}{p{16.em}|}{Hangzhou Dianzi University} & \multicolumn{1}{c|}{0.9921} & 0.0303 & 0.9922 & 0.9728 & 0.9732 \\
    \midrule
    \multicolumn{1}{|p{7.em}|}{Waseda\_Meisei \_Softbank} & \multicolumn{1}{p{16.em}|}{Waseda University, Meisei University, SoftBank Corporation} & \multicolumn{1}{c|}{0.9961} & 0.108 & 0.9964 & 0.9829 & 0.985 \\
    \midrule
    \multicolumn{1}{|p{7.em}|}{TokyoTech\_AIST (late)} & \multicolumn{1}{p{16.em}|}{Tokyo Institute of Technology} & \multicolumn{1}{c|}{0.9965} & 0.1827 & 0.9961 & 0.9824 & 0.983 \\
    \midrule
    \multicolumn{1}{|p{7.em}|}{M4D\_Team} & \multicolumn{1}{p{16.em}|}{Centre for Research and Technology Hellas} & \multicolumn{1}{c|}{  } &  &  & 0.9603 & 0.9639 \\
    \bottomrule
    \end{tabular}%
  \label{table_actev:4}%
\end{table*}%

\begin{figure*}[htbp]
\begin{centering}
\includegraphics[width=4.99967in,height=3.30000in]{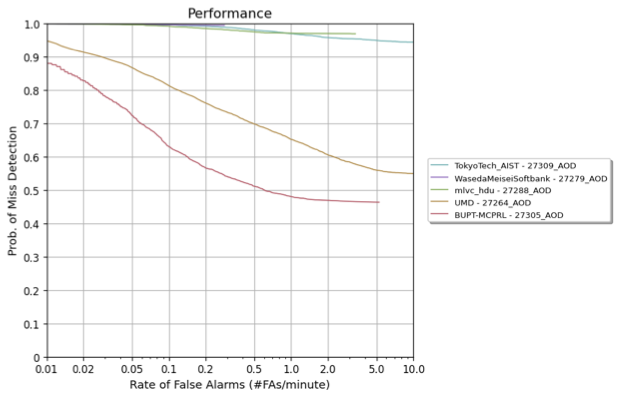}
\caption{Activity and Object Detection (AOD) DET Curve for the five teams.}
\label{fig_actev:16}
\end{centering}

\end{figure*}

Figure \ref{fig_actev:16} shows the performance based on the Activity and Object Detection (AOD) DET Curve for the 5 teams.
The x-axis is the Rate of False Alarms; The y-axis is the Prob.of Missed Detection
 and a smaller value is considered better performance. We observed that the new low for $mean\_P_{miss}@.1RFA$ of 6.7\% for team BUPT.

\begin{figure*}[htbp]
\begin{centering}
\includegraphics[width=4.9767in,height=3.00000in]{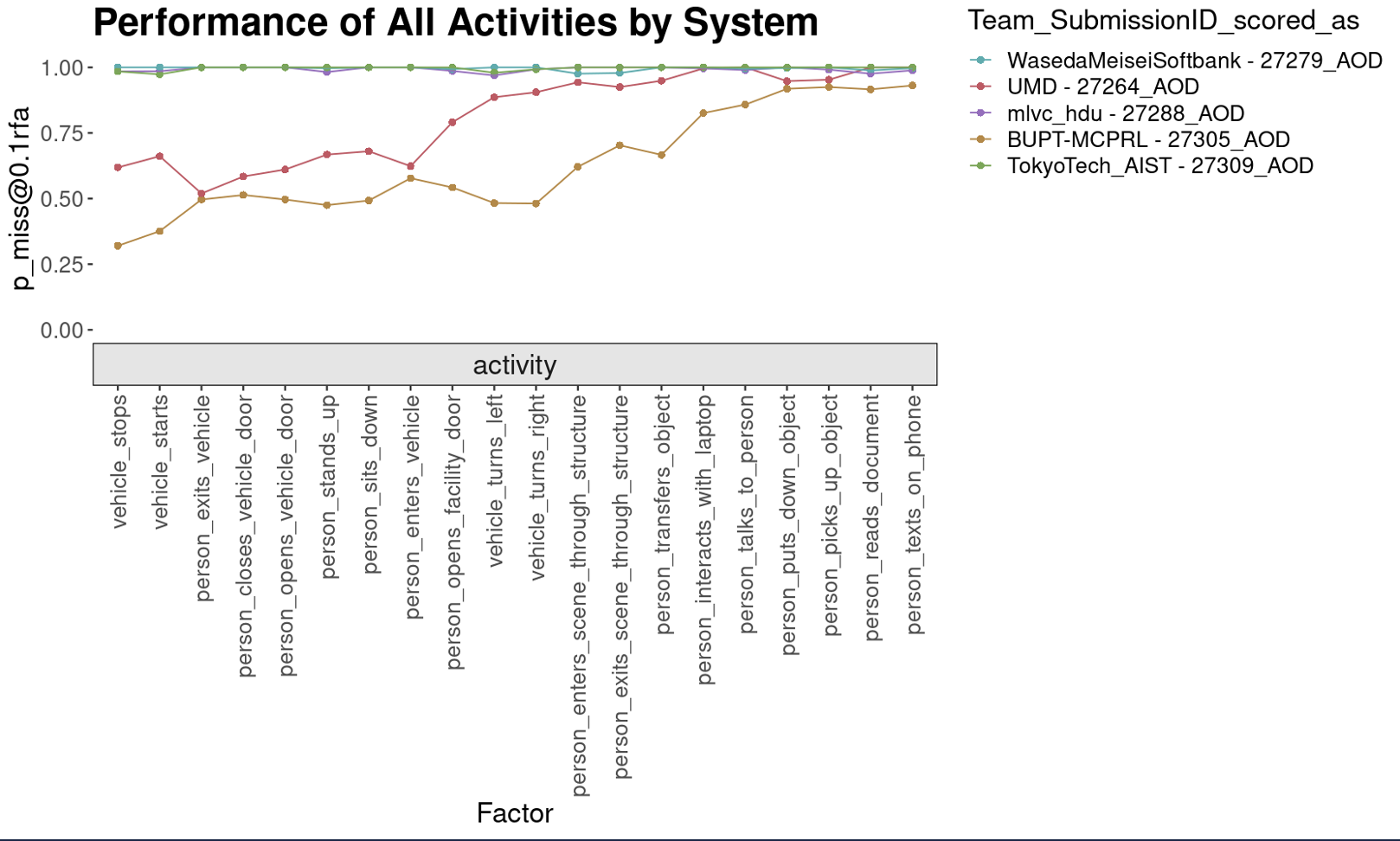}
\caption{The AOD Activity Specific Performance for the five teams}
\label{fig_actev:4}
\end{centering}
\end{figure*}

Figure \ref{fig_actev:4} shows the AOD performance for all individual activities for all the teams. The x-axis shows the 20 activities and the y-axis shows the $mean\_P_{miss}@.1RFA$. The vehicles activities remain easier than people only activities and people and object interaction activities.

\begin{figure*}[htbp]
\begin{centering}
\includegraphics[width=4.6067in,height=3.60000in]{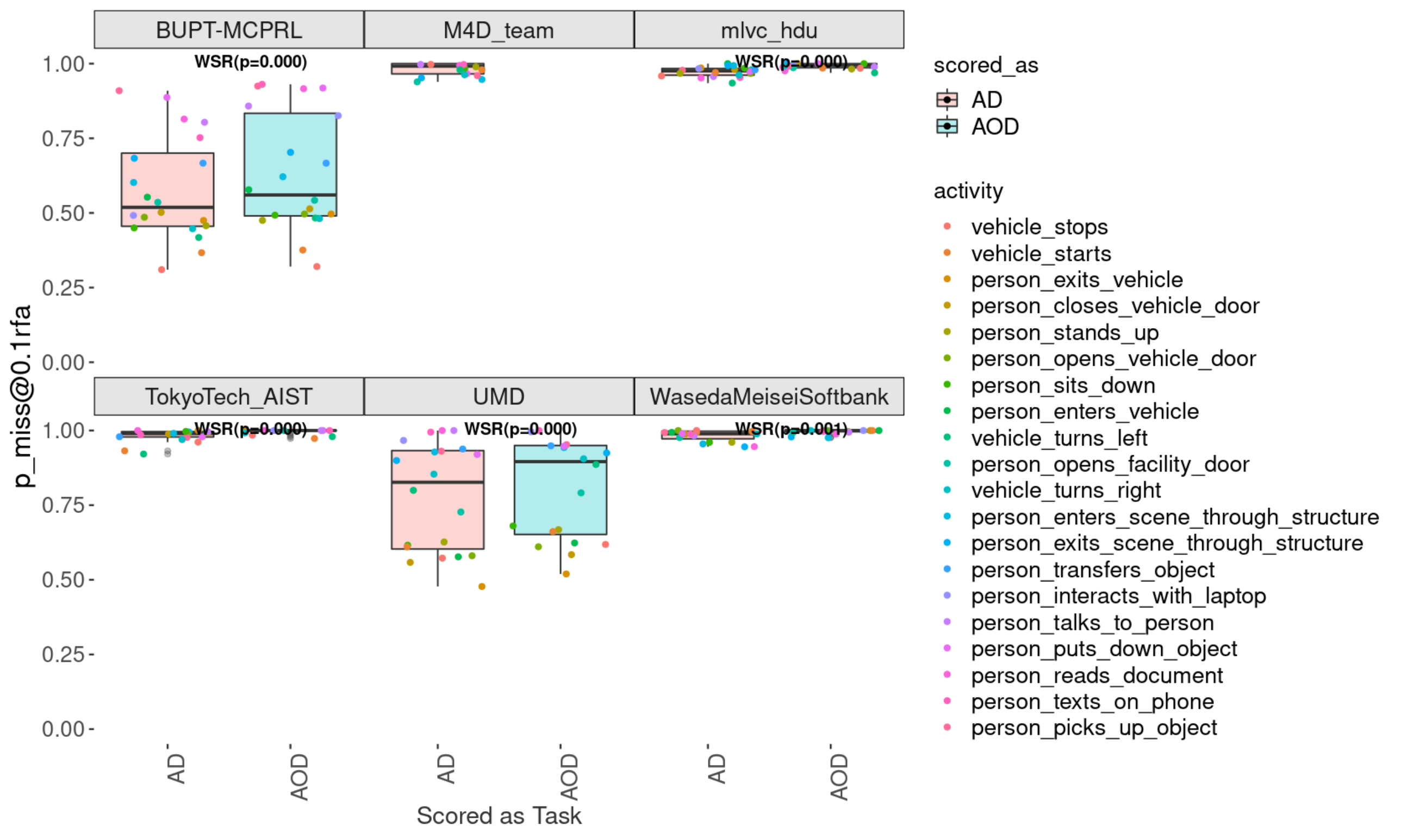}
\caption{AD vs. AOD Detection Performance}
\label{fig_actev:45}
\end{centering}
\end{figure*}

 Figure \ref{fig_actev:45} shows the AD vs. AOD Detection Performance for the six teams for all the activities. The x-axis shows the scores for AD and AOD task and y-axis shows the  $mean\_P_{miss}@.1RFA$.
As expected for every team, their AOD system has higher $mean\_P_{miss}@.1RFA$ rates than AD.

\begin{figure*}[htbp]
\begin{centering}
\includegraphics[width=4.86667in,height=3.40000in]{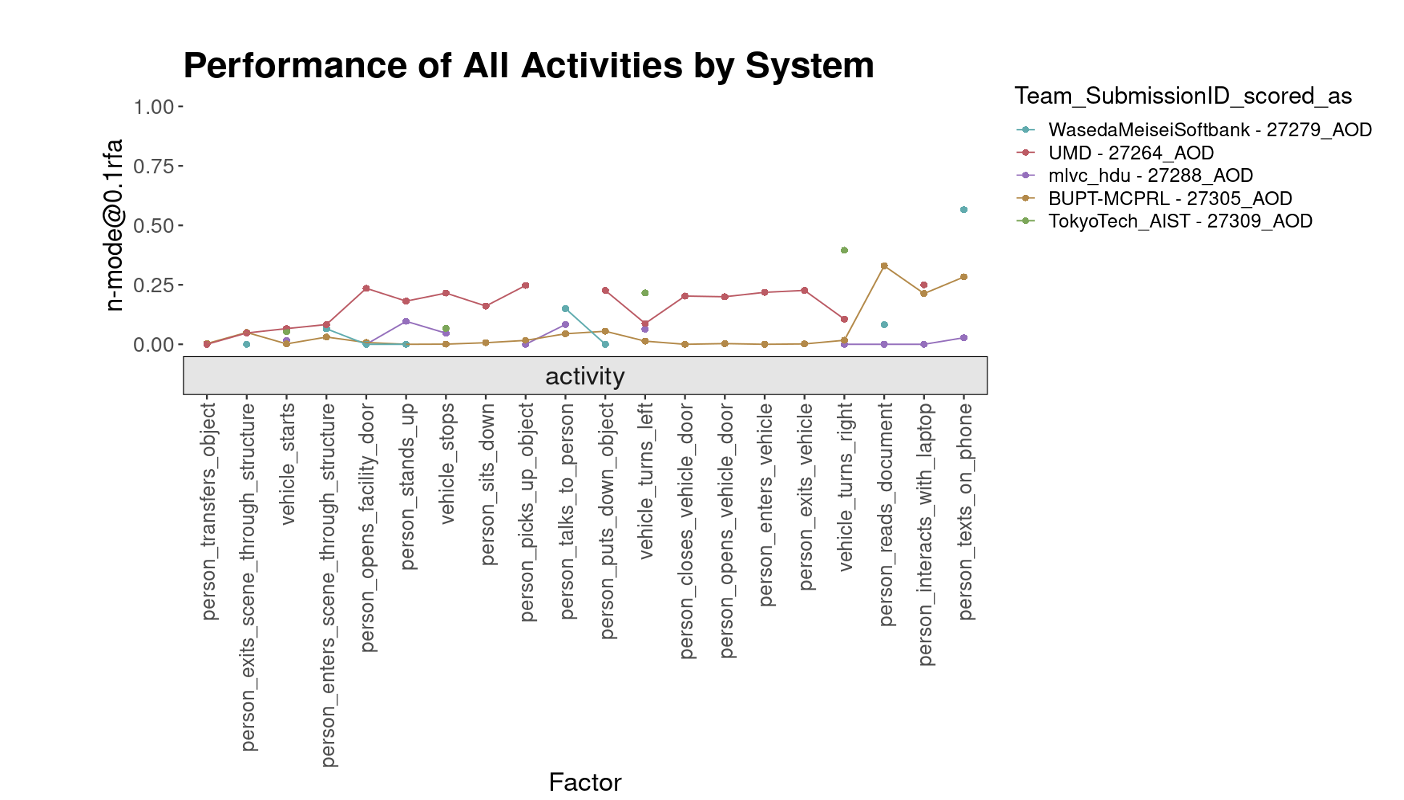}
\caption{Localization Performance for Correct AOD Instances }
\label{fig_actev:46.1}
\end{centering}
\end{figure*}

To examine the localization performance for correct AOD instances, Figure \ref{fig_actev:46.1} shows the localization performance varies across the 5 teams that participated in AOD evaluations. The x-axis shows the 20 activities and y-axis shows the localization performance $nMODE@0.1RFA$.  The missing points in the graph indicate no correct AOD detections. The BUPT-MCPRL team localizes well for most of the activities.

\subsubsection{Summary}\label{ActEV_Conclusion}

In this section, we presented the TRECVID ActEV22 evaluation task, the  performance metric and results for human activity detection for both the Activity and Object Detection and the Activity Detection tasks. We primarily focused on the activity detection task only and the time-based false alarms were used to have a better understanding of the system's behavior and to be more relevant to the use cases. The ActEV22 evaluation was based on the MEVA \cite{MEVAdata} dataset and have 20 target activities in total. Six teams from 4 countries participated in the ActEV22 evaluation and made a total of 38 submissions. We observed that, given the datasets and systems, the vehicles activities remain easier than people and people and object interaction activities. The teams MLVC\_hdu and WadsedaMeiselSoftbank participated for the first time in the ActEV evaluation. The BUPT team had the top performing system followed by the UMD team.

The TRECVID ActEV22 evaluation provided researchers an opportunity to evaluate their activity detection algorithms on a self-reported leaderboard. We hope the TRECVID ActEV22 evaluation, and the associated datasets will facilitate the development of activity detection algorithms. This will in turn provide an impetus for more research worldwide in the field of activity detection in videos.

\subsection{Movie Summarization}
An important need in many situations involving video collections (archive video search/reuse, personal video organization/search, movies, tv shows, etc.) is to summarize the video in order to reduce the size and concentrate the amount of high value information in the video track. In 2022 we began the Movie Summarization (MSUM) track in TRECVID, replacing the previous Video Summarization (VSUM) track. This track made use of a licensed movie dataset from Kinolorberedu\footnote{\url{https://www.kinolorber.com/}}, in which the goal was to summarize the storylines and roles of specific characters during a full movie.


The goals for this track are to:
\begin{enumerate}
    \item Efficiently capture important facts about certain persons during their role in the movie storyline.
    \item Assess how well video summarization and textual summarization compare in this domain.
\end{enumerate}

\subsubsection{Movie Summarization Data}
This track made use of the Kinolorberedu dataset of 10 movies, with 5 movies reserved for the training set and another 5 movies used for the testing set. Further information about movies' genres and duration are provided below in Tables \ref{trainset} and \ref{testset}.

\begin{center}
\begin{table}
\centering{
 \begin{tabular}{|p{2.55cm} | p{2.7cm} | p{1.5cm}|} 
 \hline
 \textbf{Movie} & \textbf{Genre} & \textbf{Duration}\\ [0.5ex] 
 \hline\hline
 Calloused Hands & Drama & 92\thinspace min\\
 \hline
 Liberty Kid & Drama & 88\thinspace min \\
 \hline
 Like Me & Horror / Thriller & 79\thinspace min\\
 \hline
 Losing Ground & Comedy / Drama & 81\thinspace min\\
 \hline
 Memphis & Drama & 79\thinspace min\\
 \hline
\end{tabular}
\caption{The full MSUM training set}
\label{trainset}
}
\end{table}
\end{center}

\begin{center}
\begin{table}
\centering{
 \begin{tabular}{|p{2.55cm} | p{2.7cm} | p{1.5cm}|} 
 \hline
 \textbf{Movie} & \textbf{Genre} & \textbf{Duration}\\ [0.5ex] 
 \hline\hline
 Archipelago & Drama & 110\thinspace min\\
 \hline
 Bonneville & Drama & 93\thinspace min\\
 \hline
 Chained For Life & Comedy / Drama & 88\thinspace min \\
 \hline
 Heart Machine & Drama & 84\thinspace min \\
 \hline
 Littlerock & Drama & 82\thinspace min \\
 \hline
\end{tabular}
\caption{The full MSUM testing set}
\label{testset}
}
\end{table}
\end{center}

\subsubsection{System task}
This track is comprised of two main tasks:
\begin{enumerate}
    \item Video Summary
    \item Text Summary
\end{enumerate}

\textbf{Video Summary}
Given a movie, a character, and image / video examples of that character, generate a video summary highlighting major \textbf{key-fact events} about the character (similar to TV20 \& TV21 VSUM). Video summaries are limited by a maximum summary length. See below for further details on what constitutes a key-fact event and for details on annotation and assessment. 

\textbf{Text Summary}
Given a movie, a character, and image / video examples of that character, generate a textual summary to include \textbf{key-fact events} about the character role in the movie. Textual summaries are limited by a maximum number of sentences. See below for further details on what constitutes a key-fact event and for details on annotation and assessment.

\textbf{What is a key-fact event?}
\begin{itemize}
    \item Any events that are important and critical in the character story-line.
    \item They should cover his/her role from the start to the end of the movie.
    \item Example: From the sample short movie “Super Hero” – Character: Jeremy
    \begin{itemize}
        \item Charlie bullies Jeremy.
        \item Charlie and Jeremy fight at the playground.
        \item Jeremy's mother reveals to the principal that Jeremy has a terminal illness.
        \item Jeremy gets admitted to the hospital.
        \item Jeremy passes away.
    \end{itemize}
\end{itemize}
A key-fact event regarding a character does not necessarily require that character to be visible in the scene. In the above example `Super Hero', In one scene Jeremy's mother revealed to the principal that Jeremy had a terminal illness. This would clearly count as key-fact regarding Jeremy even though he was not present in that scene.

The purpose of this task was to summarize the important key-facts for a character. As such, this is different from a movie trailer. Key events should appear in the order in which they become apparent in the movie, and should ideally capture that characters story-line.

The number of allowed key facts is limited per movie and character. One of the major challenges of the task is to separate major key facts from non consequential things. One example could be: `Daryl broke up with his girlfriend over breakfast' is more likely to be a major key fact than `Daryl had eggs and toast for breakfast'.

\subsubsection{Topics (Characters to Summarize)}
Each topic consisted of a movie, the character to summarise the key-fact events for, and a set of image/video examples of that character. For video summaries, a maximum summary time (in seconds) was specified for each character. While for text summaries, the maximum number of sentences was specified for each character as well. A sentence for text summary could be either a keyfact (the focus of the task), or a filler sentence. The maximum number of sentences a run could submit for a given character includes all keyfacts and filler sentences. 

For selection of queries, human assessors watched movies from the test set, then marked out all key facts for either 1 or 2 main characters from each movie. Key-facts were later double-checked and spell checked to remove any clear and obvious errors. Full queries and their maximum length are listed in Table \ref{msumqueries}.

\begin{table*}[h]
\centering
\caption{Movie summarization queries and specifics}
\label{msumqueries}
 \begin{tabular}{|p{3.6cm}|p{2.6cm}|p{2.8cm}|p{2.6cm}|} 
 \hline
 \textbf{Movie} & \textbf{Character} & \textbf{Max. Length} & \textbf{Max. Lines} \\
 \hline\hline
 Archipelago & Cynthia & 190 seconds & 38 \\ 
 \hline
 Archipelago & Edward & 140 seconds & 28 \\
 \hline
 Bonneville & Arvilla & 190 seconds & 38 \\
 \hline
 Chained for Life & Mabel & 130 seconds & 26 \\
 \hline
 Heart Machine & Cody & 160 seconds & 32 \\
 \hline
 Heart Machine & Virginia & 110 seconds & 22 \\
 \hline
 Littlerock & Atsuko & 190 seconds & 38 \\
 \hline
 Littlerock & Cory & 160 seconds & 32 \\
 \hline
\end{tabular}
\end{table*}

\subsubsection{Evaluation}

\begin{figure*}[!htbp]
\begin{center}
\includegraphics[height=3.3in,width=6.5in]{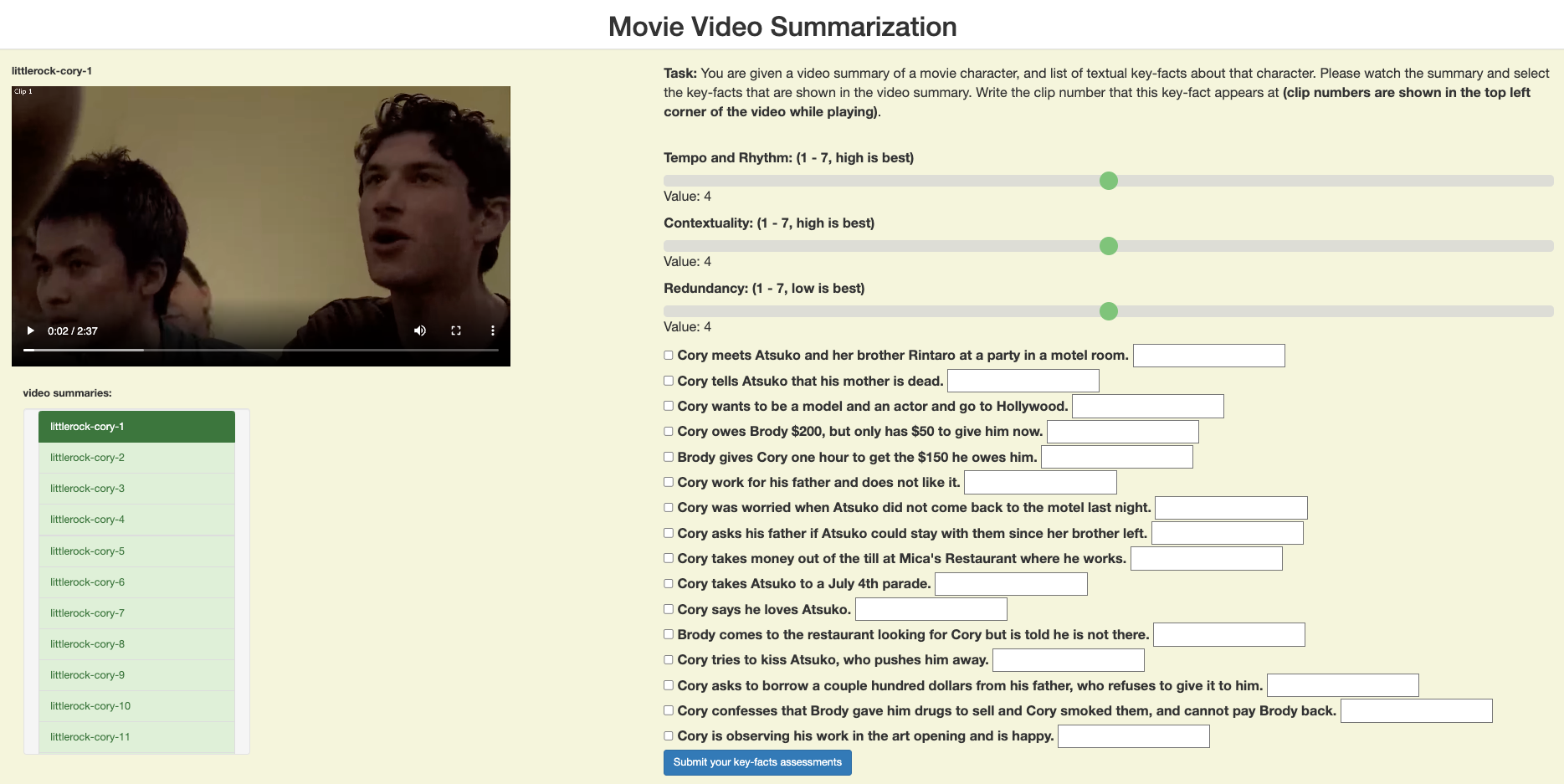}
\caption{Video summary assessment tool}
\label{vidsum}
\end{center}
\end{figure*}

\textbf{Video Summary evaluation:}
For evaluation of video summaries, assessors watched the submitted summaries. Assessors were also provided with a list of key-facts as provided by a different set of earlier assessors for this task. For any matching key-facts, assessors marked the matching key-fact and provided the clip number from the submitted summary for it to be independently verified later. Subjective ratings for tempo, contextuality, and redundancy were also provided. Figure \ref{vidsum} shows a screenshot of the tool used for assessments of video summaries.

\begin{figure*}[!htbp]
\begin{center}
\includegraphics[height=3.3in,width=6.5in]{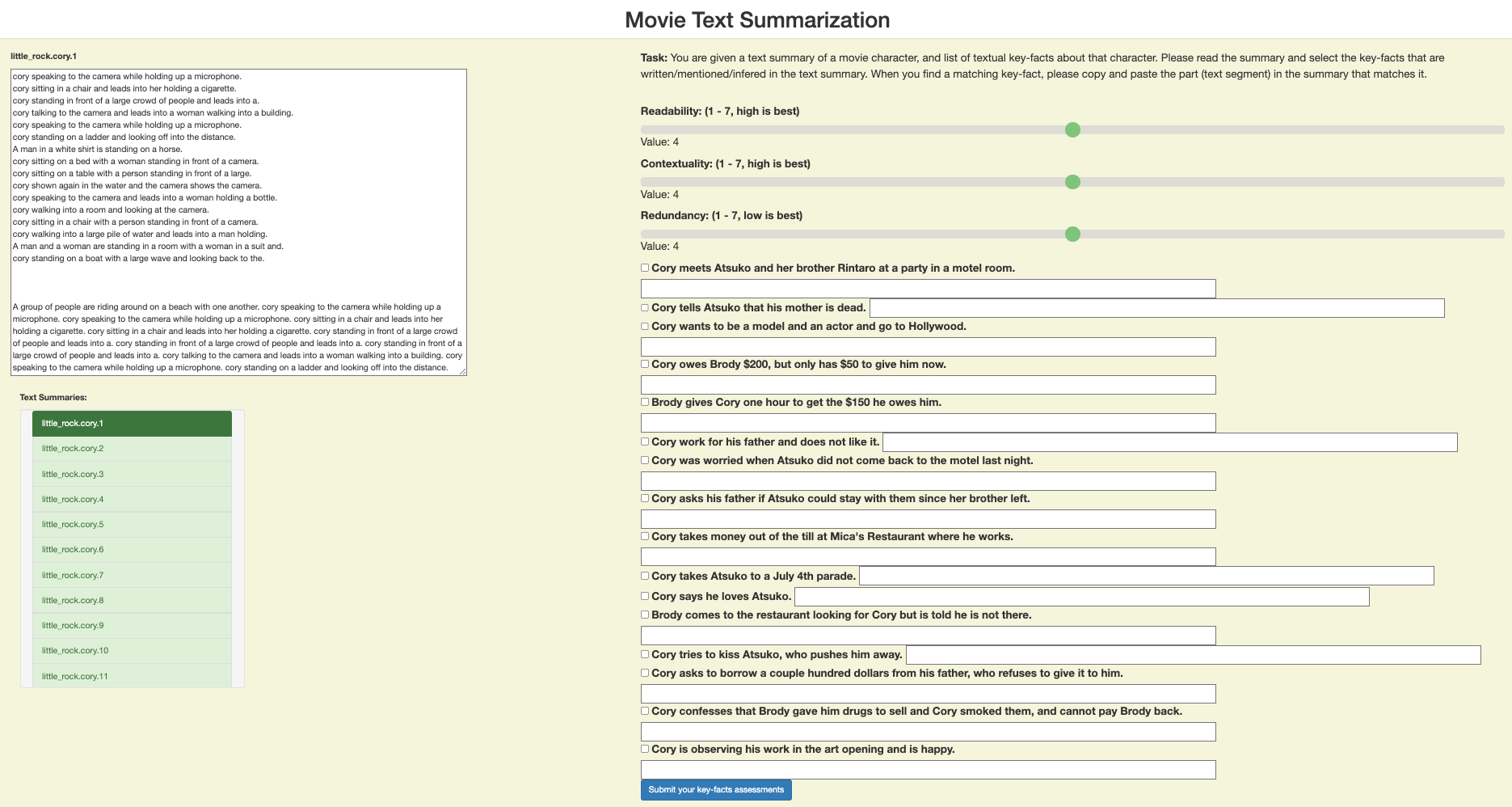}
\caption{Textual summary assessment tool}
\label{textsum}
\end{center}
\end{figure*}
    
\textbf{Textual Summary evaluation:}
For evaluation of text summaries, assessors first read a list of marked key-fact sentences in the submitted test summary which were indicated by being marked `k' for key-fact on the xml submissions. These were compared with the list key-facts for each query provided by a different set of earlier assessors. Matching key-facts were marked by assessors and the matching submitted sentence was provided so that these could be later verified. Following this, assessors then read the entire submission comprised of both key-facts and filler sentences. Subjective ratings for readability, contextuality, and redundancy were then provided. Figure \ref{textsum} shows a screenshot of the tool used for assessments of text summaries.

\subsubsection{Metrics}
Scores for this task were calculated by assigning marks to correct key-facts and to subjective attributes such as readability and contextuality. The objective-all rating refers to the percentage of correct key-facts in the submitted textual or video summary. For example, 5 key facts retrieved out of a possible 13 would give an objective-all score of 0.385. Precision refers to the number of correctly retrieved key facts divided by correct key-facts + incorrectly retrieved key-facts, i.e.\ \textit{p} = (correct / (correct + false)). 

Tempo is a video specific subjective score which rates how well a video summary comes together. Readability is the textual specific alternative which rates the overall readability of a text summary. Contextuality measures how well context is provided for important information within the summary. Redundancy measures content considered unnecessary or superfluous. All three subjective ratings are measured on a scale of 1 to 7, with 7 being best except for redundancy. All three subjective ratings are combined into an overall subjective-all score.

\subsubsection{Results - Video}
\begin{figure}[htbp]
\begin{center}
\includegraphics[height=2.5in,width=3in,angle=0]{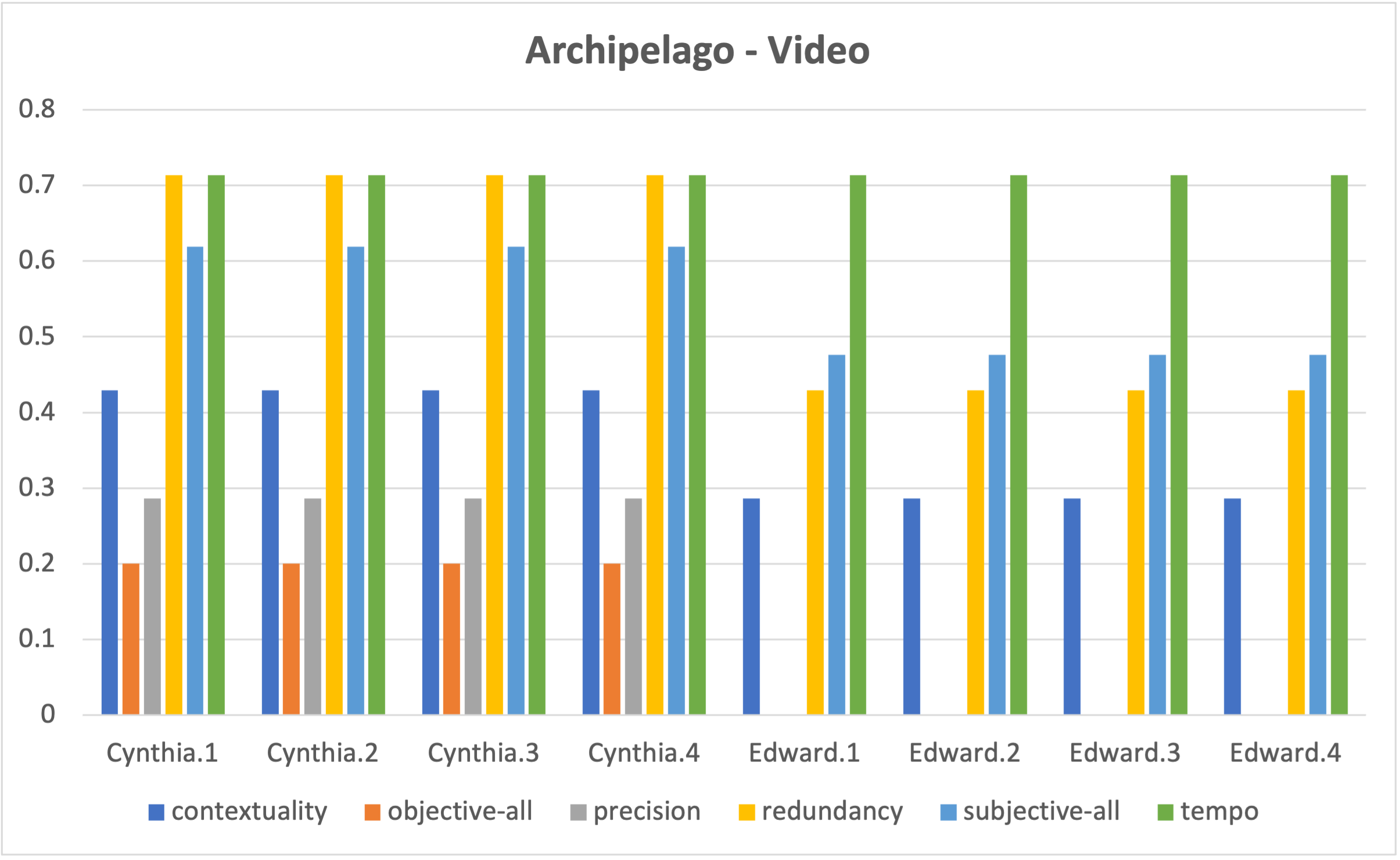}
\caption{MSUM video summarization: Archipelago}
\label{msum.arch.vid}
\end{center}
\end{figure}

Figure \ref{msum.arch.vid} shows the video summarization task results for the movie \textit{Archipelago}. Best results are seen for the character Cynthia. Subjective scores for tempo and redundancy are quite good, however objective scores of 0.2 are quite poor. There are no objective scores for the character Edward.

\begin{figure}[htbp]
\begin{center}
\includegraphics[height=2.5in,width=3in,angle=0]{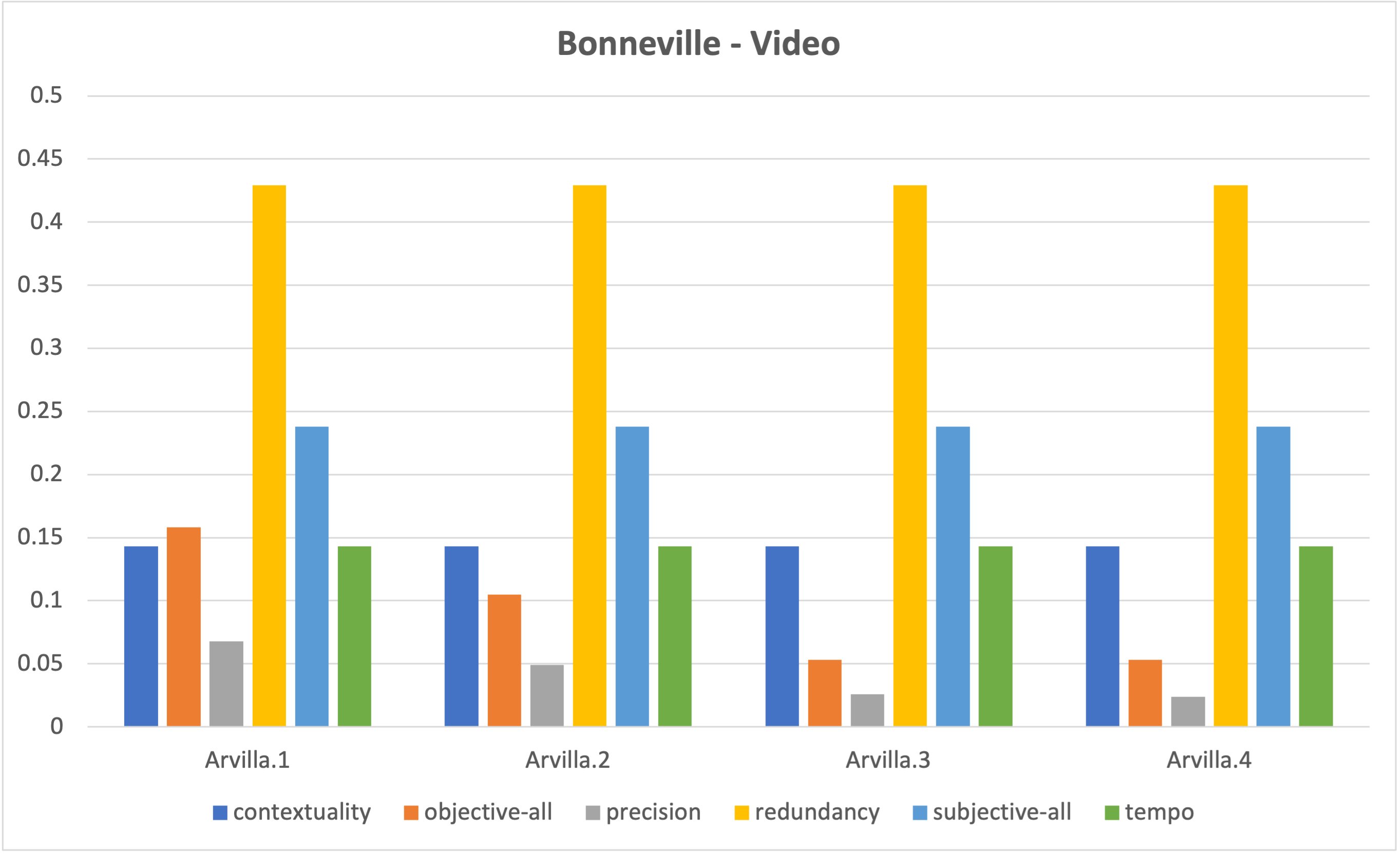}
\caption{MSUM video summarization: Bonneville}
\label{msum.bville.vid}
\end{center}
\end{figure}

Figure \ref{msum.bville.vid} shows the video summarization task results for the movie \textit{Bonneville}. There was only one character for summarization in this movie, Arvila. Generally, both subjective and objective scores for this movie and character are quite poor.

\begin{figure}[htbp]
\begin{center}
\includegraphics[height=2.5in,width=3in,angle=0]{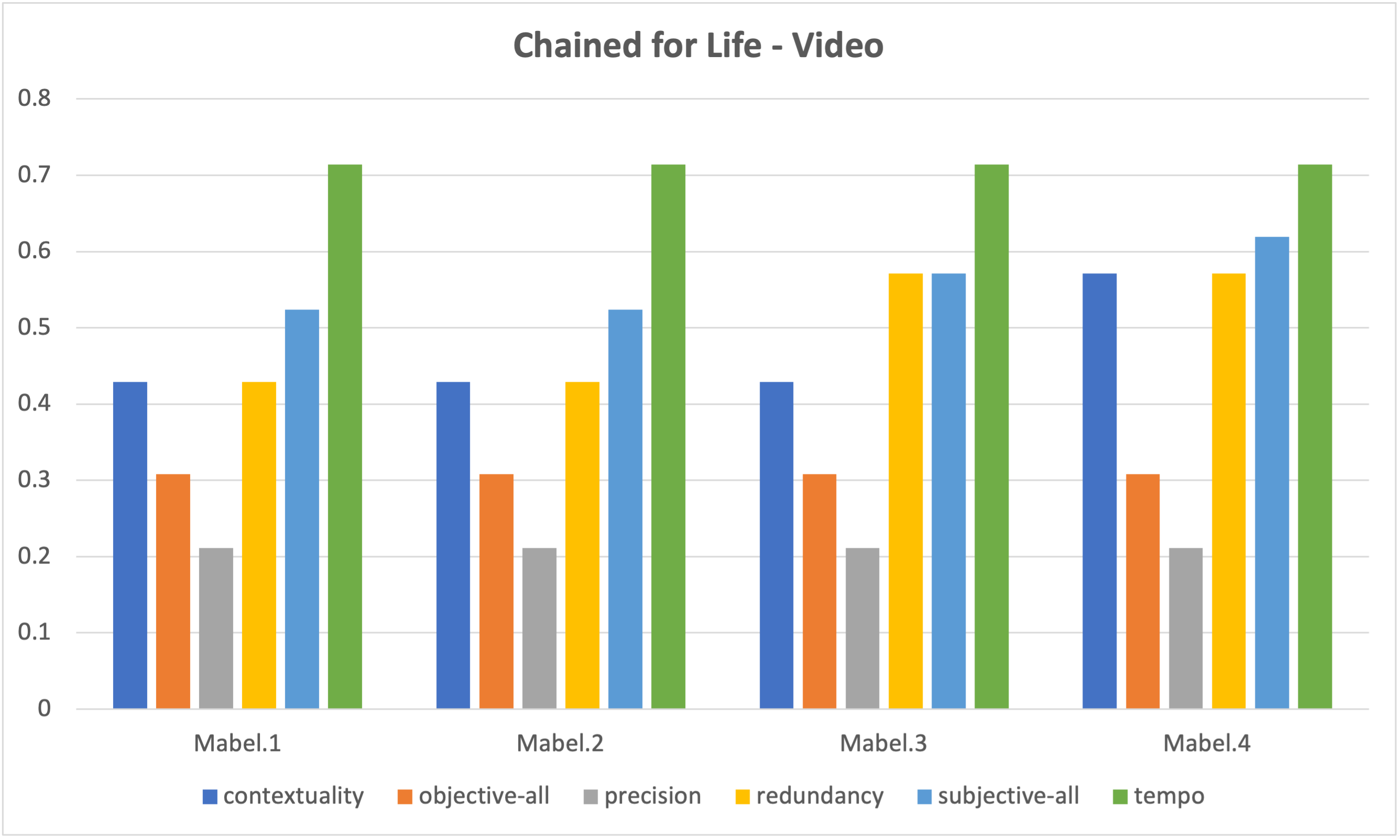}
\caption{MSUM video summarization: Chained for Life}
\label{msum.cfl.vid}
\end{center}
\end{figure}

Figure \ref{msum.cfl.vid} shows the video summarization task results for the movie \textit{Chained for Life}. This movie also contained just one character for summarization, Mabel. Subjective scores were not as high as for Archipelago - Cynthia, however objective scores are slightly better.

\begin{figure}[htbp]
\begin{center}
\includegraphics[height=2.5in,width=3in,angle=0]{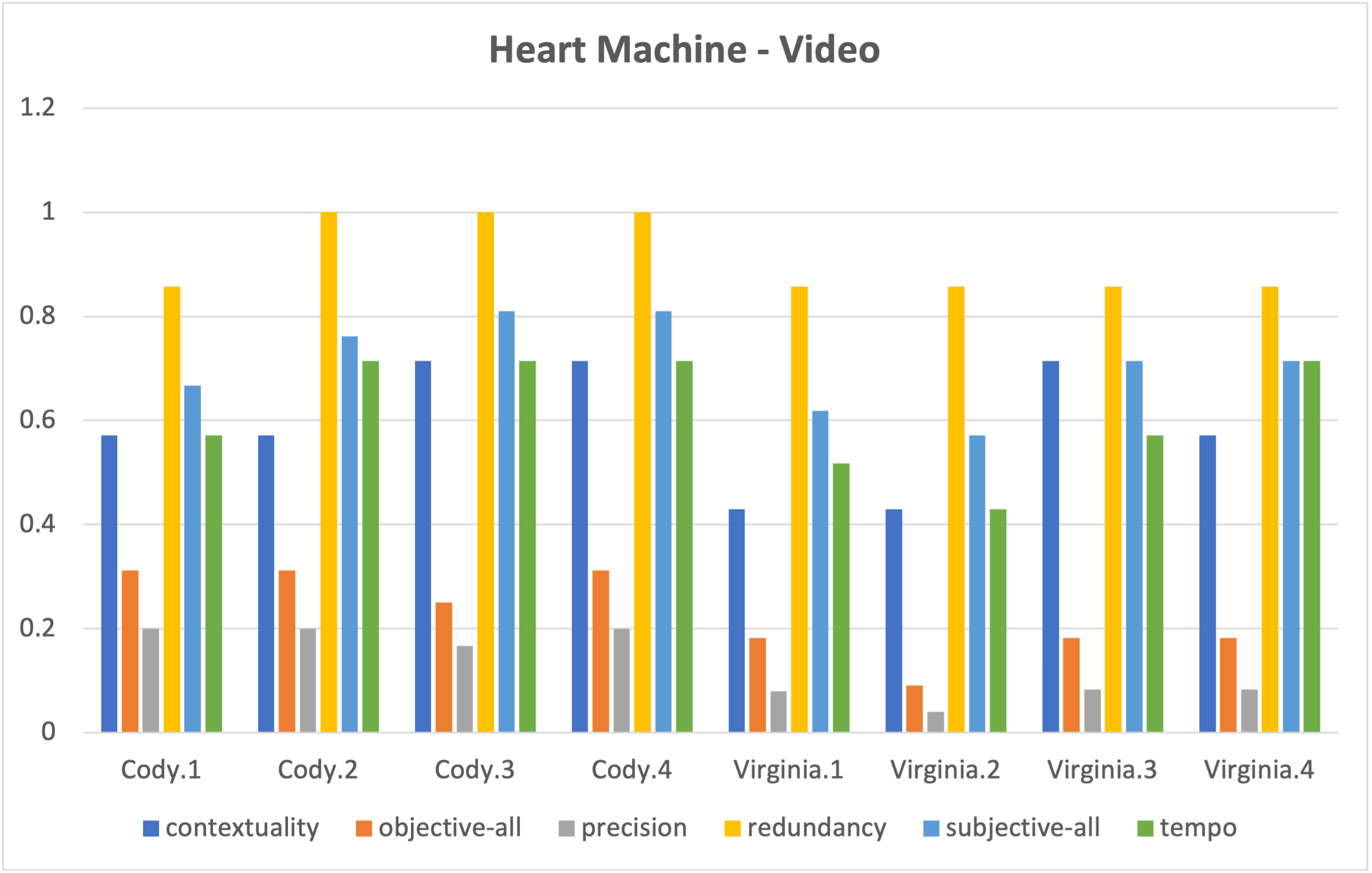}
\caption{MSUM video summarization: Heart Machine}
\label{msum.hm.vid}
\end{center}
\end{figure}

Figure \ref{msum.hm.vid} shows the video summarization task results for the movie \textit{Heart Machine}. This movie contained two characters for summarization, Cody and Virginia. Cody performs slightly better than Virginia, however there is not much difference between characters. Cody achieves perfect redundancy scores for three of the four runs. Objective scores are not great for either character, however Cody scores slightly better.

\begin{figure}[htbp]
\begin{center}
\includegraphics[height=2.5in,width=3in,angle=0]{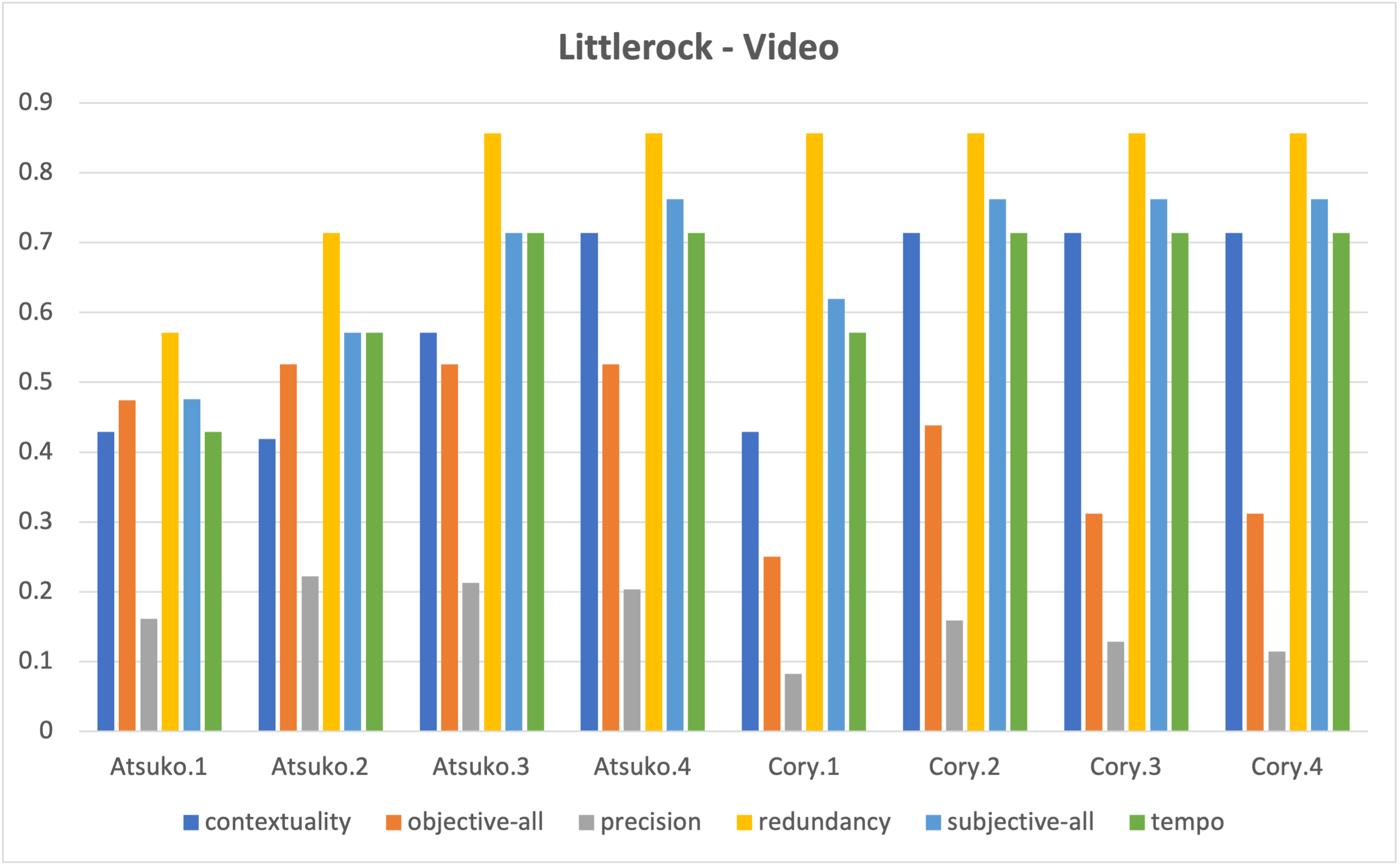}
\caption{MSUM video summarization: Littlerock}
\label{msum.lr.vid}
\end{center}
\end{figure}

Figure \ref{msum.lr.vid} shows the video summarization task results for the movie \textit{Littlerock}. This movie contained two characters for summarization, Atsuko and Cory. Atsuko performs best on objective measures while Cory performs best on subjective measures. Scores for objective measures were generally better in this movie. Cory achieves very good all round subjective scores while underachieving on objective scores in comparison to Atsuko.

\subsubsection{Results - Text}
\begin{figure}[htbp]
\begin{center}
\includegraphics[height=2.5in,width=3in,angle=0]{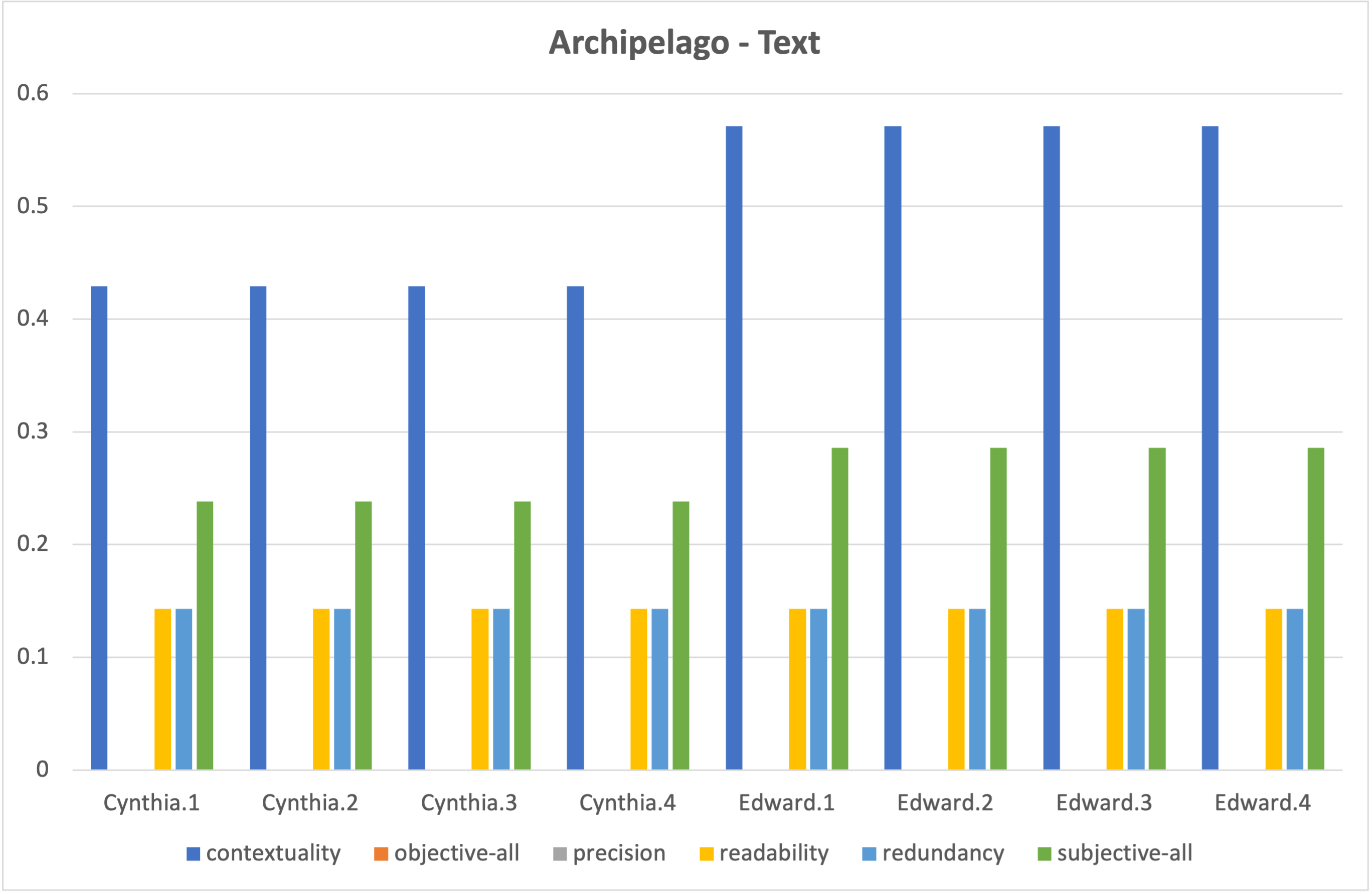}
\caption{MSUM text summarization: Archipelago}
\label{msum.arch.text}
\end{center}
\end{figure}

Figure \ref{msum.arch.text} shows the text summarization task results for the movie \textit{Archipelago}. Best results are seen for the character Edward, however these are only for subjective scores. Neither character achieves any marks for objective scores.

\begin{figure}[htbp]
\begin{center}
\includegraphics[height=2.5in,width=3in,angle=0]{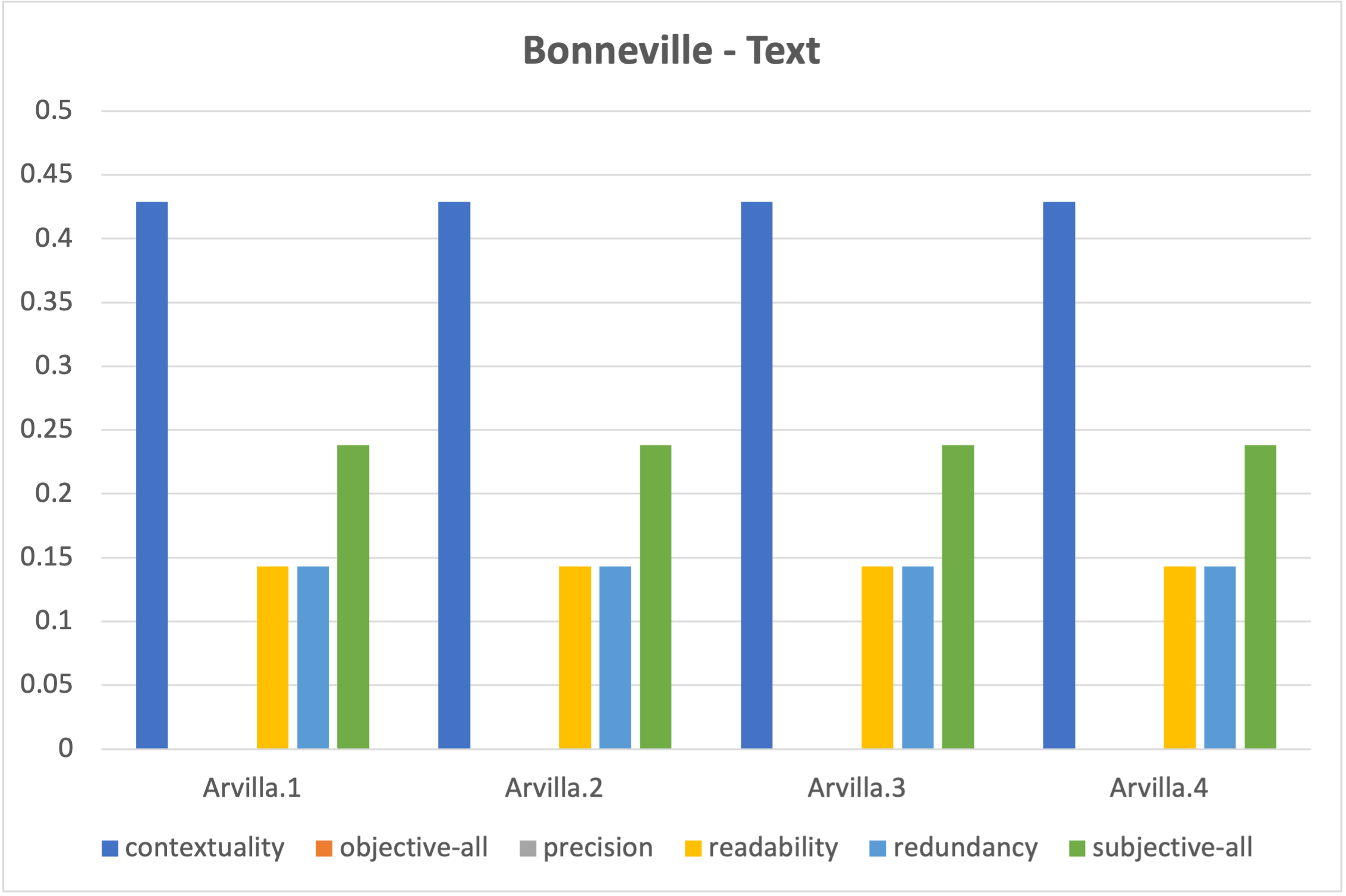}
\caption{MSUM text summarization: Bonneville}
\label{msum.bville.text}
\end{center}
\end{figure}

Figure \ref{msum.bville.text} shows the text summarization task results for the movie \textit{Bonneville}. There was only one character for summarization in this movie, Arvila. Generally, subjective scores for this movie and character are quite poor, while there are zero marks for objective scores.

\begin{figure}[htbp]
\begin{center}
\includegraphics[height=2.5in,width=3in,angle=0]{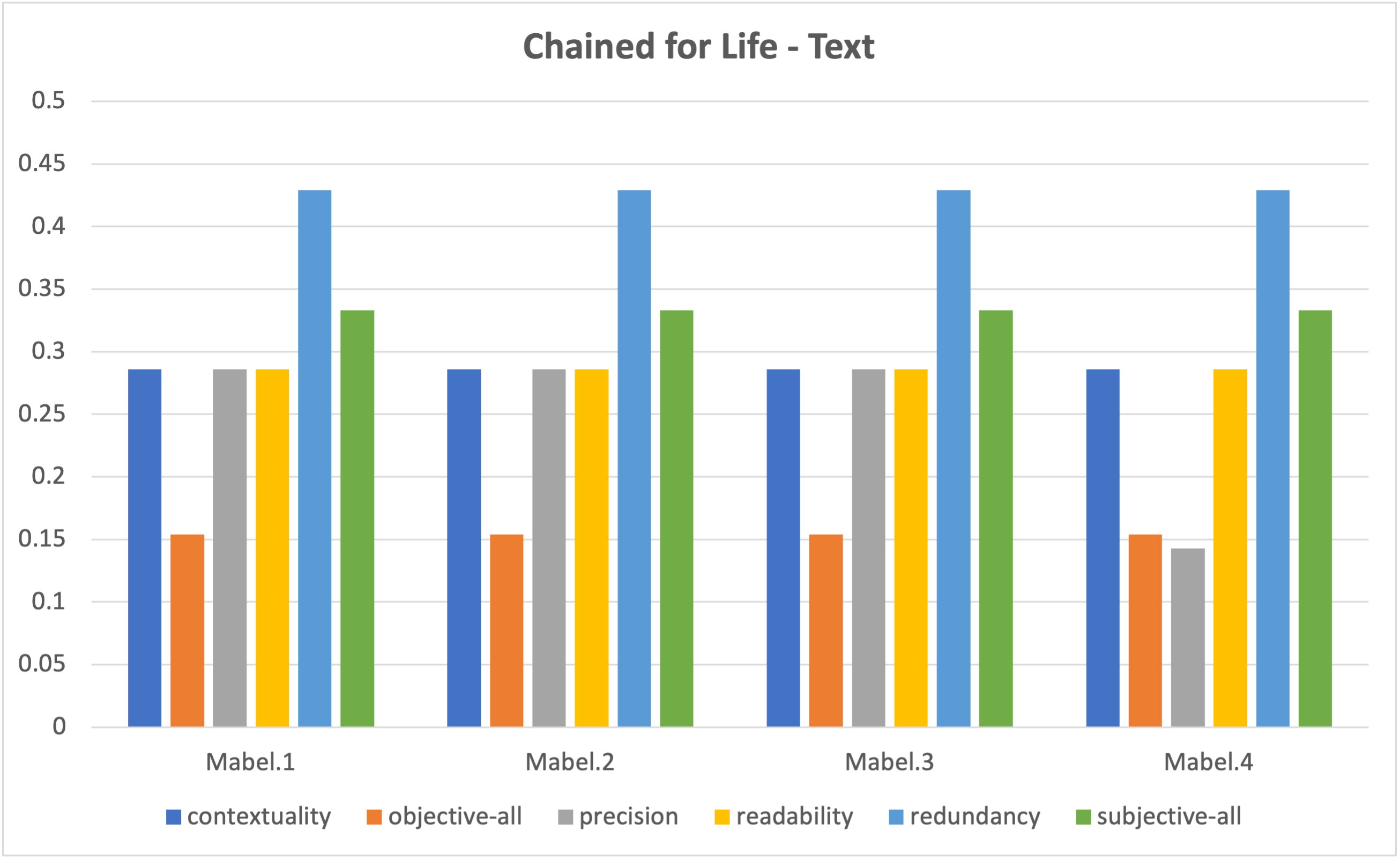}
\caption{MSUM text summarization: Chained for Life}
\label{msum.cfl.text}
\end{center}
\end{figure}

Figure \ref{msum.cfl.text} shows the text summarization task results for the movie \textit{Chained for Life}. This movie also contained just one character for summarization, Mabel. Subjective scores for this movie were higher than for the above text summarization movies, while some objective scores were also achieved in this movie.

\begin{figure}[htbp]
\begin{center}
\includegraphics[height=2.5in,width=3in,angle=0]{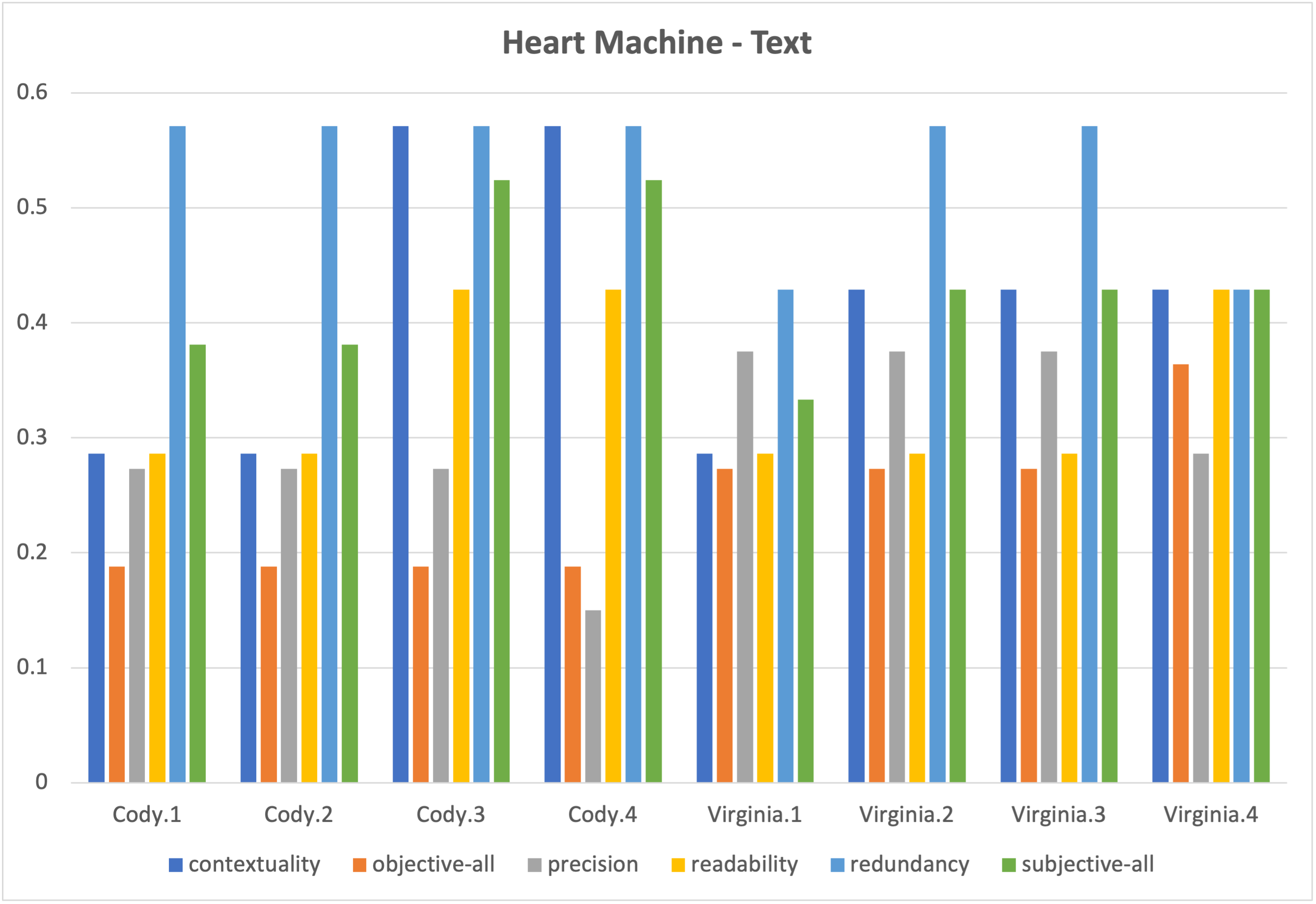}
\caption{MSUM text summarization: Heart Machine}
\label{msum.hm.text}
\end{center}
\end{figure}

Figure \ref{msum.hm.text} shows the text summarization task results for the movie \textit{Heart Machine}. This movie contained two characters for summarization, Cody and Virginia. Virginia score higher than Cody for objective scores, while there is very little difference between characters for subjective scores. Objective scores achieved on this movie were higher again than on previously mentioned movies on the text summarization task.

\begin{figure}[htbp]
\begin{center}
\includegraphics[height=2.5in,width=3in,angle=0]{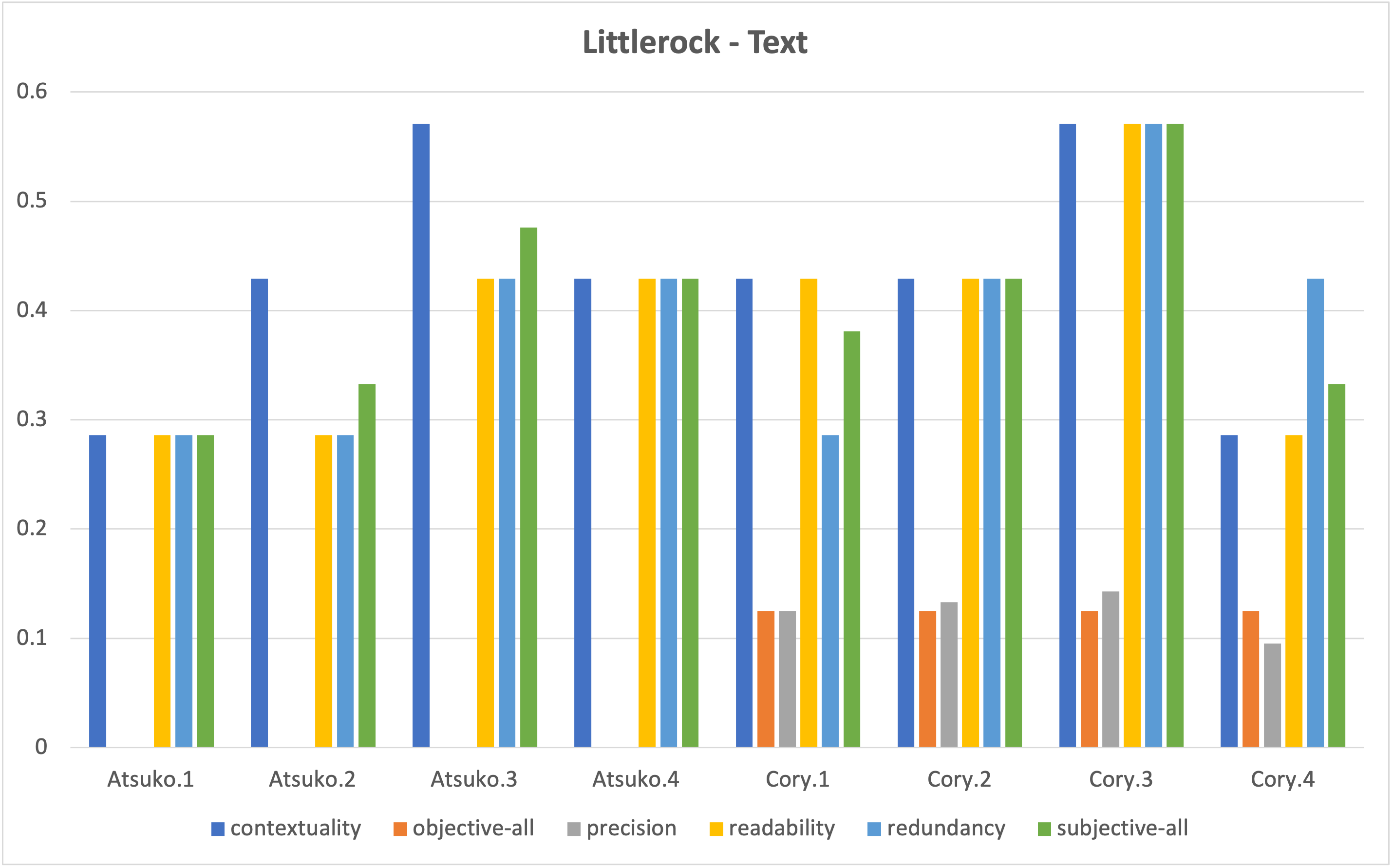}
\caption{MSUM text summarization: Littlerock}
\label{msum.lr.text}
\end{center}
\end{figure}

Figure \ref{msum.lr.text} shows the text summarization task results for the movie \textit{Littlerock}. This movie contained two characters for summarization, Atsuko and Cory. Cory performs better than Atsuko on both objective and subjective metrics. Atsuko achieved zero marks on objective metrics while Cory at least achieves some marks on objective metrics.

\subsubsection{Observations}
This is the first year of the movie summarization task, replacing the previous video summarization which had run for two years. The main difference between these tasks is that the movie summarization task uses individual movies in the dataset rather than BBC Eastenders TV series, and also contains a text summarization subtask. One other issue of note this year was the poor performance for the text summarization subtask. Many movie characters for summarization did not achieve any marks for objective metrics, leading to questions as to whether text summarization techniques are advanced enough for it to be included as a TRECVID task.

We now summarize the approach taken by the finishing team. NII\_UIT first segmented the movie into shots using TransNet\_v2. They then passed shots through a pipeline to calculate a `face similarity ranking score' and `text similarity ranking score'. Scores were then fused to generate a single importance score per shot to determine final shots. To calculate the `face similarity ranking score', MTCNN \cite{zhang2016joint} was used for face detection and VGGFace2 \cite{cao2018vggface2} was used for  face representation. Cosine similarity was then used to match faces in the shot and the input face query. For calculation of `text similarity ranking score', two sources were used: `Audio to Text Similarity Score' and `Video Captioning Similarity Score'. For the audio to text, shots were first passed through a Video Captioning Network, Generated captions were then compared with key-facts from the training set. Bert was used for feature extraction and cosine distance was used for text similarity measurement. For video captioning, shots were passed through an audio to text system. Generated audio captions then went through the same process as for `Audio to Text'. A time score was used to penalize shots which were too long before final fusion of scores to obtain higher scores.

\subsubsection{Conclusions}
This was the first year of the Movie Summarization task. This comprised two subtasks - video summarization and text summarization. Teams were asked to produce video and text summaries of specified characters from 5 movies provided to teams. Between one and two characters were chosen per movie, totaling 8 characters from 5 movies to be summarized. The main challenge of this task was to select only shots that contribute to a character's main story-line. Another big challenge was that these shots did not necessarily need to contain that specific character, as other characters may have talked about him/her in a way that exposes quite a lot of the story to the viewer.

Objective results (recall of key-facts) were better for Littlerock video summary than for any other submitted video summaries. For text summaries, objective results (recall of key-facts) were better for Heart Machine than for any other submitted text summaries. In general, objective results were also much better for video summaries than for text summaries. Subjective results (tempo, contextuality, redundancy) were better for Heart Machine video summary than for any other summaries. For text summaries, subjective results (tempo, contextuality, redundancy) were very similar for Heart Machine and Littlerock. Both performed much better than any other text summaries. In general, results for text summaries were much lower than results for video summaries. 

Only one participating team finished this task, far below the desired participation rate. While video summary submissions were satisfactory, including some quite good summaries, text summary submissions were quite disappointing and below the expected standard. This leads to questions as to whether this should be continued in subsequent years. If this task is to remain, a debate still needs to happen about whether to include the text summary sub-task. A lot of submitted text summaries were well below expectations, particularly the readability of these summaries.

\section{Summing up and moving on}
In this overview paper to TRECVID 2022, we provided basic information for all
tasks we run this year and particularly on the goals, data, evaluation mechanisms, and metrics used. 
Further details about each particular group's approach and performance for each task 
can be found in that group's site report. The raw results for each submitted run 
can be found at the online proceeding of the workshop \cite{tv22pubs}.
Finally, we are looking forward to continuing a new evaluation cycle in 2023 after
refining the current tasks and introducing any potential new tasks.
\section{Authors' note}
TRECVID would not have happened in 2022 without support from the
National Institute of Standards and Technology (NIST). The research
community is very grateful for this. Beyond that, various individuals
and groups deserve special thanks:
\begin{itemize}

\item{Koichi Shinoda of the TokyoTech team agreed to host a copy of 
IACC.2 data.}

\item{Georges Qu\'{e}not provided the master shot reference for the
IACC.3 videos.}

\item{The LIMSI Spoken Language Processing Group and Vocapia Research
provided ASR for the IACC.3 videos.}

\item{Luca Rossetto of University of Basel for providing the V3C dataset collection.}

\item{Jeffrey Liu and Andrew Weinert of MIT Lincoln Laboratory for supporting the DSDI task
      by making the LADI dataset available and helping with the testing dataset preparations.} 

\item{Baptiste Chocot of NIST associate for supporting the previous ActEV task.}
\end{itemize}

Finally, we want to thank all the participants and other contributors
on the mailing list for their energy and perseverance.

\section{Acknowledgments} 
The ActEV NIST work was partially supported by the Intelligence Advanced Research Projects Activity (IARPA), agreement~IARPA-16002. The authors would like to thank Kitware, Inc. for annotating the dataset. 
The Video-to-Text work has been partially supported by Science Foundation Ireland (SFI) as a part of the Insight Centre at Dublin City University (12/RC/2289) and grant  number  13/RC/2106 (ADAPT Centre for Digital Content Technology, \url{www.adaptcentre.ie})  at  Trinity College Dublin.
We would like to thank Tim Finin and Lushan Han of University of Maryland, Baltimore County for providing access to the semantic similarity metric.
Finally, the TRECVID team at NIST would like to thank all external coordinators for their efforts across the different tasks they helped to coordinate.

\bibliography{video}

\clearpage
\onecolumn

\appendix
\section{Ad-hoc 2022 query topics} 
\label{appendixA}
\begin{description}\itemsep0pt \parskip0pt

\item[701]  A man with a white beard
\item[702]  A room with blue wall
\item[703]  A construction site
\item[704]  A parked white car
\item[705]  A type of cloth hanging on a rack, hanger, or line
\item[706]  Building with columns during daytime
\item[707]  A person is mixing ingredients in a bowl, cup, or similar type of containers
\item[708]  A female person bending downwards
\item[709]  A person is in the act of swinging
\item[710]  A person wearing a light t-shirt with dark or black writing on it
\item[711]  A woman wearing a head kerchief
\item[712]  A man wearing black shorts
\item[713]  A kneeling man outdoors
\item[714]  Two or more persons in a room with a fireplace
\item[715]  An Asian bride and groom celebrating outdoors
\item[716]  A drone landing or rising from the ground
\item[717]  A black bird seen on a dry area sitting, walking, or eating
\item[718]  A large stone building from the outside
\item[719]  A piece of heavy farm equipment or machine seen outdoors
\item[720]  A clock on a wall in a room
\item[721]  Two persons are seen while at least one of them is speaking in a non-English language outdoors
\item[722]  A woman is eating something outdoors
\item[723]  A person is biking through a path in a forest
\item[724]  A man and a bike in the air after jumping from a ramp
\item[725]  A woman holding or smoking a cigarette
\item[726]  Two teams playing a game where one team have their players wearing white t-shirts.
\item[727]  Two persons wearing white outfits and black belts demonstrate martial arts in a room with floor mats
\item[728]  Two adults are seated in a flying paraglider in the air
\item[729]  A ring shown on the left hand of a person
\item[730]  A man is holding a knife in a non-kitchen location

\end{description}

\section{Ad-hoc query topics - 20 progress topics} 
\label{appendixB}
\begin{description}\itemsep0pt \parskip0pt
\item[681]  A woman with a ponytail
\item[682]  A person's Hands with a red nail polish
\item[683]  A building with balconies seen from the outside during daytime
\item[684]  A room with a wood floor
\item[685]  A wooden bridge
\item[686]  A round table
\item[687]  A person is throwing an object away
\item[688]  A person is washing oneself or another thing
\item[689]  A man wearing a lanyard around his neck
\item[690]  A man is seen at a gas station
\item[691]  A vehicle driving under a tunnel
\item[692]  A big building that is being camera panned or tilted from the outside
\item[693]  A person is lying on the ground outdoors
\item[694]  A person is rubbing part of their face using their hands
\item[695]  A man holding a gun but not shooting
\item[696]  A person is pouring liquid into a type of container
\item[697]  A man holding a fishing rod while being dipped in a body of water
\item[698]  A person holding a long stick which is not a drum stick outdoors
\item[699]  A person wearing a ring in their nose
\item[700]  A man wearing a dark colored hooded jacket outdoors

\end{description}

\end{document}